\documentclass[preprint,12pt]{elsarticle}

\usepackage{graphicx}
\usepackage{amsmath,amsfonts,amssymb}
\usepackage{booktabs}
\usepackage{hyperref}
\usepackage{cleveref}
\usepackage{multirow}
\usepackage{caption}
\usepackage{subcaption}

\usepackage{threeparttable} 
\usepackage{array}

\begin{document}

\begin{frontmatter}

\title{GS‐Light: Training-Free Multi-View Extension of IC-Light for Textual Position-Aware Scene Relighting}

\author[a]{Jiangnan Ye}
\author[a]{Jiedong Zhuang}
\author[a]{Lianrui Mu}
\author[a]{Wenjie Zheng}
\author[a]{Jiaqi Hu}
\author[a]{Xingze Zou}
\author[a]{Jing Wang}
\author[a]{Haoji Hu\corref{cor1}}

\address[a]{College of Information Science and Electronic Engineering, Zhejiang
University, Hangzhou, 310027, Zhejiang, China}

\cortext[cor1]{Corresponding author.}

\cortext[email]{Email addresses: 
jiangnan\_ye@zju.edu.cn (Jiangnan Ye), zhuangjiedong@zju.edu.cn (Jiedong Zhuang), mulianrui@zju.edu.cn (Lianrui Mu), wenjie\_zheng@zju.edu.cn (Wenjie Zheng), jiaqi\_hu@zju.edu.cn (Jiaqi Hu), zeezou@zju.edu.cn (Xingze Zou), j\_wang@zju.edu.cn (Jing Wang), haoji\_hu@zju.edu.cn (Haoji Hu)}

\begin{highlights}
\item GS-Light enables fast, position-aware multi-view relighting in 3DGS.
\item Lighting priors extracted via LVLM + geometry \& semantics.
\item Cross-View Attention enforce multi-view consistency in relit outputs.
\item Outperforms baselines with better quality in diverse scenes.
\end{highlights}

\begin{abstract}
We introduce \textbf{GS-Light}, an efficient, textual position-aware pipeline for text-guided relighting of 3D scenes represented via Gaussian Splatting (3DGS). GS-Light implements a \textbf{training-free extension} of a single-input diffusion model to handle multi-view inputs. Given a user prompt that may specify lighting direction, color, intensity, or reference objects, we employ a large vision-language model (LVLM) to \textbf{parse the prompt into lighting priors}. Using off-the-shelf estimators for geometry and semantics (depth, surface normals, and semantic segmentation), we fuse these lighting priors with view-geometry constraints to compute illumination maps and generate initial latent codes for each view. These meticulously derived init latents guide the diffusion model to generate relighting outputs that more accurately reflect user expectations, especially in terms of lighting direction. By feeding multi-view rendered images, along with the init latents, into our multi-view relighting model, we produce high-fidelity, artistically relit images. Finally, we fine-tune the 3DGS scene with the relit appearance to obtain a fully relit 3D scene. We evaluate GS-Light on both indoor and outdoor scenes, comparing it to state-of-the-art baselines including per-view relighting, video relighting, and scene editing methods. Using quantitative metrics (multi-view consistency, imaging quality, aesthetic score, semantic similarity, etc.) and qualitative assessment (user studies), GS-Light demonstrates consistent improvements over baselines. Code and assets will be made available upon publication.
\end{abstract}


\end{frontmatter}

\newcolumntype{C}[1]{>{\centering\arraybackslash}m{#1}}

\begin{figure}[h!]
\centering
\resizebox{\textwidth}{!}{
\setlength{\tabcolsep}{0pt}
\begin{tabular}{c@{\hspace{0.3em}}ccc@{\hspace{0.3em}}cccccc}
\hline
& \multicolumn{3}{c}{Relit Video} & \multicolumn{6}{c}{Relit Gasussian Splatting} \\
\textbf{\small Source} & \textbf{\small Ours} & \textbf{\small RelightVid} & \textbf{\small Lumen} & \textbf{\small Ours} & \textbf{\small DGE} & \textbf{\small EditSplat} & \textbf{\small IN2N} & \textbf{\small IGS2GS} & \textbf{\small IGS2GS-IC}\\

\vspace{-0.34em}
\includegraphics[width=0.16\textwidth]{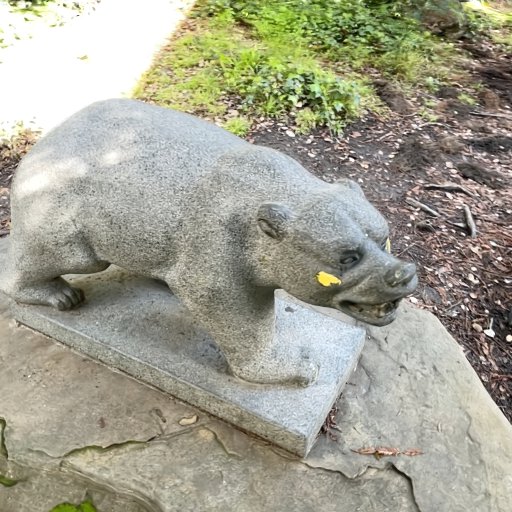} &
\includegraphics[width=0.16\textwidth]{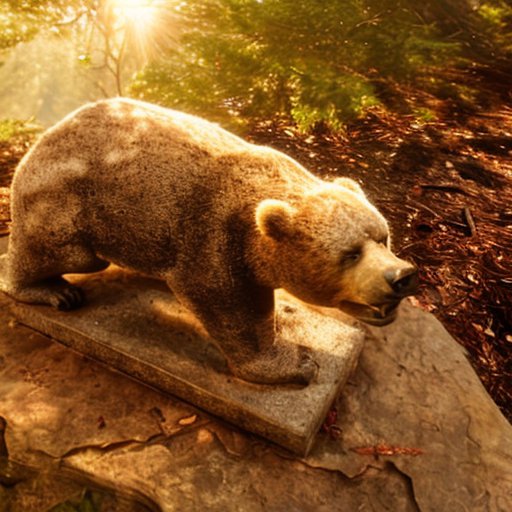} &
\includegraphics[width=0.16\textwidth]{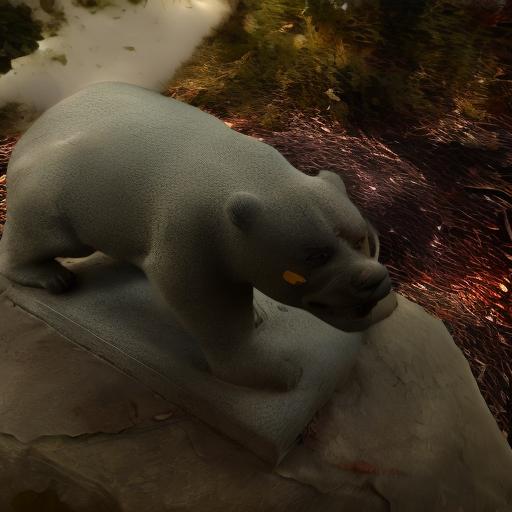} &
\includegraphics[width=0.16\textwidth]{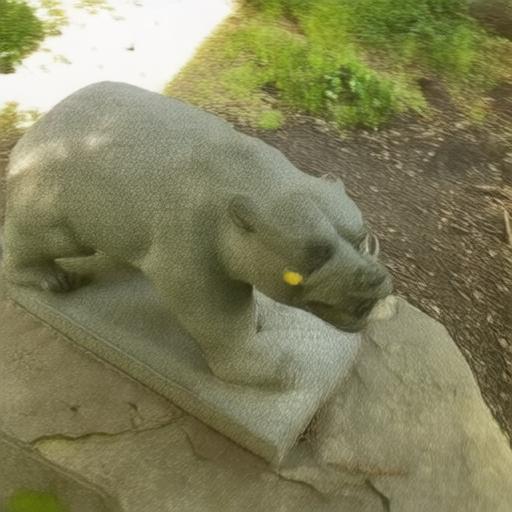} &
\includegraphics[width=0.16\textwidth]{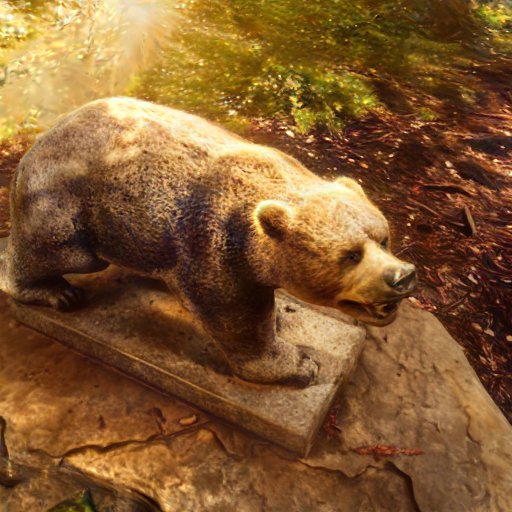} &
\includegraphics[width=0.16\textwidth]{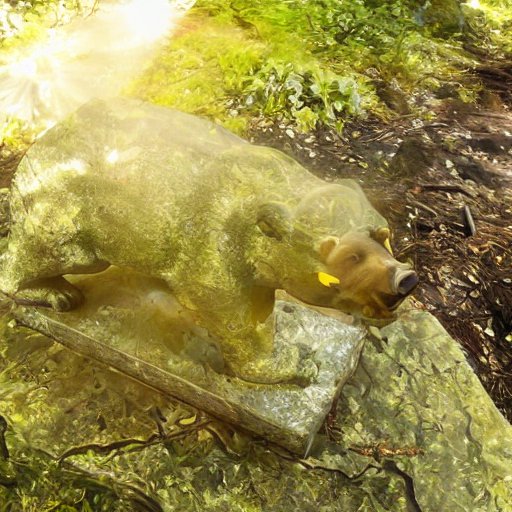} &
\includegraphics[width=0.16\textwidth]{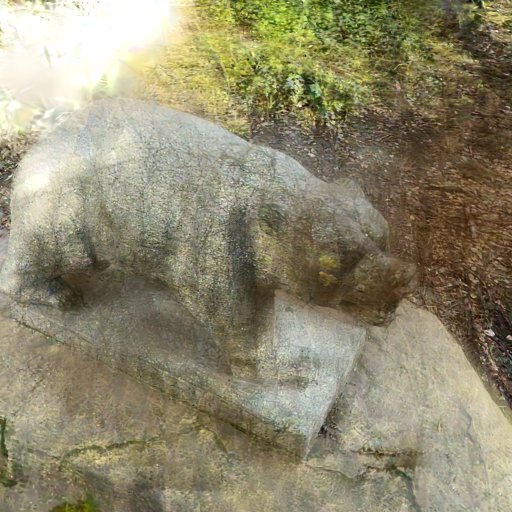} & \includegraphics[width=0.16\textwidth]{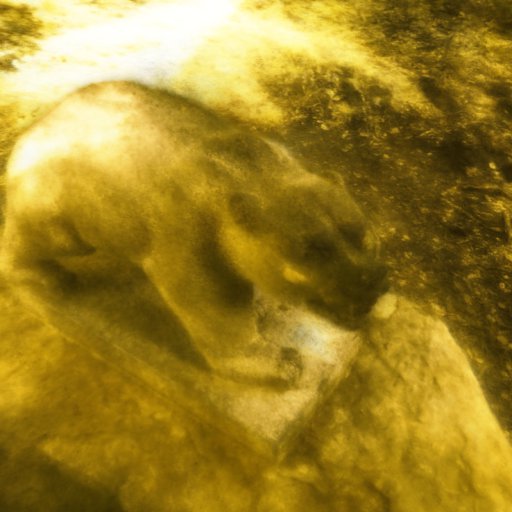} &
\includegraphics[width=0.16\textwidth]{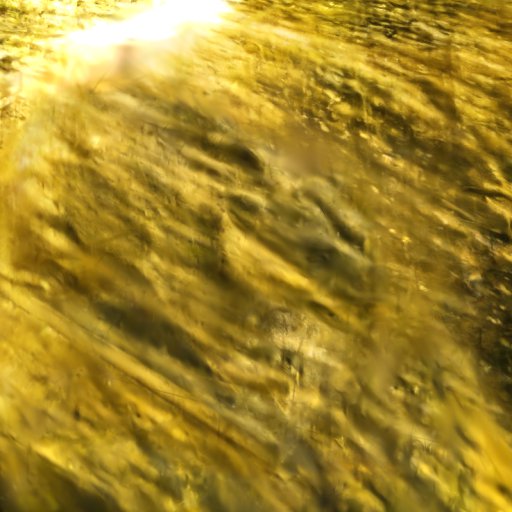} &
\includegraphics[width=0.16\textwidth]{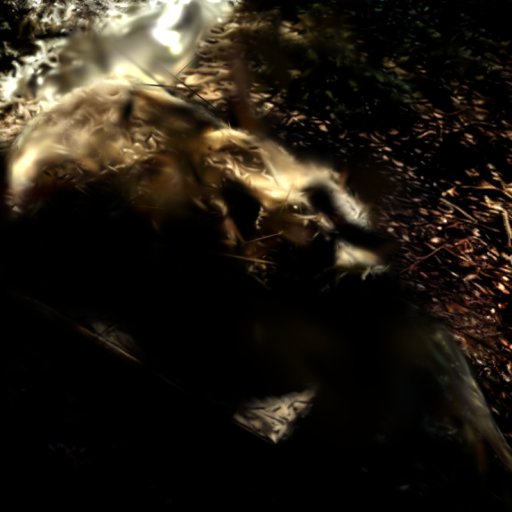} \\

\includegraphics[width=0.16\textwidth]{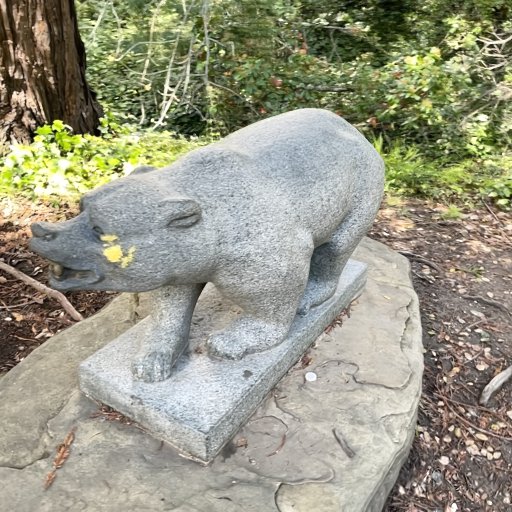} &
\includegraphics[width=0.16\textwidth]{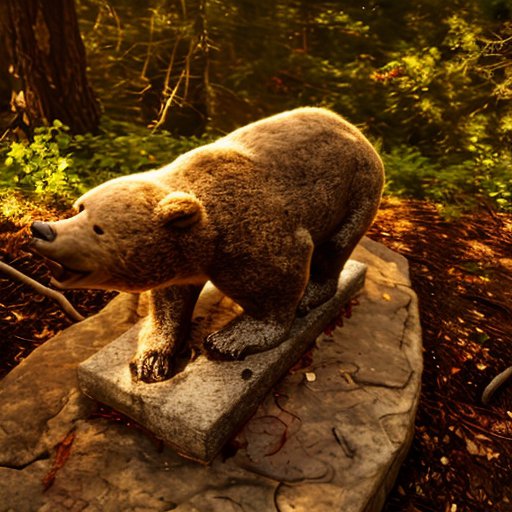} &
\includegraphics[width=0.16\textwidth]{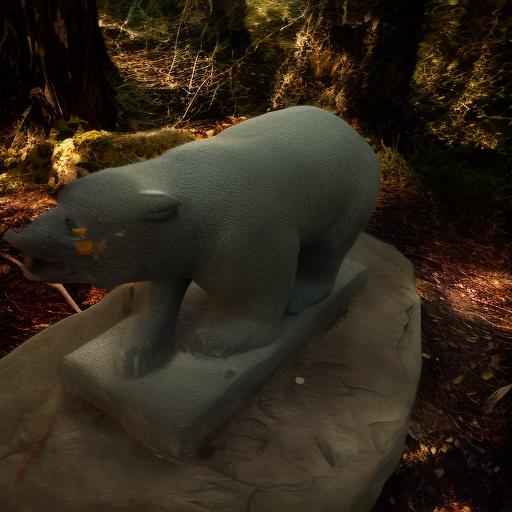} &
\includegraphics[width=0.16\textwidth]{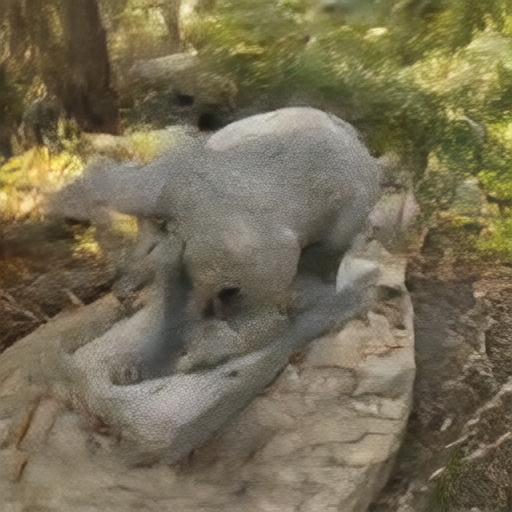} &
\includegraphics[width=0.16\textwidth]{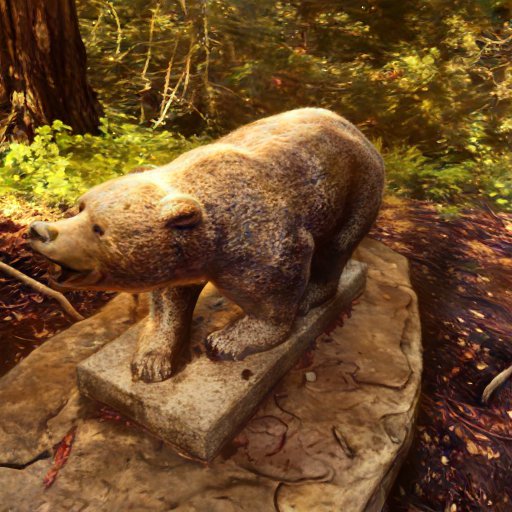} &
\includegraphics[width=0.16\textwidth]{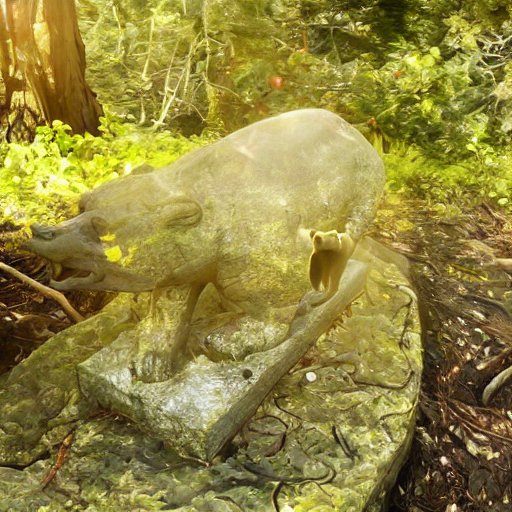} &
\includegraphics[width=0.16\textwidth]{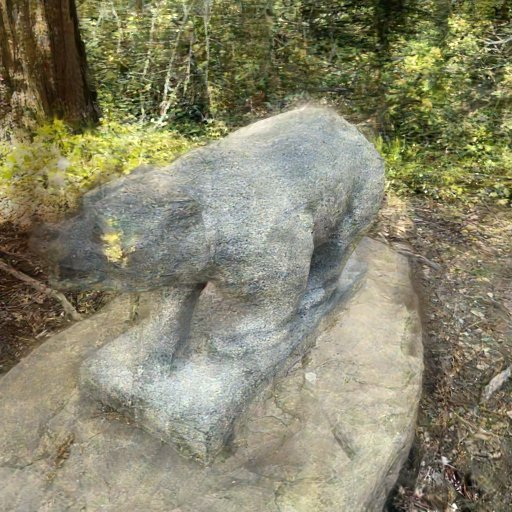} & \includegraphics[width=0.16\textwidth]{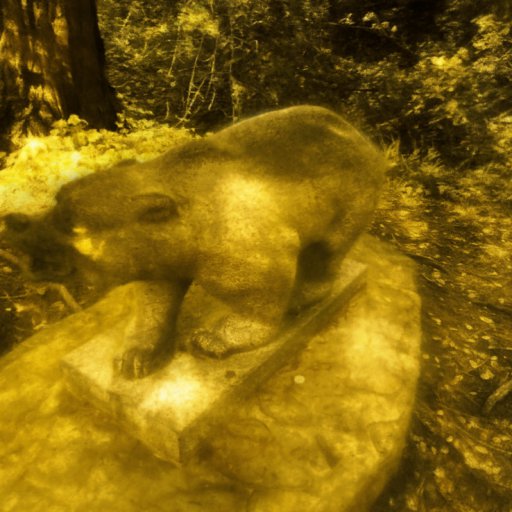} &
\includegraphics[width=0.16\textwidth]{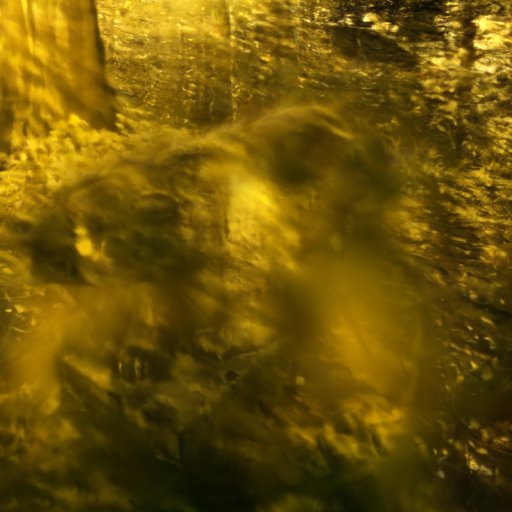} &
\includegraphics[width=0.16\textwidth]{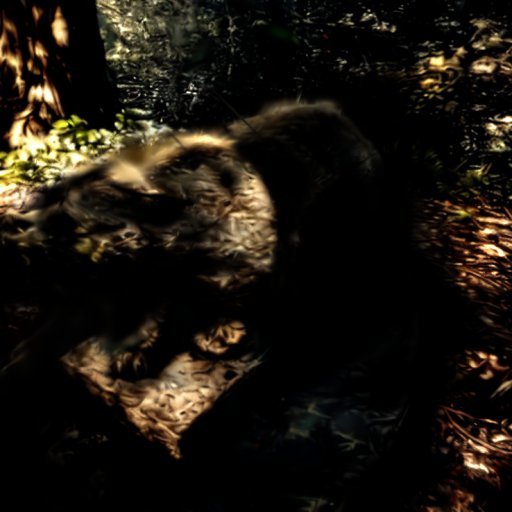} \\

\multicolumn{10}{c}{"\textit{bear, forest, sunlight filtering through trees, natural lighting, warm atmosphere, \textbf{light from top}}"} \\

\vspace{-0.34em}
\includegraphics[width=0.16\textwidth]{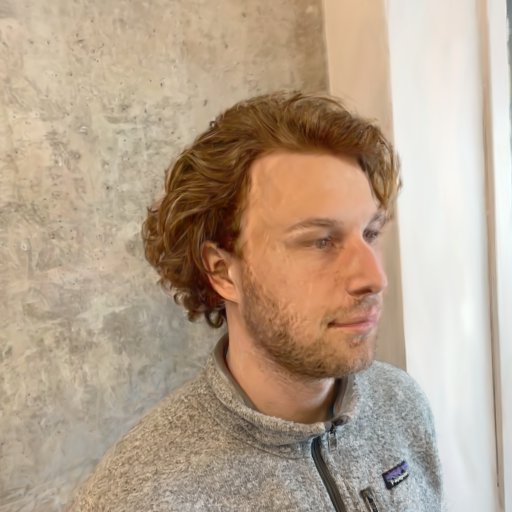} &
\includegraphics[width=0.16\textwidth]{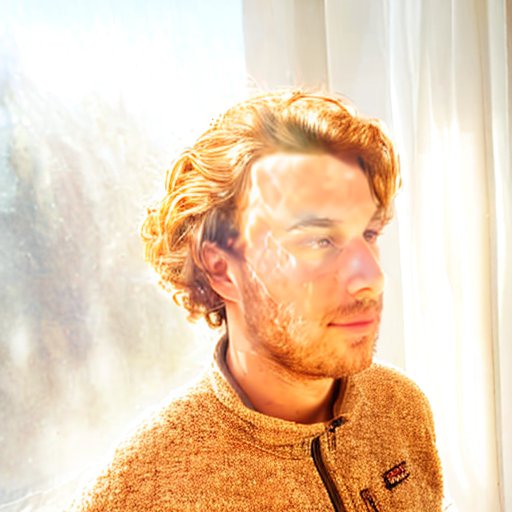} &
\includegraphics[width=0.16\textwidth]{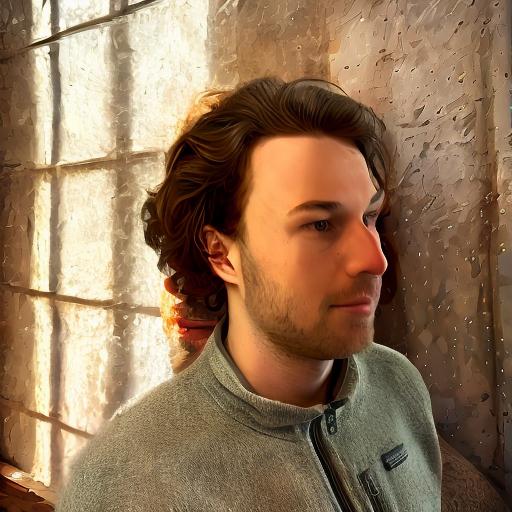} &
\includegraphics[width=0.16\textwidth]{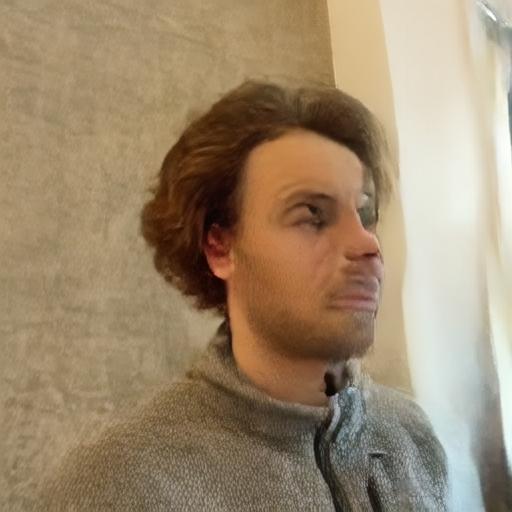} &
\includegraphics[width=0.16\textwidth]{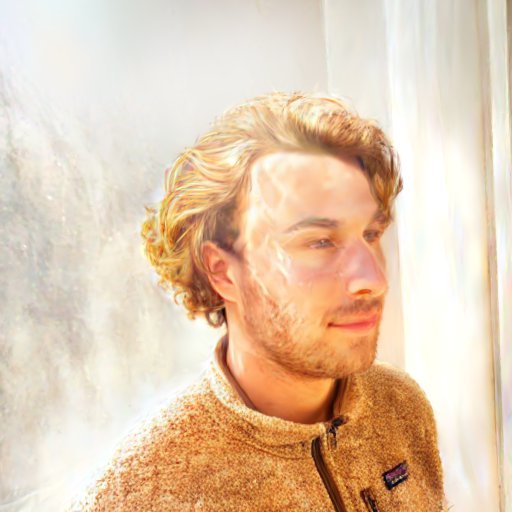} &
\includegraphics[width=0.16\textwidth]{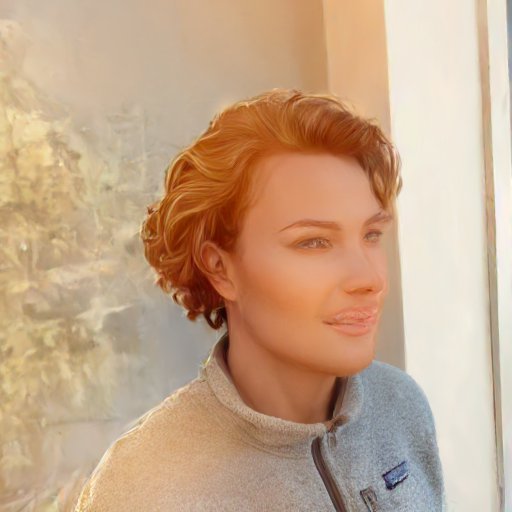} &
\includegraphics[width=0.16\textwidth]{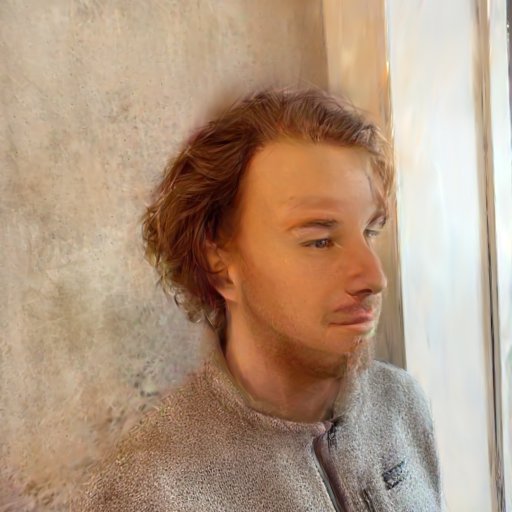} & \includegraphics[width=0.16\textwidth]{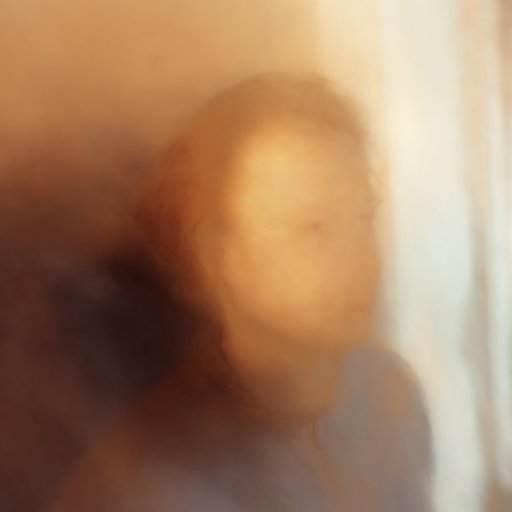} &
\includegraphics[width=0.16\textwidth]{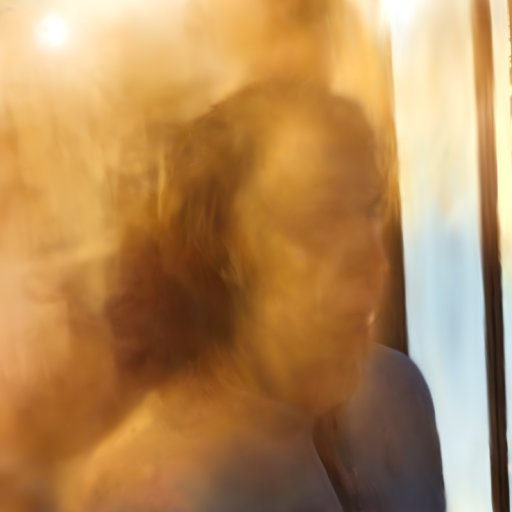} &
\includegraphics[width=0.16\textwidth]{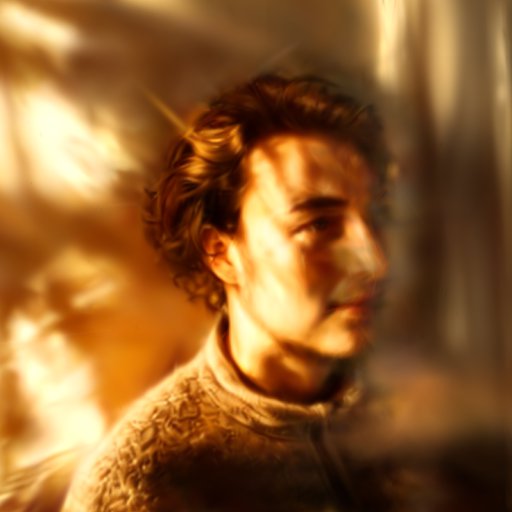} \\

\includegraphics[width=0.16\textwidth]{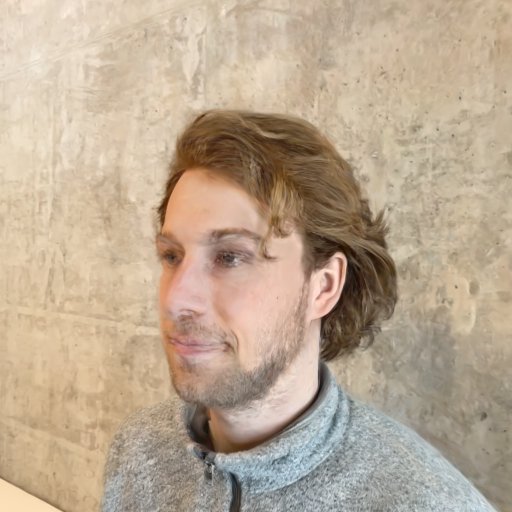} &
\includegraphics[width=0.16\textwidth]{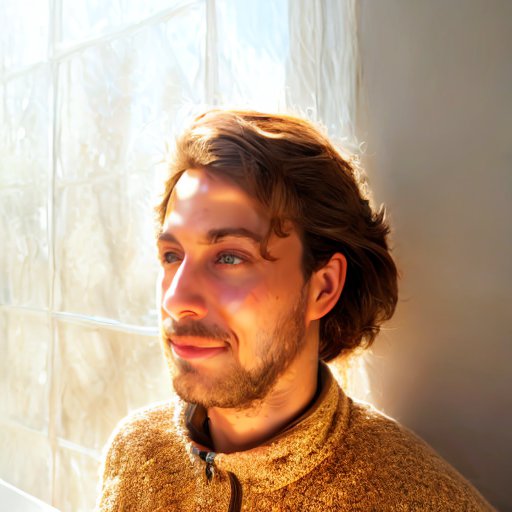} &
\includegraphics[width=0.16\textwidth]{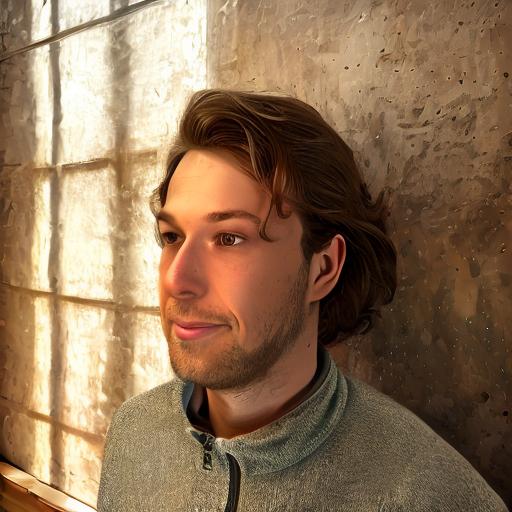} &
\includegraphics[width=0.16\textwidth]{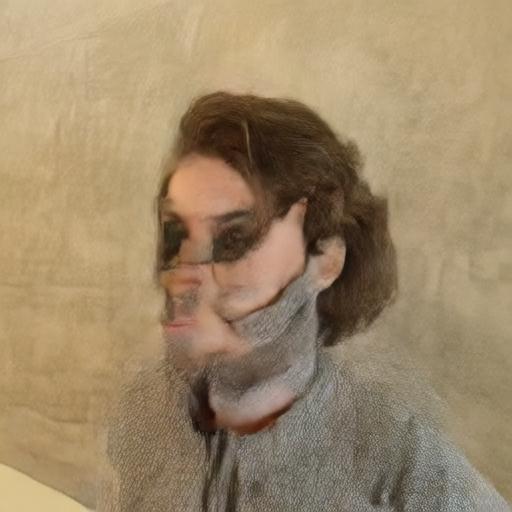} &
\includegraphics[width=0.16\textwidth]{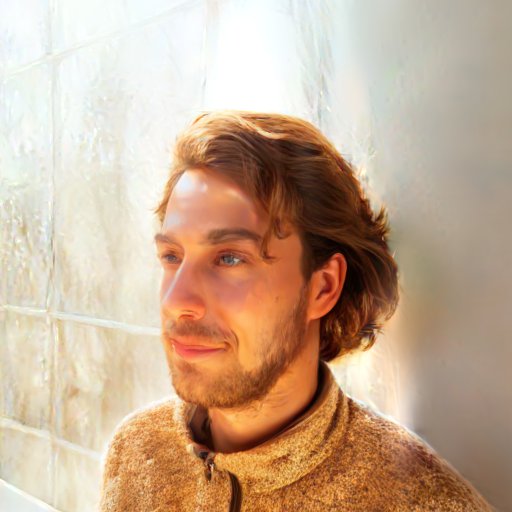} &
\includegraphics[width=0.16\textwidth]{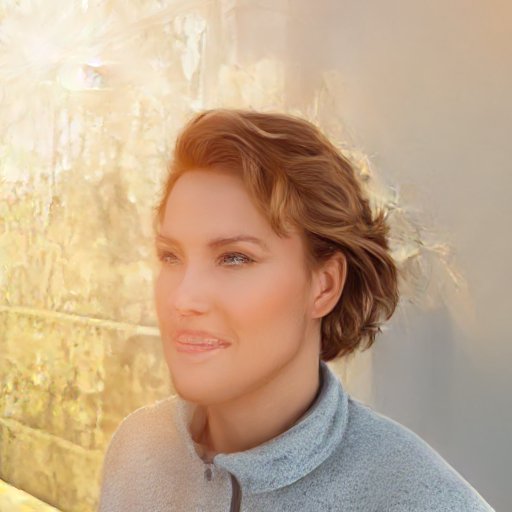} &
\includegraphics[width=0.16\textwidth]{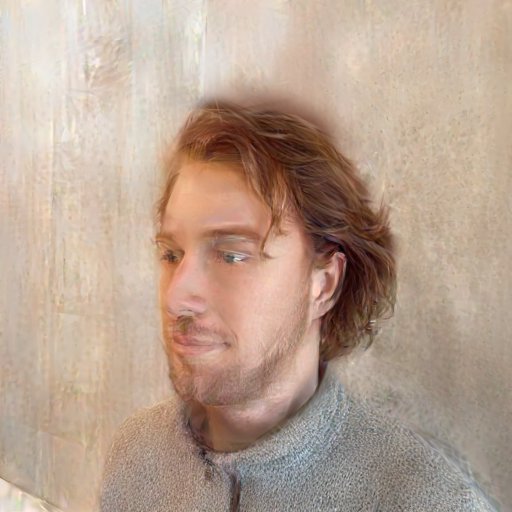} & \includegraphics[width=0.16\textwidth]{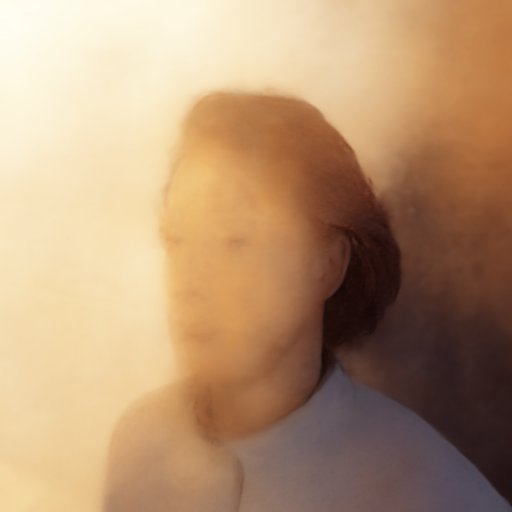} &
\includegraphics[width=0.16\textwidth]{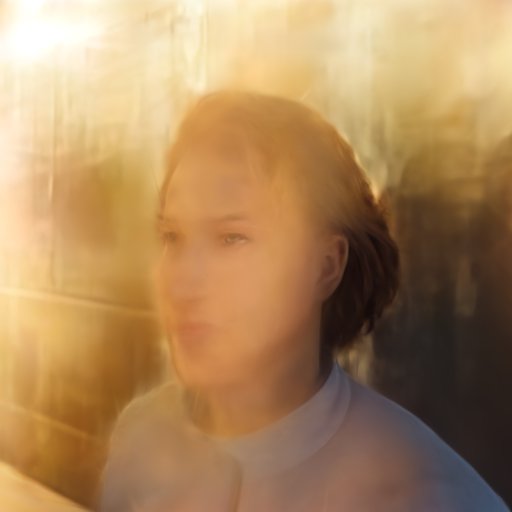} &
\includegraphics[width=0.16\textwidth]{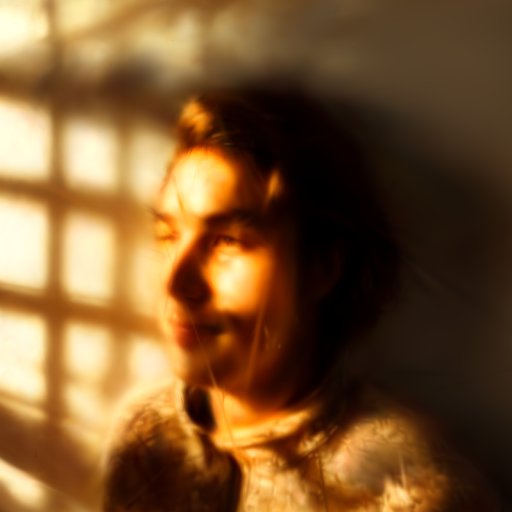} \\

\multicolumn{10}{c}{"\textit{portrait, detailed face, sunshine from window, warm atmosphere, \textbf{light from the left}}"} \\

\vspace{-0.34em}
\includegraphics[width=0.16\textwidth]{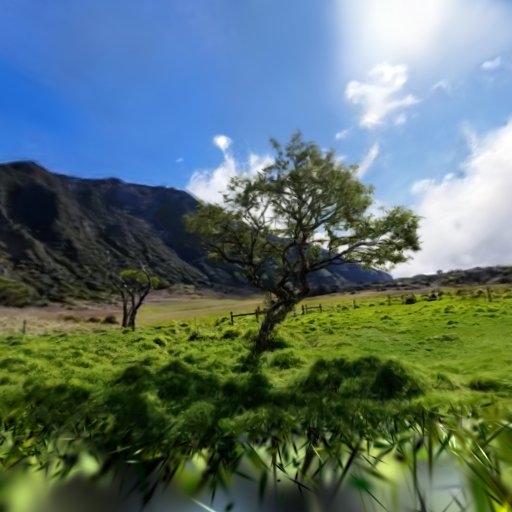} &
\includegraphics[width=0.16\textwidth]{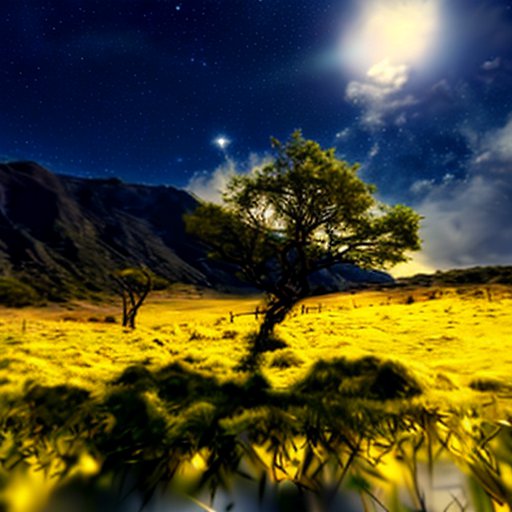} &
\includegraphics[width=0.16\textwidth]{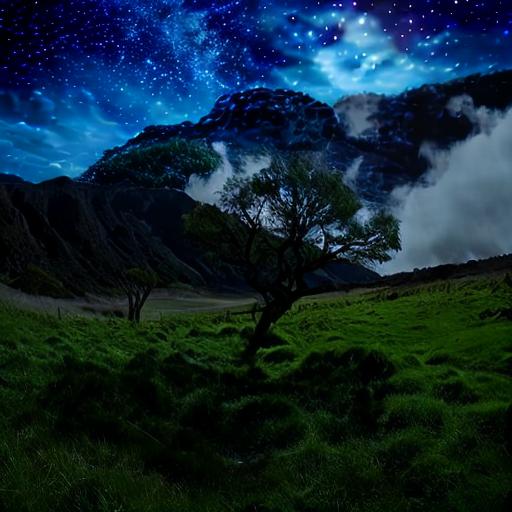} &
\includegraphics[width=0.16\textwidth]{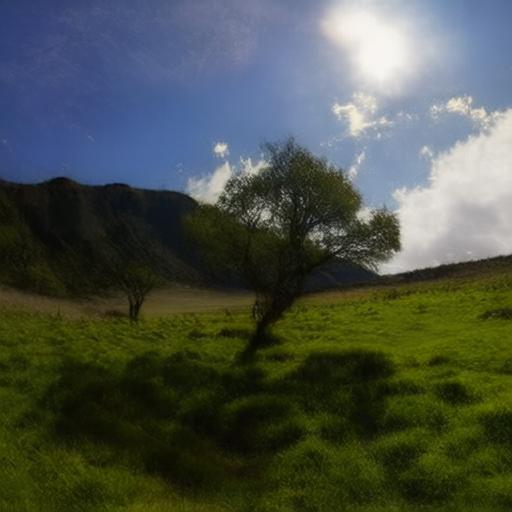} &
\includegraphics[width=0.16\textwidth]{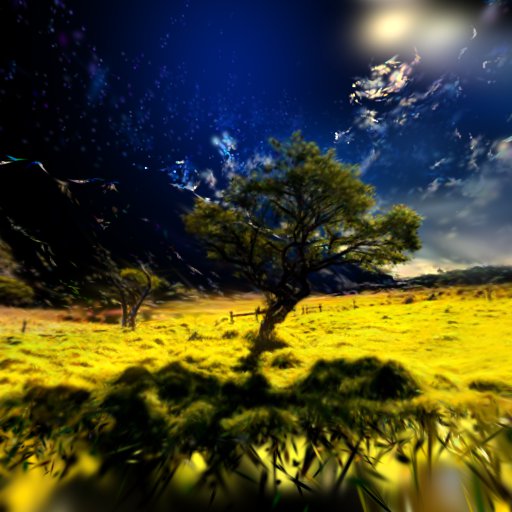} &
\includegraphics[width=0.16\textwidth]{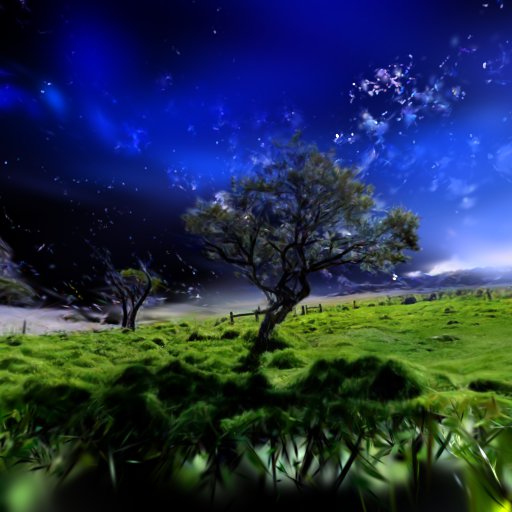} &
\includegraphics[width=0.16\textwidth]{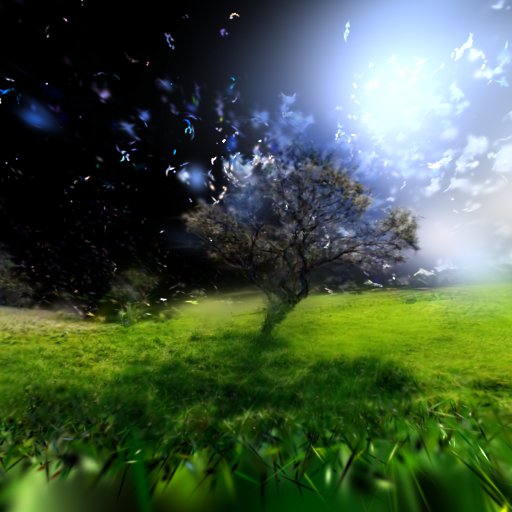} & \includegraphics[width=0.16\textwidth]{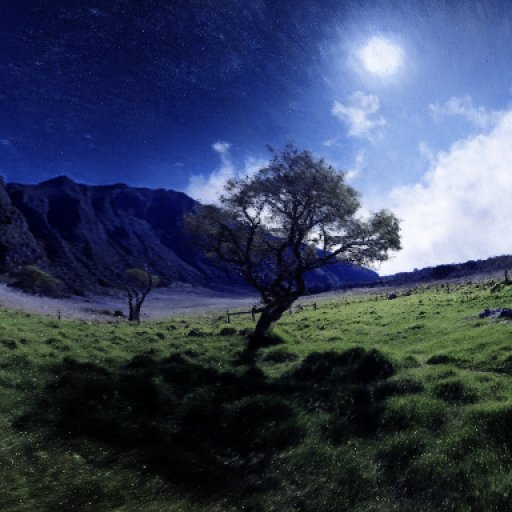} &
\includegraphics[width=0.16\textwidth]{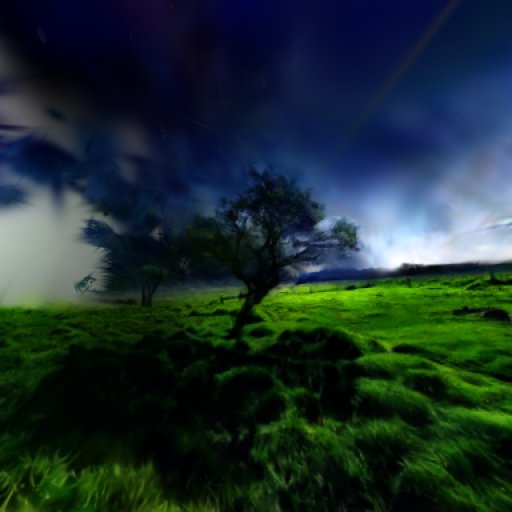} &
\includegraphics[width=0.16\textwidth]{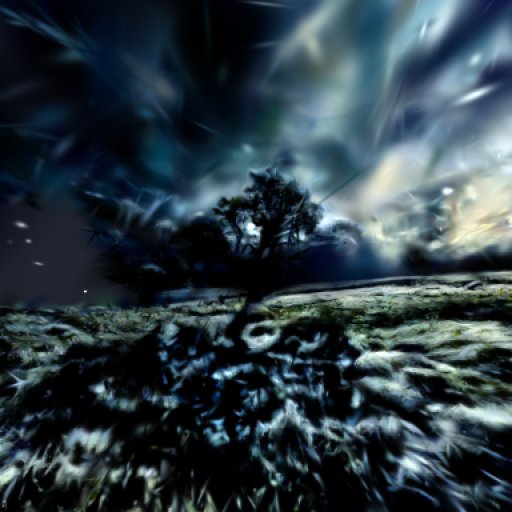} \\

\includegraphics[width=0.16\textwidth]{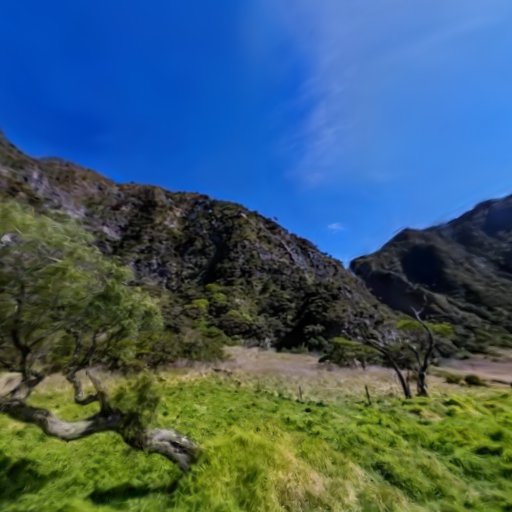} &
\includegraphics[width=0.16\textwidth]{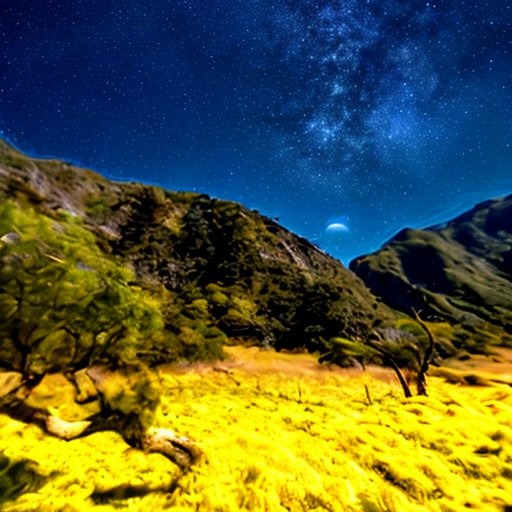} &
\includegraphics[width=0.16\textwidth]{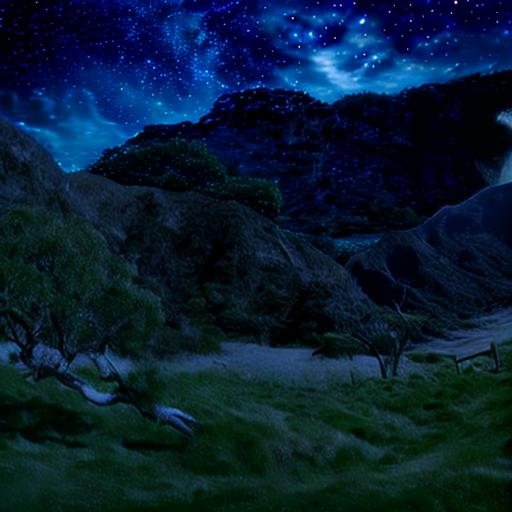} &
\includegraphics[width=0.16\textwidth]{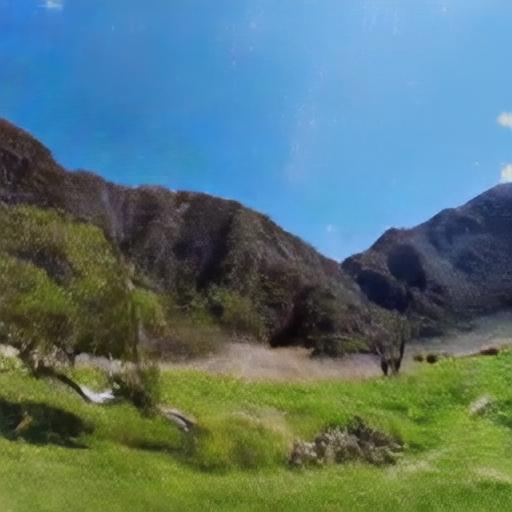} &
\includegraphics[width=0.16\textwidth]{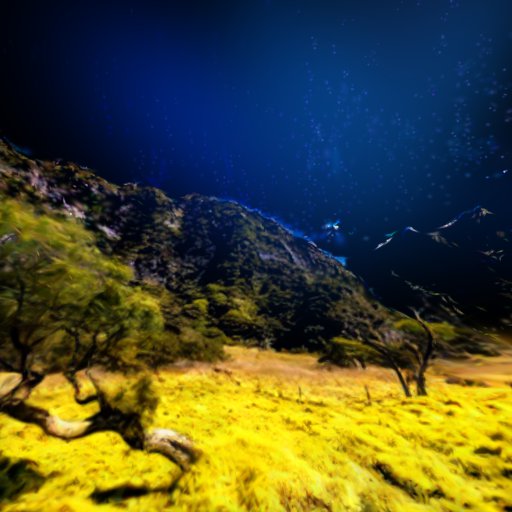} &
\includegraphics[width=0.16\textwidth]{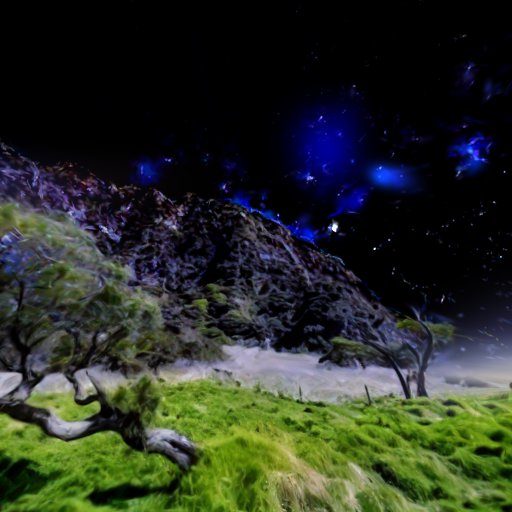} &
\includegraphics[width=0.16\textwidth]{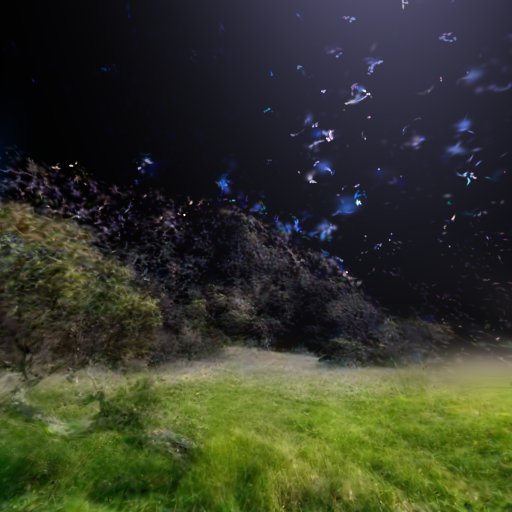} & \includegraphics[width=0.16\textwidth]{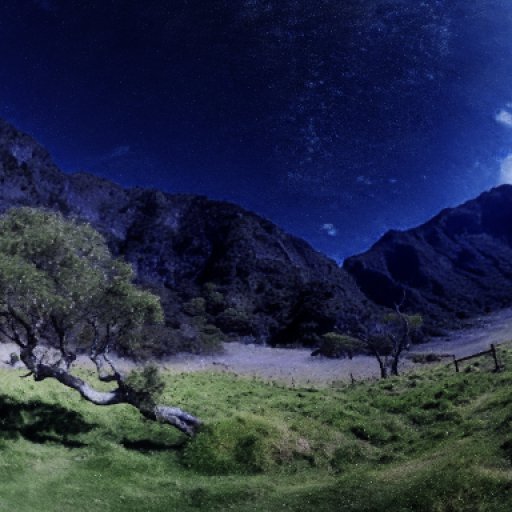} &
\includegraphics[width=0.16\textwidth]{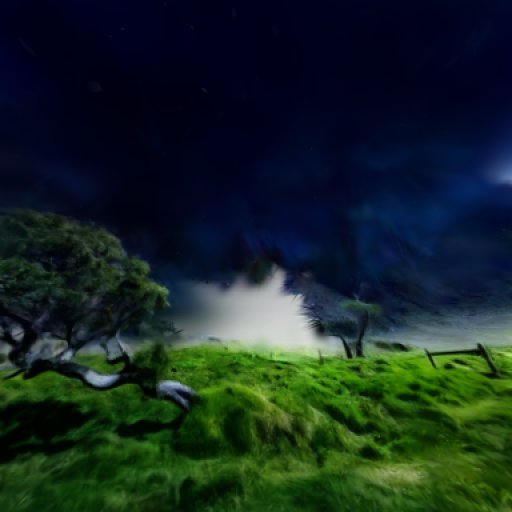} &
\includegraphics[width=0.16\textwidth]{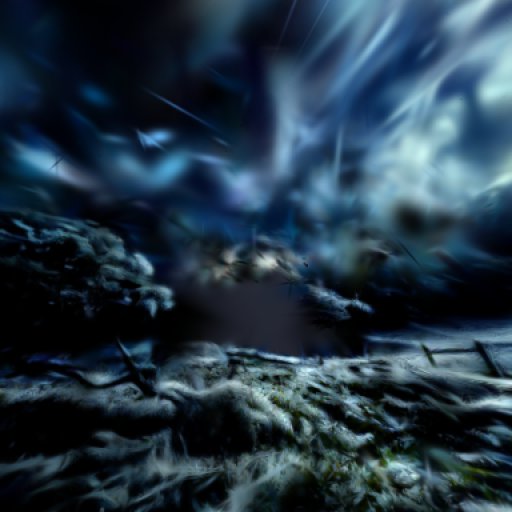} \\

\multicolumn{10}{c}{"\textit{field, outdoor, starry night, silver moonlight, tranquil tone}"} \\

\vspace{-0.34em}
\includegraphics[width=0.16\textwidth]{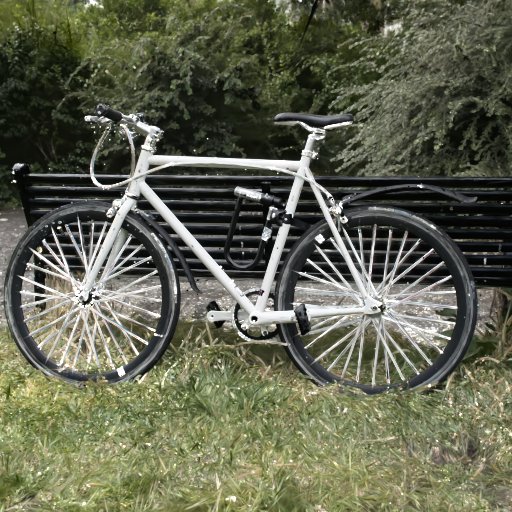} &
\includegraphics[width=0.16\textwidth]{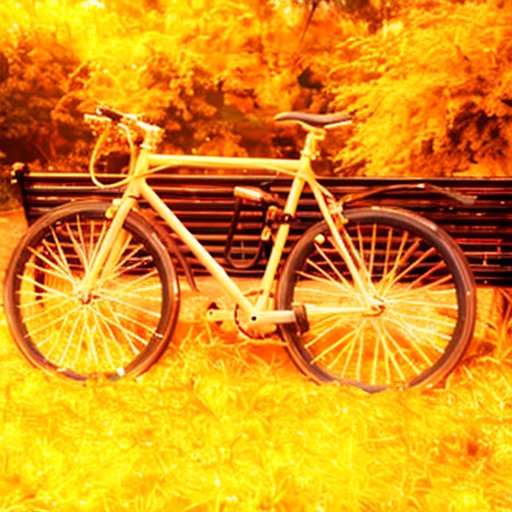} &
\includegraphics[width=0.16\textwidth]{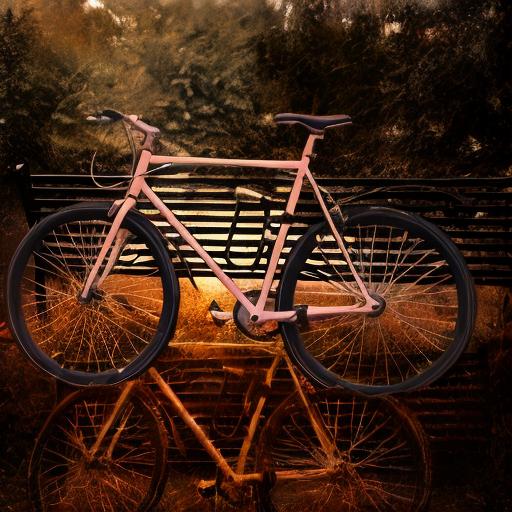} &
\includegraphics[width=0.16\textwidth]{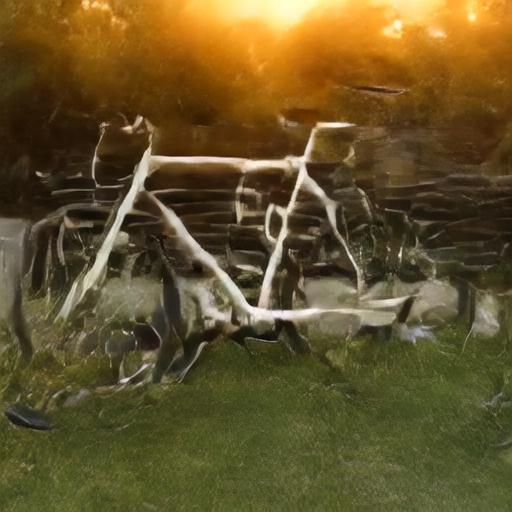} &
\includegraphics[width=0.16\textwidth]{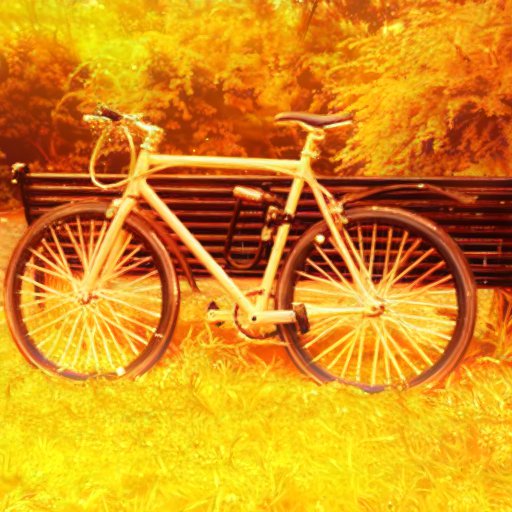} &
\includegraphics[width=0.16\textwidth]{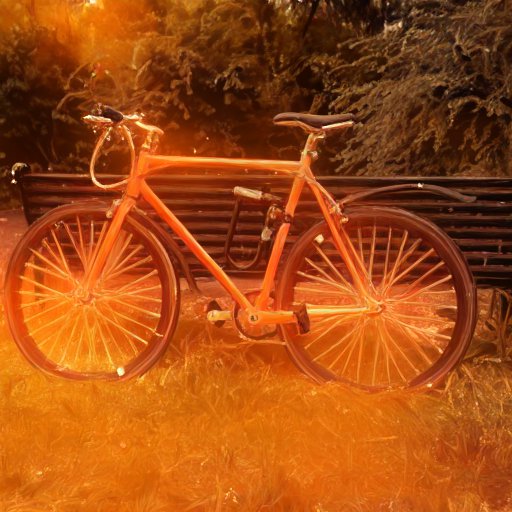} &
\includegraphics[width=0.16\textwidth]{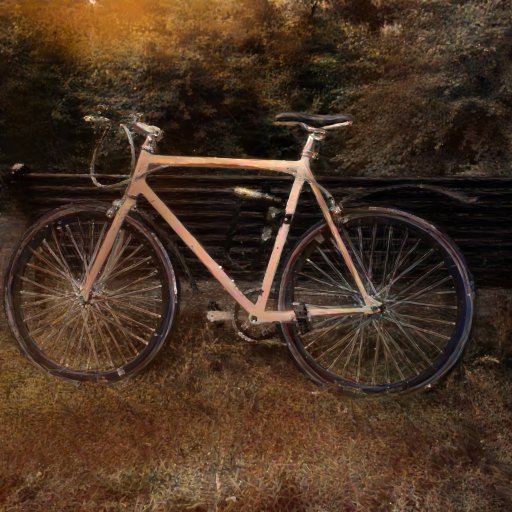} & \includegraphics[width=0.16\textwidth]{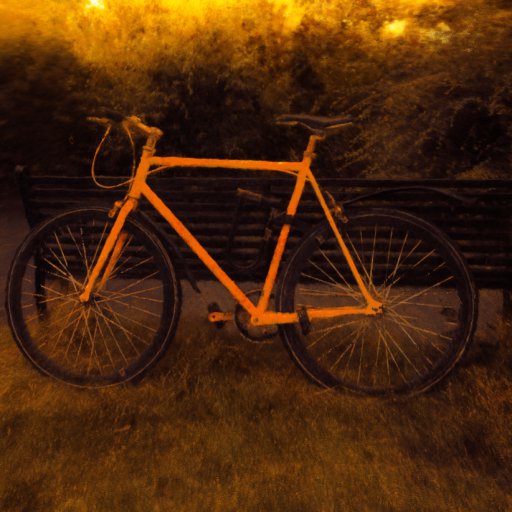} &
\includegraphics[width=0.16\textwidth]{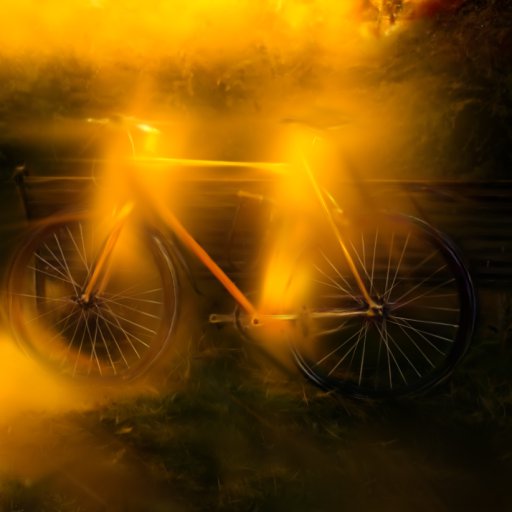} &
\includegraphics[width=0.16\textwidth]{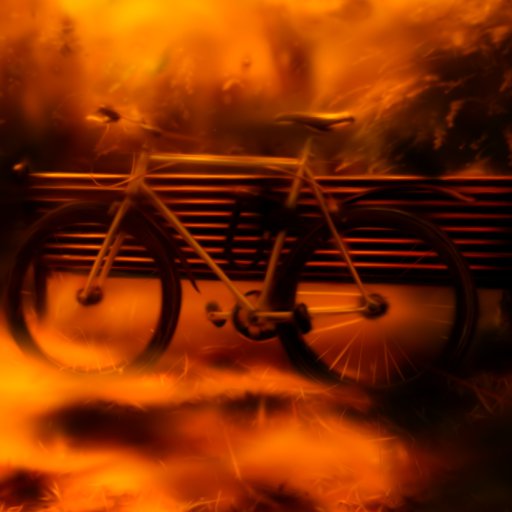} \\

\includegraphics[width=0.16\textwidth]{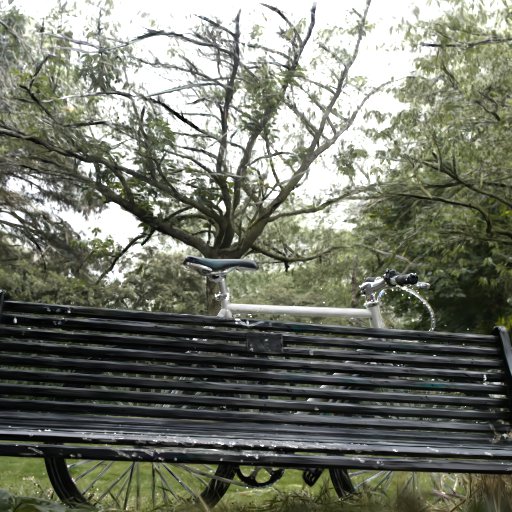} &
\includegraphics[width=0.16\textwidth]{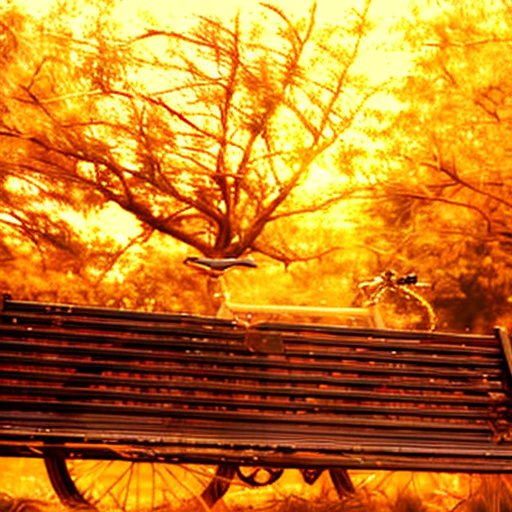} &
\includegraphics[width=0.16\textwidth]{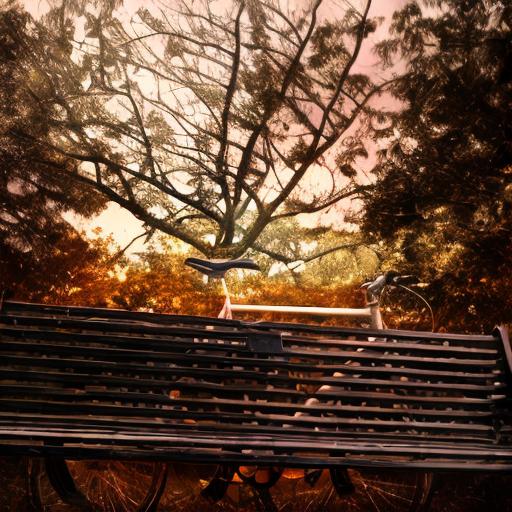} &
\includegraphics[width=0.16\textwidth]{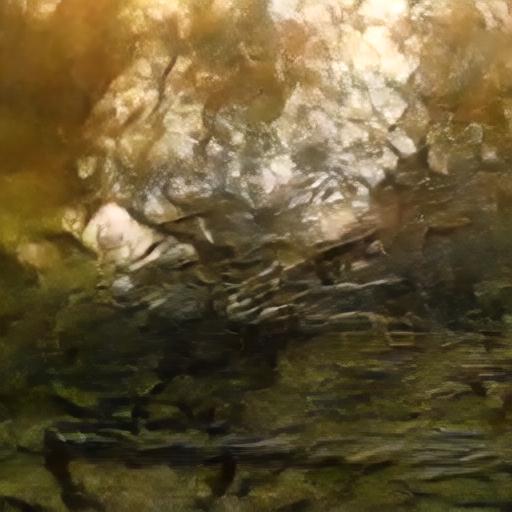} &
\includegraphics[width=0.16\textwidth]{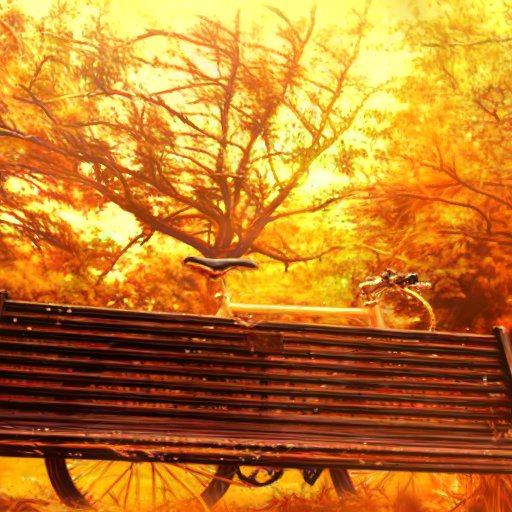} &
\includegraphics[width=0.16\textwidth]{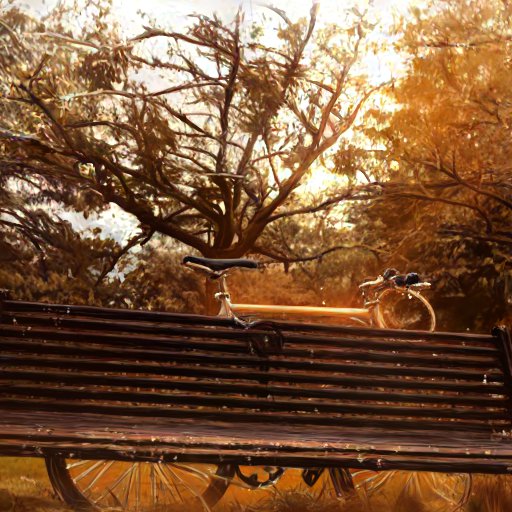} &
\includegraphics[width=0.16\textwidth]{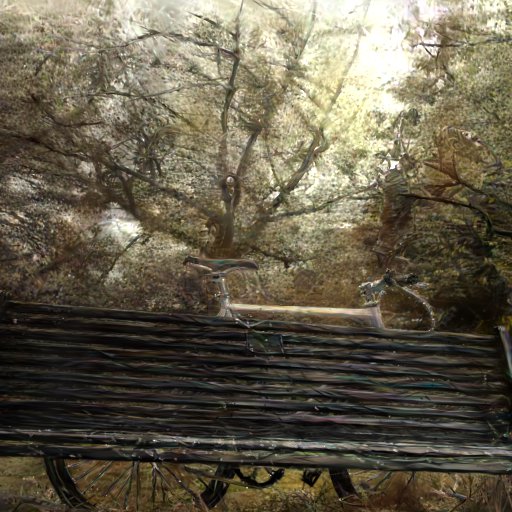} & \includegraphics[width=0.16\textwidth]{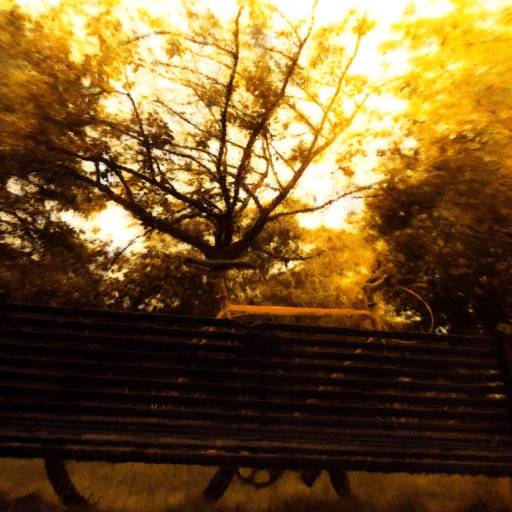} &
\includegraphics[width=0.16\textwidth]{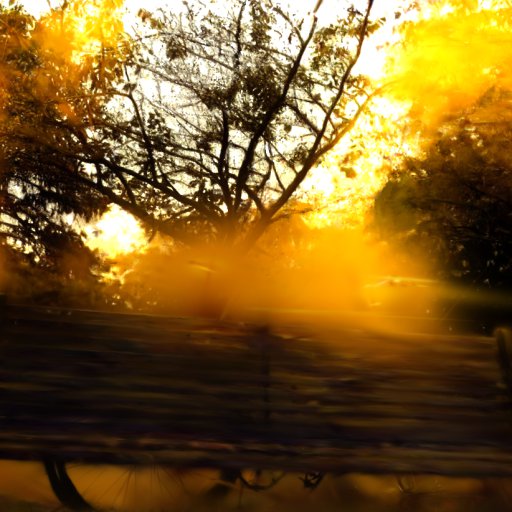} &
\includegraphics[width=0.16\textwidth]{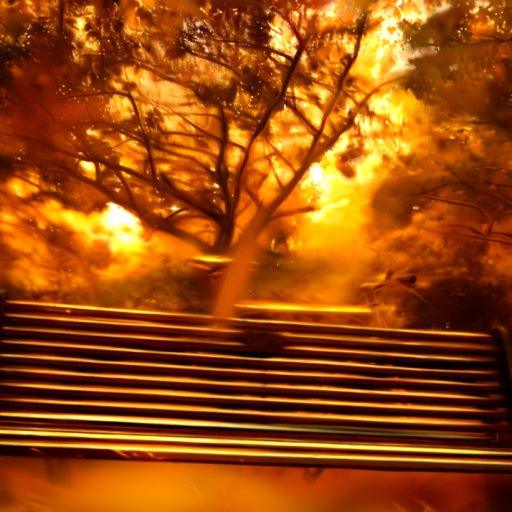} \\

\multicolumn{10}{c}{"\textit{bicycle, sunset, orange glow, long shadows}"} \\

\hline
\end{tabular}
}
\caption{Qualitative comparison between our \textbf{GS-Light} and other prior work on relighting videos or gaussian splatting scenes. Our GS-Light produces results with higher fidelity and aesthetic, improved multi-view consistency, stronger semantic relevance, and enhanced controllability of lighting direction through text prompts.}
\label{fig:banner}
\end{figure}


\section{Introduction}
\label{sec:intro}

Relighting a 3D scene ― changing its illumination while preserving its geometry, material attributes, and view coherence ― is a central problem in computer graphics and vision, with applications spanning augmented/virtual reality, film production, and content creation. Modern neural scene representation methods such as Neural Radiance Fields (NeRF~\cite{nerf}) and 3D Gaussian Splatting (3DGS~\cite{3dgs}) have enabled impressive novel view synthesis, and separately, generative 2D image editing / relighting models (especially diffusion-based ones, e.g. IC-Light~\cite{iclight}) have delivered strong image-level lighting edits. However, combining these two worlds to achieve \textit{multi-view consistent}, \textit{textually controllable}, and \textit{position-aware} relighting remains nontrivial. Key challenges include inferring plausible scene lighting from text, ensuring consistency over views and efficiently dealing with geometry/light interaction (shadows, normals, occlusion).

With the advent of IC-Light, its powerful 2D image relighting capability — covering illumination harmonization, identity preservation, and lighting artistry — has been rapidly adopted in a wide range of AIGC-related downstream applications. However, since ICLight is a model fine-tuned from Stable Diffusion~\cite{ldm} v1.5, it inherently introduces 3D inconsistency in multi-view editing tasks, posing challenges to lifting its prior knowledge to 3D scene editing. In addition, our experiments reveal that current SD-based models and multimodal models are generally insensitive to positional information in textual prompts (e.g., \textit{left}, \textit{right}), which further hinders the alignment between relighting results and user intent. Although some visual grounding works (e.g., VPP-LLaVA~\cite{vpp_llava}) strengthen training in this aspect and achieve promising results, experiments show that they mainly learn patterns between key location-related texts and the ground truth in the limited training data, rather than developing a semantic understanding of abstract spatial information. This is reflected in their poor generalization performance when tested on out-of-domain data.

To address the aforementioned challenges, we propose \textbf{GS-Light}, a light-weight iterative Gaussian Splatting optimization pipeline that bridges text-guided 2D relighting diffusion models with 3D scene representations to achieve \textbf{position-aware}, \textbf{multi-view consistent relighting}. Our framework is built upon two key components. The first is the \textbf{Position-Align Module (PAM)}, which aligns positional semantic information between the inputs—comprising scene-rendered images and user editing instructions—and the relighting model, ensuring accurate interpretation of spatial cues. The second is \textbf{MV-ICLight}, a multi-view relighting model based on IC-Light, specifically designed to enforce cross-view consistency, enabling coherent and faithful relighting across different viewpoints. 

\textbf{PAM} is primarily responsible for the preprocessing of input data. It extracts information from both image and text modalities and further estimates the scene’s geometry and lighting, generating the position-aligned illumination map, which is required by the subsequent module. Given a user prompt, we employ an LVLM (e.g., Qwen2.5-vl~\cite{qwen25vl}) to extract lighting priors such as direction, color temperature or hue, intensity, and optionally reference objects from the given rendered images and input instructions, using a constrained question–answer template to ensure outputs are limited to relighting-relevant parameters. For each view, we further estimate scene geometry and semantics via off-the-shelf predictors, including depth (e.g., VGGT~\cite{vggt}), surface normals (e.g., StableNormal~\cite{stablenormal}), and semantic masks (e.g., LangSAM~\cite{sam2}), which provide cues for shading, occlusion, shadow casting, and object identification. These priors are then fused with geometry constraints to compute per-pixel illumination maps, using a simple Phong diffuse illumination model~\cite{phong}, which are further initialized as the init latents of the diffusion model denoising process, enabling fine-grained control over lighting in the editing results. 

\textbf{MV-ICLight} is a variant implementation of IC-Light’s cross-view attention mechanism, enabling a single-image editing model to perform edits consistently across multiple views. Inspired by DGE~\cite{dge}, we introduce an improved epipolar constraint to realize a memory-efficient, training-free extension of the relighting model from single-view input to multi-view input. Conditioned on the illumination maps and aligned latents, we then generate per-view relit images that maintain illumination consistency across views. Finally, these relit images are integrated into the 3D Gaussian Splatting representation, producing relit scenes that remain faithful to user instructions while ensuring multi-view coherence.

GS-Light is designed to achieve high-quality relighting of 3DGS scenes, faithfully adhering to the user’s editing intent. In tests on both indoor and outdoor datasets, our method demonstrates clear competitiveness against various state-of-the-art approaches in terms of reconstruction consistency, semantic editing similarity, and user subjective evaluation. Our contributions are:

\begin{itemize}
\item We propose GS-Light, the first efficient method that supports textual, position-aware relighting over Gaussian Splatting scenes with multi-view consistency. It takes around 3 minutes to generate per-scene priors once and around 3 minutes for each scene's relighting.
\item We develop a lighting prior extraction scheme via questioning an LVLM, combined with geometry \& semantic estimators, to produce view-wise illumination and latent initialization.
\item We enforce viewwise consistency using advanced epipolar constraints, ensuring coherence in relit views and latent space. It worth noting that our extension framework is compatible with other diffusion models (e.g., DiT~\cite{dit}), which makes sure the great scalability of our method.
\item We validate GS-Light across a variety of scenes (indoor and outdoor), demonstrating superior performance vs baselines in both objective and perceptual metrics, while maintaining fast inference.
\end{itemize}

\section{Related Work}
\label{sec:related}

\subsection{2D Illumination Harmonization / Relighting}

Image relighting methods based on deep neural networks first became mainstream. To enable image-based relighting, Light Stage~\cite{lightstage} was introduced to capture the reflectance functions of human faces, while ~\cite{deep_iamge_based_relight} significantly reduced the number of required input images. SfSNet~\cite{sfsnet} leveraged deep neural networks to model 3D faces and decompose material properties for portrait relighting. ~\cite{portrait_transfer}, using a mass transport approach, achieved portrait illumination transfer. ~\cite{single_portrait_relight} constructed training pairs using One-Light-At-a-Time (OLAT) scans, and ~\cite{physics_guided_relight} further divided the process into diffuse rendering and non-diffuse residual stages. Going a step further, SwitchLight~\cite{switchlight} decomposed source images into intrinsic components, and predicted them with separate networks.

With the development of diffusion models~\cite{dm, ldm, ddim, dit}, fine-tuning pretrained diffusion models for image-to-image tasks such as editing~\cite{diffusionclip, instructpix2pix, null_text}, style transfer~\cite{style_align, deadiff, instantstyle, instantstyle_plus}, geometry estimation~\cite{stablenormal, diffusion_depth, instant3d}, and relighting has proven to be an effective approach. Relightful Harmonization~\cite{relight_harmoni} used the background image as a condition to generate harmonized foreground illumination. IC-Light~\cite{iclight}, by imposing the principle of consistent light transport and leveraging a carefully curated large-scale dataset, fine-tuned Stable Diffusion~\cite{ldm} v1.5 to achieve state-of-the-art performance in 2D image relighting.

\subsection{3DGS Editing / Stylization Models}

One category of methods directly learns features in 3D space and constrains the loss of the target task through specific regularization terms. ~\cite{reference_based} employs a texture-guided control mechanism to directly constrain the parameters of Gaussians. ARF~\cite{arf} proposes the NNFM loss, which provides better 3D consistency compared with the widely used Gram-matrix-based loss~\cite{style_transfer} in style transfer tasks. Building on this, G-Style~\cite{gstyle} introduces a CLIP similarity term and a total variation term, while StyleGaussian~\cite{stylegaussian} embeds VGG features into a radiance field.

Another category of methods leverages the strong priors of 2D editing models to assist in 3D scene editing, mainly addressing inconsistencies in multi-view editing. Instruct-NeRF2NeRF~\cite{in2n} first introduces the Dataset Update strategy to ensure convergence during scene optimization. Based on this, ViCA-NeRF~\cite{vicanerf} adopts multi-view image and depth warping \& mixup methods to provide 2D models with richer 3D information. ConsistDreamer~\cite{consistdreamer} proposes using 3D-consistent structured noise in 2D diffusion denoising, which yields better multi-view consistency. InstantStyleGaussian~\cite{instantstylegaussian} transfers the Instruct-NeRF2NeRF~\cite{in2n} approach to the style transfer task. ProEdit~\cite{proedit} employs a progressive editing strategy to reduce the feasible space of text-aligned editing tasks, thereby mitigating multi-view inconsistency. DGE~\cite{dge} directly modifies the 2D model’s self-attention into cross-view self-attention across keyframes and further applies epipolar-constrained matching to propagate attention results into non-keyframe feature maps, significantly reducing memory demand during inference. EditSplat~\cite{editsplat} introduces multi-view fusion guidance, which aligns multi-view information with text prompts and source images to ensure multi-view consistency. By collecting a large scale illumination video dataset, RelightVid~\cite{relightvid} finetuned IC-Light to adapted to video relighting task, while Lumen~\cite{lumen} training an end to end DiT-based model. 

Our work builds on the improved epipolar-constrained matching scheme of DGE~\cite{dge} to achieve more consistent editing results under reduced memory usage. To further exploit the powerful relighting capability of IC-Light, we adapt IC-Light into the DGE framework without modifying or fine-tuning its weights, enabling a high-quality, multi-view consistent relighting module across multiple images, which we call \textbf{MV-ICLight}.

These methods work well on the multiview images or videos editing or relighting though, however, they can hardly tackle the challenge about the unfaithful result related to the positional concept (e.g., \textit{light from right}) input by users.

\subsection{3DGS with Inverse Rendering / Physically Based Rendering}

GI-GS~\cite{gigs}, GUS-IR~\cite{gusir} take a set of pretrained 3D Gaussians with normals and intrisics, then perform a differentiable PBR to model the direct light and global illumination. GS3~\cite{gs3} presents spatial and angular Gaussians with intrinsics and lighting/view directions, using mlps to predict shadow and global illumination. RNG~\cite{rng} add an extra latent vector that describes the reflectance for each Gaussian to model the surface with soft boundaries like fur or fabric. GeoSplatting~\cite{geosplating} proposes MGadapter to differentiably construct a surface-grounded 3DGS from an optimizable mesh guidance, enabling a precise light transport calculation.

These physically based rendering methods can achieve highly realistic relighting effects; however, their computational cost — including both training and rendering — is often very expensive. Moreover, these methods require prior knowledge of the light source parameters, such as position, color, and intensity. In our relighting task, however, such information must be inferred from the user’s textual instructions, which is typically beyond the capability of this class of methods.
\subsection{Positional Alignment between Images and Text Prompts}

LISA~\cite{lisa} and GLaMM~\cite{glamm} employ the SAM decoder to achieve segmentation at the pixel level on reference images, whereas LLaVA Grounding~\cite{llava_grounding} extends the model with a dedicated grounding head to output bounding boxes. VPP-LLaVA~\cite{vpp_llava} enhances MLLMs’ visual grounding by introducing visual position prompts and a curated 0.6M-sample dataset, achieving state-of-the-art localization performance with strong zero-shot generalization. WhatsUp~\cite{whatsup} shows that despite strong performance on VQAv2~\cite{vqav2}, VL models struggle with basic spatial relations, and they introduce new benchmarks to highlight this limitation.

\section{Method}
\label{sec:method}

\begin{figure}[t]  
    \centering
    \includegraphics[width=0.9\linewidth]{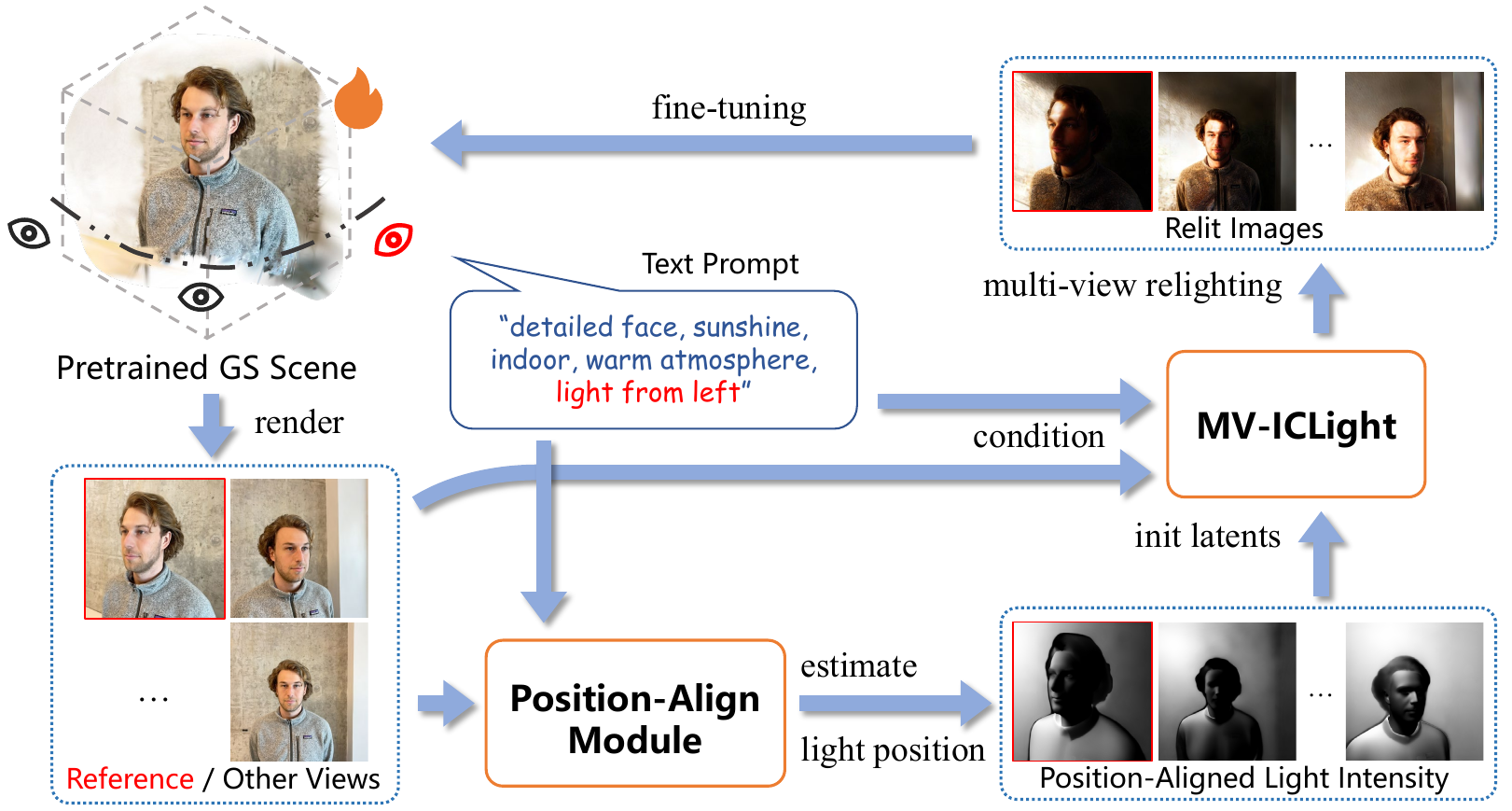}  
    \caption{Circulation pipeline of GS-Light. Starting from a pre-trained Gaussian Splatting (GS) scene and a text prompt specifying the relighting instruction, we first render images from all training views. One training view is selected as the reference view to align the positional information in the prompt. Through our proposed \textbf{Position-Align Module} (PAM), we generate position-aligned light intensity maps for all views. These intensity maps are then provided as initialization latents to our \textbf{Multi-View ICLight}, producing multi-view consistent relit images. Finally, the relit images are used to fine-tune the opacity and color parameters of the GS scene, forming a closed-loop tuning circulation. Repeating this circulation multiple times ensures that the relit GS converges to a stable and consistent result.}
    \label{fig:pipeline}  
\end{figure}

In this section, we present the details of \textbf{GS-Light}, our proposed pipeline for text-driven relighting of 3D Gaussian Splatting (3DGS) scenes. Our goal is to design a method that (i) faithfully adheres to user instructions expressed in natural language, especially the light direction, (ii) ensures multi-view consistency, and (iii) operates efficiently without requiring per-scene retraining. To achieve this, GS-Light integrates a \textbf{Position-Align Module (PAM)} for prompt–geometry alignment with an extended \textbf{MV-ICLight} for training-free multi-view diffusion-based relighting. An overview of the pipeline is shown in Fig.~\ref{fig:pipeline}.

\subsection{Preliminaries}

\subsubsection{Diffusion Editors}
Diffusion-based image editing models have achieved state-of-the-art performance in tasks such as inpainting, illumination harmonization, and relighting. A diffusion model~\cite{dm,ldm} learns the data distribution $p_{\text{data}}(x)$ by reversing a gradual noising process. Specifically, the forward process adds Gaussian noise to a clean image $x_0$:
\begin{equation}
q(x_t \mid x_0) = \mathcal{N}(x_t; \sqrt{\bar{\alpha}_t}x_0, (1-\bar{\alpha}t)I),
\end{equation}
while the reverse process uses a neural network $\epsilon\theta(x_t,t,c)$ to predict and remove noise under condition $c$, thereby generating or editing images consistent with the guidance.

\paragraph{Editing with Diffusion}
Diffusion-based editing methods can be broadly divided into two categories. On the one hand, training-free approaches reuse a pretrained diffusion model and perform modifications by initializing from a noised version of an existing image $x_0$, then denoising under a new condition $c_{\text{edit}}$:
\begin{equation}
x_0' = \text{Denoise}(x_t, c_{\text{edit}}), \quad x_t \sim q(x_t \mid x_0),
\end{equation}
which enables preserving the structure of $x_0$ while applying edits guided by $c_{\text{edit}}$. On the other hand, training-based approaches rely on finetuning or retraining with triplet data $(x_{\text{src}}, c_{\text{edit}}, x_{\text{tgt}})$ to directly learn the conditional distribution $p\left(x_{\text{tgt}}|x_{\text{src}},  c_{\text{edit}}\right)$ for editing. In this work, we adopt the \textbf{IC-Light} model, adhered to the latter paradigm, which is specifically designed for illumination control and artistic lighting editing, and finetuned on a large-scale and carefully curated dataset on the base of Stable Diffusion v1.5 model. IC-Light introduces dedicated conditioning channels for illumination harmonization, enabling edits that respect global lighting style while preserving content fidelity. However, being trained purely in the 2D image domain, IC-Light lacks mechanisms to enforce cross-view consistency, which limits its direct applicability to 3D scene editing.

\subsubsection{3D Gaussian Splatting}

A 3D Gaussian distribution defines a probability density in 3D:

\begin{equation}
G(\boldsymbol{x}) = e^{-\tfrac{1}{2}(\boldsymbol{x}-\boldsymbol{\mu})^\top \boldsymbol{\Sigma}^{-1}(\boldsymbol{x}-\boldsymbol{\mu})}
\end{equation}

where $\boldsymbol{\mu} \in \mathbb{R}^3, \boldsymbol{\Sigma} \in \mathbb{R}^{3\times 3}$ define the mean and covariance matrix.

3D Gaussian Splatting (3DGS)~\cite{3dgs} represents a scene with a set of anisotropic Gaussian primitives $\mathcal{G} = {\{ (\boldsymbol{\mu}_i , \Sigma_i, \boldsymbol{c}_i, \alpha_i) \}}_{i=1}^G$ where $\boldsymbol{\mu}_i, \Sigma_i$ determine each Gaussian primitive's distribution, with associated attributes color $\boldsymbol{c}_i \in [0,1]^3$ and opacity $\alpha_i \in [0,1]$. During rendering, Gaussians are projected onto the image plane and rasterized in tiles for parallel efficiency. For a pixel $\boldsymbol{p}$, which is receiving contributions from a sorted set of Gaussians ${G_{i}}$ along the view direction, its color $\boldsymbol{C}$ is obtained via differentiable $\alpha$-blending:
\begin{equation}
\boldsymbol{C}(\boldsymbol{p}) = \sum_{i=1}^{G} {\alpha_i T_i \boldsymbol{c}_i}
\end{equation}
where $T_i = \prod_{j < i}(1-\alpha_j)$ is the accumulated transmittance from front to back, $G$ is the total number of Gaussians.

The entire process is fully differentiable. GS initializes primitives through the Structure from Motion~\cite{sfm} (SfM) process and applies densification and pruning during training based on gradient magnitudes in NDC coordinates to control the number of Gaussians, making training and finetuning both efficient and simple. GS-Light finetunes GS using relit images by freezing the position and shape of each Gaussian and disabling the densification strategy, which allows rapid optimization of scene appearance while avoiding memory overhead issues.

\subsection{Position-Align Module (PAM)}

\begin{figure}[ht]
\centering

\begin{subfigure}[t]{0.23\textwidth}
  \centering
  \includegraphics[width=\linewidth]{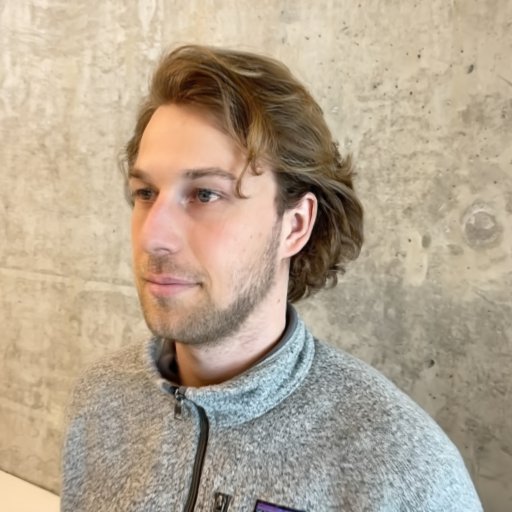}
  \caption*{Input}
\end{subfigure}
\begin{subfigure}[t]{0.23\textwidth}
  \centering
  \includegraphics[width=\linewidth]{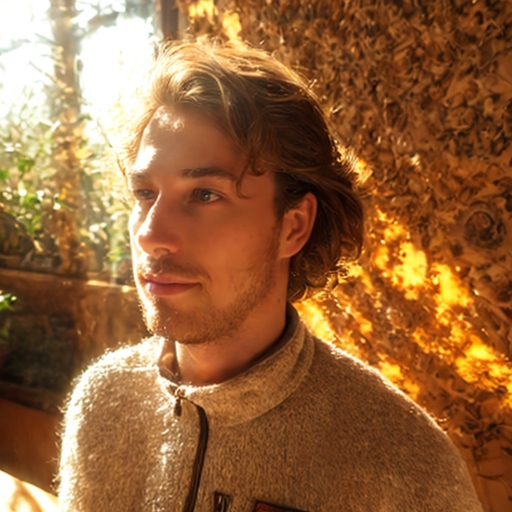}
  \caption*{
    \begin{minipage}{0.8\linewidth}
    \raggedright
    \textit{"detailed face, sunshine, indoor, warm atmosphere"}
    \end{minipage}
    }
\end{subfigure}
\begin{subfigure}[t]{0.23\textwidth}
  \centering
  \includegraphics[width=\linewidth]{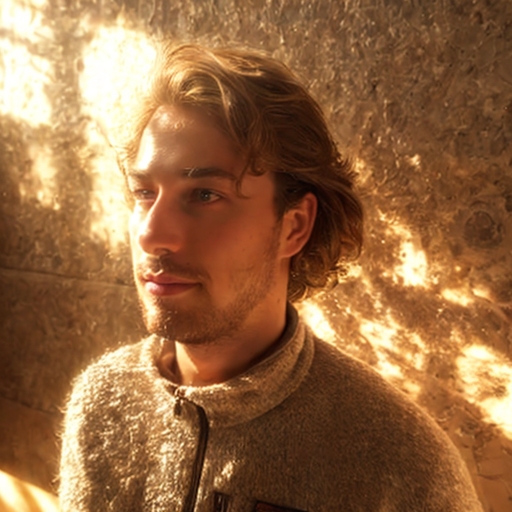}
  \caption*{
    \begin{minipage}{\linewidth}
    \raggedright
    \textit{"..., warm atmosphere, \textbf{light from left}"}
    \end{minipage}
    }
\end{subfigure}
\begin{subfigure}[t]{0.23\textwidth}
  \centering
  \includegraphics[width=\linewidth]{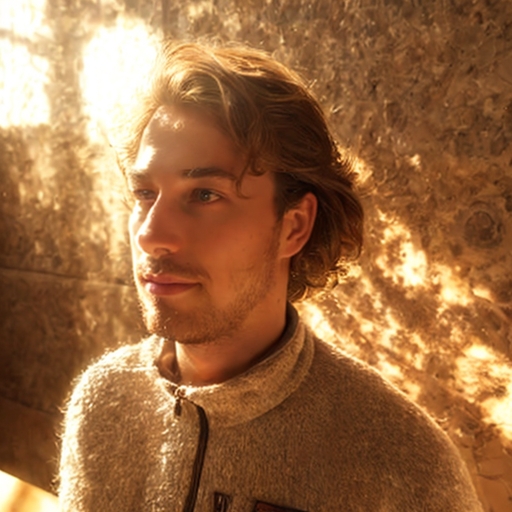}
  \caption*{
    \begin{minipage}{\linewidth}
    \raggedright
    \textit{"..., warm atmosphere, \textbf{light from right}"}
    \end{minipage}
    }
\end{subfigure}

\caption{IC-Light relighting results on different light direction instructions, which show a \textbf{weak} or \textbf{wrong response} towards the position-related information.}
\label{fig:iclight_position}

\end{figure}

WhatsUp~\cite{whatsup} points out that existing multimodal large models have a relatively weak semantic understanding of positional relationships in the textual modality. Experiments show that the same phenomenon also occurs in text-to-image models, as illustrated in Fig.~\ref{fig:iclight_position}. The Position-Align Module is responsible for bridging the semantic gap between text prompts and geometric cues, producing position-aware illumination maps that are later used to initialize the diffusion process, which is shown in Fig.~\ref{fig:pam}. PAM consists of 2 stages:

\begin{figure}[!t]  
    \centering
    \includegraphics[width=0.9\linewidth]{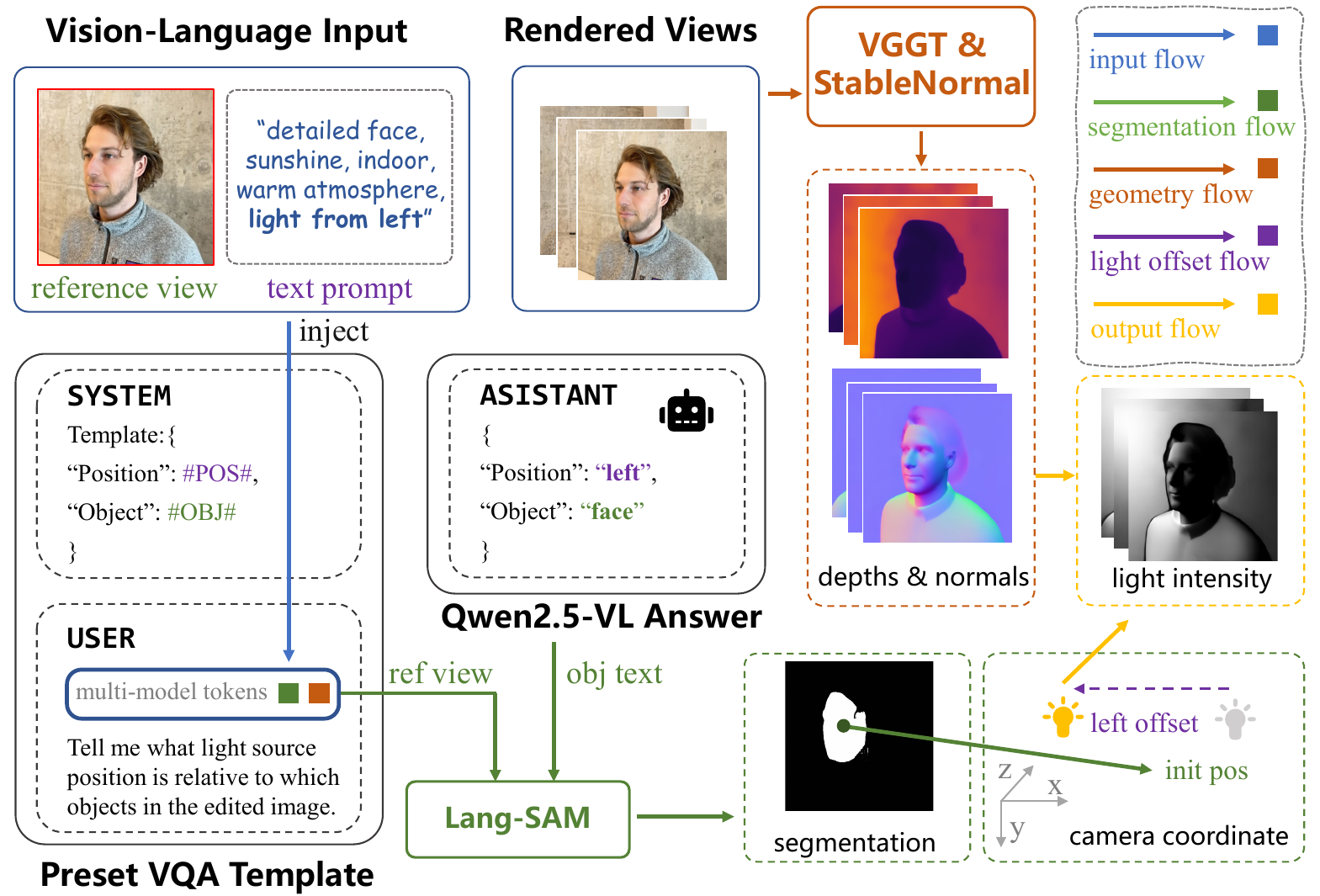}  
    \caption{Details of \textbf{PAM}. Given the rendered views and a text prompt, Qwen2.5-VL is employed with a preset VQA template to parse the user’s intended lighting direction and reference object. Pretrained models VGGT, StableNormal, and Lang-SAM are then applied to estimate the initial light position and scene geometry. By combining these estimates with the parsed light-position offset, PAM produces light-intensity maps that are spatially aligned with the input positional intent across all views.}
    \label{fig:pam}  
\end{figure}

\subsubsection{Parsing Lighting Priors via LVLM}
\label{subsubsec:PAM}
Given a user instruction, we employ a large vision-language model (LVLM) such as Qwen2.5-vl~\cite{qwen25vl} to extract structured lighting priors. The priors include lighting direction (e.g., \textit{light from the left}) and reference objects (e.g., \textit{detailed face}). To ensure robustness and avoid irrelevant outputs, we design a constrained question–answer template that restricts the LVLM’s output space to lighting-relevant descriptors only. This provides reliable, semantically grounded cues for subsequent relighting.

Given a 3DGS scene $\mathcal{G}$, a set of training input views $\mathcal{V}={\{v_n\}}_{n=1}^{N_v}$, where $N_v$ is the number of training views, and the user-specified relighting instruction $c_\text{edit}$, we begin by rendering the GS scene into each view to obtain multi-view images $\mathcal{I} = {\{I_n\}}_{n=1}^{N_v}$ that contain scene information. We assume that the illumination conditions and lighting direction described by the user are referenced from a specific view $v_\text{ref}$. Without loss of generality, we consider $v_\text{ref} \in \mathcal{V}$. 

To extract the user’s intended lighting direction and the corresponding reference object, we employ a predefined Q\&A template $\mathcal{T}$ that prompts the LVLM to respond in a fixed format:
\textit{Light is on the \{DIRECTION\} of the \{OBJECT\}}. With the fixed format, we can easily parse the direction and object prompts with a simple regular expression matching.

\begin{equation}
p_\text{dir}, p_\text{obj} = \text{QwenVL}(I_\text{ref}, c_\text{eidit}, \mathcal{T})
\end{equation}

Here, QwenVL refers to Qwen2.5-VL~\cite{qwen25vl}, a powerful open-source LVLM. $p_\text{dir} \in \{\text{left}, \text{right}, \text{top}, \text{bottom}\}$ is direction prompt and $p_\text{obj}$ is the reference object prompt. By enumerating the possible cases of $p_\text{dir}$, we can determine the light source offset direction that matches the user’s intent.

Next, we also need to perform text–image alignment based on the reference object, in order to determine the initial position of the light source. Using a pretrained segmentation foundation model, we can obtain the mask of the reference object $\boldsymbol{M}_\text{obj}$:

\begin{equation}
\boldsymbol{M}_\text{obj} = \text{LangSAM}(I_\text{ref}, p_\text{dir})
\end{equation}

Here, LangSAM is a text-guided segmentation model that integrates GroundingDINO~\cite{groundingdino} and SAM2.1~\cite{sam2}, and outputs high-quality object masks based on the given prompt. With this mask, we can compute the pixel coordinates of the reference object in the image $\boldsymbol{p}_\text{obj}$:

\begin{equation}
\boldsymbol{p}_\text{obj} = \frac{1}{|\boldsymbol{M}_\text{obj}|} \sum_{\boldsymbol{p} \in \boldsymbol{M}_\text{obj}} {\boldsymbol{p}}
\end{equation}

which will be used for subsequent estimation of the light source position.

\subsubsection{Geometric and Semantic Understanding \& Latent Initialization}

\paragraph{Geometry Estimation}
To complement text-based priors, we estimate per-view scene geometry and semantics using off-the-shelf models. Scale-aligned depth maps $\mathcal{D} = {\{D_n\}}_{n=1}^{N_v}$ are obtained via VGGT~\cite{vggt}, which is a general framework designed for multi-view geometry perception with an end-to-end transformer-based architecture. Then surface normals $\mathcal{N} = {\{N_n\}}_{n=1}^{N_v}$ are estimated via StableNormal~\cite{stablenormal}, which predicts more smooth and proper normals on the base of a powerful generative diffusion prior model, rather than directly inferring from depth maps $\mathcal{D}$. 

\paragraph{Semantic Understanding}

\begin{figure}[!t]  
    \centering
    \includegraphics[width=0.9\linewidth]{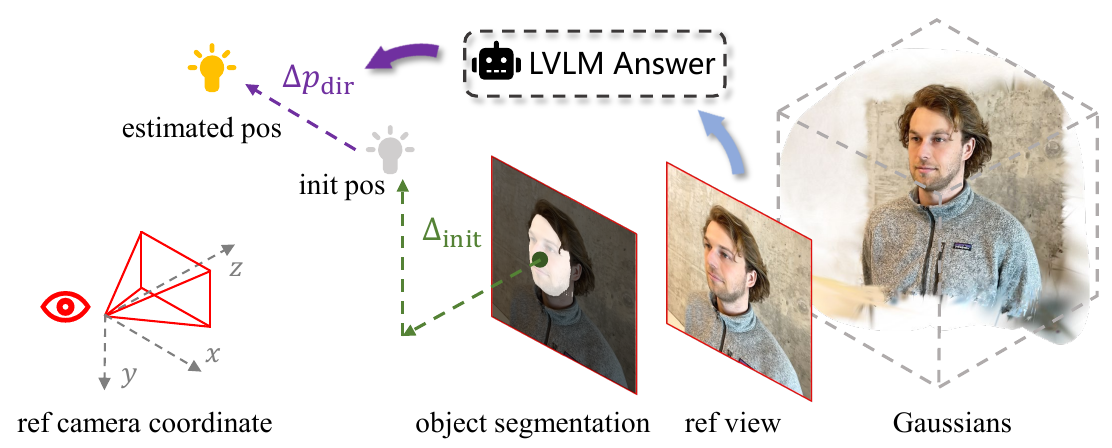}  
    \caption{The process of estimating light source position.}
    \label{fig:light_pos}  
\end{figure}

As mentioned earlier, we can use LangSAM to align the reference image $I_\text{ref}$ with the user’s instructions $c_\text{eidit}$, thereby estimating the reference object's pixel coordinates  $\boldsymbol{p}_\text{obj}$. Considering that lighting usually comes from above, we assume here that the initial light source position is located diagonally above on the side of the reference object facing the camera (the exact initial position is not critical; what matters is the change in the light source position before and after relighting). Following the convention of 3DGS, we adopt the OpenCV camera coordinate system (with the x, y, z axes of the camera pointing to the right, downward, and into the screen, respectively):

\begin{equation}
\boldsymbol{p}_l = \begin{bmatrix}
\boldsymbol{p}_{\text{obj}} \\
d_\text{ref}
\end{bmatrix} + \boldsymbol{\Delta}_{\text{init}}
\end{equation}
where $\boldsymbol{p}_l$ is the light position in camera coordinate, $d_\text{ref}$ is the depth at pixel coordinate $\boldsymbol{p}_{\text{obj}}$. $\boldsymbol{\Delta}_{\text{init}}$ is the relative offset to initialize the light position.

Based on the direction prompt $p_\text{dir}$ provided by the LVLM, we can add a corresponding offset $\boldsymbol{\Delta} p_\text{dir}$ to the initial light source position so that it aligns with the user’s description of the light source direction:

\begin{equation}
\boldsymbol{p}'_l = \boldsymbol{p}_l + \boldsymbol{\Delta} p_\text{dir}
\end{equation}

To better understand the light position estimation process, we illustrate it as Fig.~\ref{fig:light_pos}.

\paragraph{Latent Initialization}
With geometry and light source ready, we can the compute the distribution of light intensity across the views by a modified
Phong-like diffuse illuminating model~\cite{phong}:

\begin{equation}
I_d = \max(-\langle \boldsymbol{l}_\text{in}, \boldsymbol{n} \rangle, 0)^\gamma
\end{equation}

where $I_d$ is the diffuse part reflected by the surface, which indicates the light intensity. $\boldsymbol{l}_\text{in}$ denotes incident direction and  $\boldsymbol{n}$ denotes surface normal, $\langle \cdot,\cdot \rangle$ is inner product operation, $\gamma$ is a hyperparameter to balance the brightness distribution.

The illumination intensity maps $\{{I_d}\}_{n=1}^{N_v}$ serve as initialization signals in the diffusion process. Specifically, we encode the illumination maps into latent space and inject them into IC-Light’s denoising steps as init latents $l_d$ 

\begin{equation}
\label{equation:vae}
l_d = \operatorname{VAE}\left( I_d \right)
\end{equation}
where $\operatorname{VAE}$ is the image encoder of SD-1.5. This provides explicit geometry-aware lighting conditioning, allowing the model to align its generation trajectory with the desired lighting direction and intensity, significantly reducing prompt–output mismatch.

\subsection{MV-ICLight}

Although IC-Light is renowned for its powerful image relighting capability, like other 2D image editing models, it also struggles to ensure consistency in the outputs when given multi-view inputs. (Insert Image). To extend IC-Light from single-view to multi-view editing, we introduce \textbf{MV-ICLight}, which enforces cross-view coherence during relighting.

\begin{figure}[!t]  
    \centering
    \includegraphics[width=0.9\linewidth]{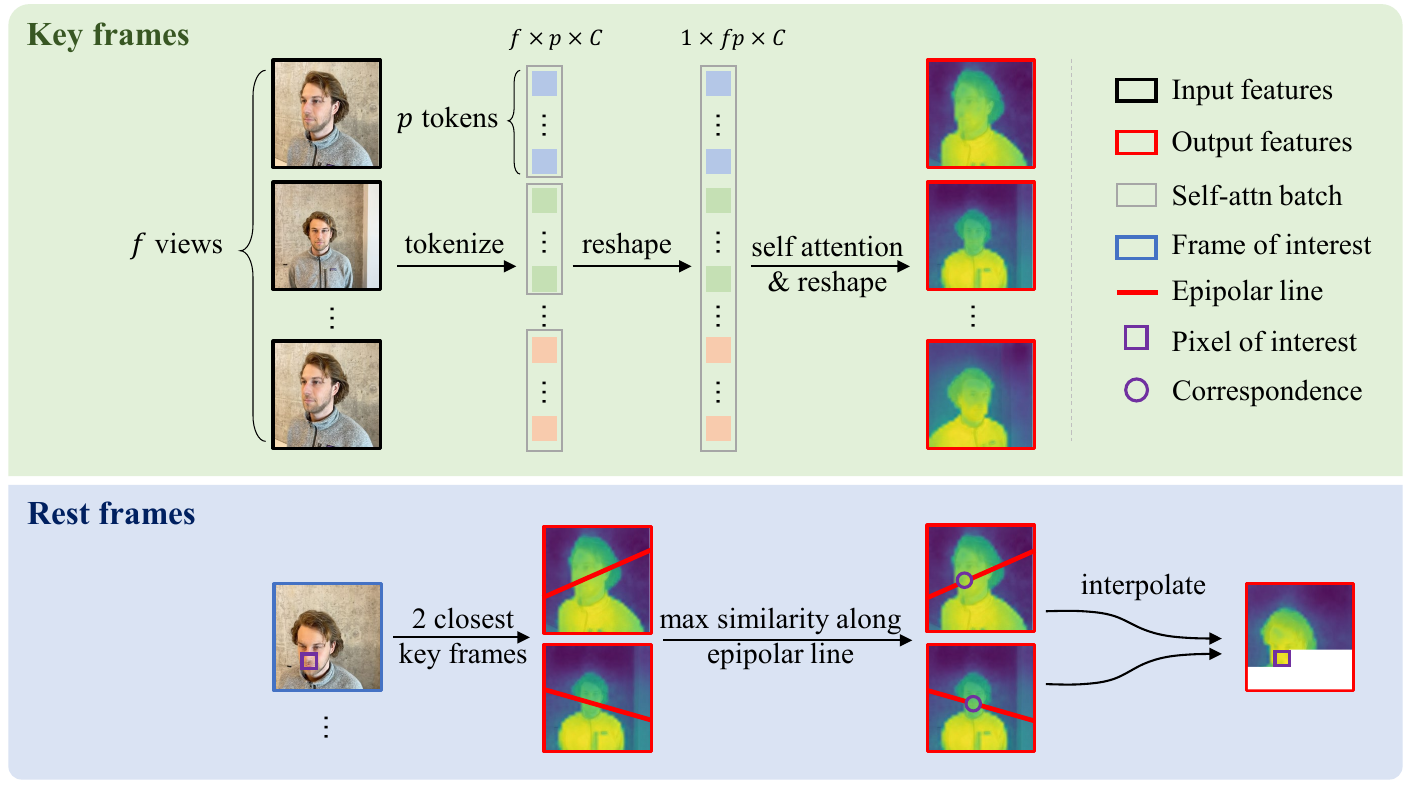}  
    \caption{Schematic diagram of self attention in MV-ICLight. Upper part: the implementation mechanism of Cross-View Attention. Lower part: epipolar constrain for non-key frames from DGE~\cite{dge}.}
    \label{fig:dge}  
\end{figure}

\subsubsection{Cross-View Attention}
To enable multi-view editing, it is necessary for the model to aggregate global information from multi-view inputs during inference, thereby establishing a mechanism for inter-view information exchange and supplementation. A natural idea is to replace the self-attention in the Diffusion U-Net with cross-view attention, as shown in the upper part of Fig.~\ref{fig:dge}. Specifically, for a given set of query and key tokens $\{Q\}_{f=1}^{F}, \{K\}_{f=1}^{F}$ from the $f$-th frame, each of $Q_f$ can perceive all tokens from other frames, instead of only the same frame which is the case of traditional self-attention. The Cross-View Attention Score can be formulated as: 

\begin{equation}
\operatorname{CVAttn}\left( Q, K, f \right) = \operatorname{Softmax}\left( \frac{Q_f \cdot \left[ K_1, \cdots, K_F\right]}{\sqrt{d}} \right)
\end{equation}
where $d$ is the dimension of the query and key embeddings.

\subsubsection{Multi-View Relighting with Advanced Epipolar Constraints}
Directly extending self-attention across multiple frames will significantly increase computational and memory overhead, since the cost of self-attention grows quadratically with the number of input tokens. Inherited from DGE~\cite{dge}, we subsample a few key frames from all training views to perform a full-size inference and implement an epipolar-matching mechanism to fill up the feature maps of the rest frames, as shown in the lower part of Fig.~\ref{fig:dge}. We keep the notations from Sec.~\ref{subsubsec:PAM} and DGE. Given a set of key frames $\mathcal{K} \subset \mathcal{V}$, where $\mathcal{V}$ is the training set, a feature map $\Psi_{v'}$ corresponding to image $I_{v'}$, where $v' \notin \mathcal{K}$, and features $\Psi_{k^*}$ of the nearest keyframe $I_{k^*}$, the correspondence map $M_{v'}$ is given by:

\begin{equation}
M_{v'}\left[ \boldsymbol{p}_u \right] = \operatorname*{argmin}_{\boldsymbol{p}_v,\boldsymbol{p}_v^{T}\hat{F}\boldsymbol{p}_u=0} D\left( \Psi_{v'}\left[\boldsymbol{p}_u\right], \Psi_{k^*}\left[\boldsymbol{p}_v\right] \right)
\end{equation}

where $D$ is the cosine distance, $\boldsymbol{p}_u$ and $\boldsymbol{p}_v$ is the pixel-wise index, $k^*$ is the key view closest to view $v'$, and unlike DGE, $\hat{F}=\frac{F}{\|F\|+\epsilon}$ is the normalized fundamental matrix corresponding to the two views $v'$ and $k^*$, where $\epsilon$  is a small constant introduced to avoid division-by-zero overflow. 

Since the fundamental matrix $F$ is defined up to an arbitrary nonzero scale factor, we normalized it to have unit norm to ensure numerical stability during inference. Experiments demonstrate that, as shown in Tab.~\ref{tab:ablation_epipolar}, by using our normalized fundamental matrix $\hat{F}$, the failure of epipolar constraints in the shallow attention blocks of the UNet is largely avoided. As a result, inference stability is greatly improved, and the generated results exhibit significantly better multi-view consistency.

During inference, as shown in Eq.~\ref{equation:vae}, the spatially aligned illumination diffuse map is encoded into latents $l_d$, which are then noised up to timestep $T'(<T_\text{max})$, as the starting point for denoising, where $T_\text{max}$ is the total denoising steps number. The latents representation of the original images $l_I$, is also used as a condition to keep the geometry and details of results consistent with original images, and concatenated with $l_d$ along the channel dimension:

\begin{equation}
l_i^{\frac{H}{8}\times\frac{H}{8}\times 2C} = \operatorname{Concat}(l_d^{\frac{H}{8}\times\frac{H}{8}\times C}, l_I^{\frac{H}{8}\times\frac{H}{8}\times C})
\end{equation}
where $l_i$ is the init noised latents to perform partial DDIM~\cite{ddim} pipeline, $H, W$ is resolution of source images and $C$ is the channel dimension of latents.

\subsubsection{GS Tuning and Iterative Dataset Updating Strategy}

Through MV-ICLight, the relit images across multiple views indeed achieve better consistency compared to independently relighting each view, though slight inconsistencies still remain. To address this, we adopt the \textbf{Iterative Dataset Update} strategy from IN2N~\cite{in2n}: after every $K_\text{int}$ steps, the GS scene is rendered from all viewpoints and relit using MV-ICLight, and the resulting images are directly updated as the training data for the corresponding views. This process is repeated $K_\text{reap}$ times. The GS scene eventually converges to a multi-view consistent relighting result, where $K_\text{int}$ and $K_\text{reap}$ are hyperparameters controlling the update interval and number of iterations during fine-tuning.

To apply the multi-view relighting edits to the GS-represented scene, we directly use the edited results as the training set to fine-tune the Gaussians:

\begin{equation}
\mathcal{G'} =  \operatorname*{argmin}_{c_i, \alpha_i \in \mathcal{G}_i, \forall \mathcal{G}_i \in \mathcal{G}}\sum_{v \in \mathcal{V}} {{\operatorname{L}\left( \operatorname{Render}(\mathcal{G},v), I_\text{relit}^v \right)}}
\end{equation}
where $\mathcal{G'}$ is the final relit GS scene,  $\operatorname{L}$ is the loss function of 3DGS training, typically composed of an L1 loss and an SSIM loss, $\operatorname{Render}(\mathcal{G},v)$ is the Gaussian rendering at viewpoint $v$, and $I_\text{relit}^v$ is the relit image generated from MV-ICLight.

\section{Experiments}
\label{sec:exp}

In this section, we first present the implementation details of our method, followed by both qualitative and quantitative comparisons with existing approaches. We then conduct an ablation study to further analyze the effectiveness of our method.

\paragraph{Datasets}
The relighting tasks are performed on several datasets of indoor and outdoor scenes, including IN2N~\cite{in2n} dataset, MipNerf360~\cite{mip360}, Scannet++~\cite{scannetpp}, etc., where 3D Gaussian Splatting models (or dense multi-view imagery) are available. We use the GPT-5~\cite{gpt5} model to generate various scene-related relighting instructions, and each instruction is iterated over possible lighting directions $p_\text{dir} \in \{\text{left}, \text{right}, \text{top}, \text{bottom}\}$ to construct the benchmark. 
Specifically, we select all scenes from IN2N and MipNeRF360, along with the first 10 scenes from ScanNet++, resulting in a total of 25 scenes. For each scene, we assign three randomly generated relighting prompts and random lighting directions, leading to 75 relighting tasks in total, which constitute our benchmark.
We also conducted a user study to collect participants’ preferences regarding the relighting results produced by different models. 

\paragraph{Implementation Details}
Under our 2D-to-MV framework, the capability of the 2D image editor largely determines the quality of the final 3D editing results. For fairness, we adopt several prior works based on IP2P~\cite{instructpix2pix} for 3D scene editing and IC-Light~\cite{iclight} for scene/video relighting as our baselines. To control the memory consumption during inference, we use 50–70 input images of 512×512 resolution for all datasets. We adopt classifier-free guidance with a text guidance scale of 7.5, applied throughout the generation process. For the Dataset Update Strategy, we set the update interval $K_\text{int}$ to 500 and the number of updates $K_\text{reap}$ to 2, consistent with DGE~\cite{dge}. During Gaussian Splatting fine-tuning, we disable gradient computation for all parameters except color and opacity, and also deactivate the densification strategy.  

\begin{table}[t]
\centering
\caption{
Quantitative comparison on IN2N~\cite{in2n} dataset. Best and second results are highlighted in \textbf{bold} and with \underline{underline}.
}
\label{tab:quantitative_results2}
\setlength{\tabcolsep}{3.8pt}
\resizebox{\textwidth}{!}{
\begin{threeparttable}
\begin{tabular}{lcccccc}
\toprule
\multirow{3}{*}{\textbf{Method}} & 
\multirow{3}{*}{\textbf{CLIP-T}↑} & 
\multirow{3}{*}{\textbf{CLIP-D}↑} &
\multicolumn{4}{c}{\textbf{VBench}} \\
\cline{4-7}
&  &  & \textbf{Subject} & \textbf{Background} & \textbf{Aesthetic} & \textbf{Image} \\
&  &  & \textbf{Consistency}↑ & \textbf{Consistency}↑ & \textbf{Quality}↑ & \textbf{Quality}↑ \\
\midrule
\multicolumn{7}{c}{\textbf{Relighting on Videos}} \\
RelightVid & \underline{0.2383} & \underline{0.0611} & \underline{0.7806} & \underline{0.8624} & \underline{0.5991} & \underline{61.31} \\
Lumen* & 0.1754 & -0.0336 & 0.7215 & 0.8336 & 0.4720 & 53.88 \\
\textbf{Ours} & \textbf{0.2580} & \textbf{0.1170} & \textbf{0.8320} & \textbf{0.8884} & \textbf{0.6317} & \textbf{62.22} \\
\midrule
\multicolumn{7}{c}{\textbf{Relighting on Gaussian Splatting}} \\
DGE & \underline{0.2369} & \underline{0.0918} & 0.8614 & 0.9017 & \underline{0.5619} & \textbf{60.65} \\
EditSplat & 0.1983 & -0.0021 & 0.8730 & 0.9094 & 0.5203 & 53.94 \\
IN2N & 0.2222 & 0.0410 & \textbf{0.8757} & 0.9098 & 0.5425 & 51.08 \\
IGS2GS$^\dagger$ & 0.2055 & 0.0111 & 0.8594 & 0.9268 & 0.4249 & 30.27 \\
IGS2GS-IC$^\ddagger$ & 0.1800 & -0.0102 & 0.8642 & \textbf{0.9330} & 0.3938 & 26.76 \\
\textbf{Ours} & \textbf{0.2580} & \textbf{0.1171} & \underline{0.8748} & \underline{0.9270} & \textbf{0.6043} & \underline{57.80} \\
\bottomrule
\end{tabular}
\begin{tablenotes}
\item[*] the generating resolution setting of benchmark is 512$\times$512, however the Lumen's recommended resolution is 480$\times$832, which may cause a performance decline.
\item[$^\dagger$] denotes Instruct-GS2GS, an improved version adapted from IN2N for GS training, using InstructPix2Pix~\cite{instructpix2pix} as the 2D image editor.
\item[$^\ddagger$] denotes replacing the 2D image editor InstructPix2Pix with IC-Light~\cite{iclight}.
\end{tablenotes}
\end{threeparttable}
}
\end{table}

\begin{table}[t]
\centering
\caption{
Quantitative comparison on MipNerf360~\cite{mip360} and Scannet++~\cite{scannetpp} dataset. Best and second results are highlighted in \textbf{bold} and with \underline{underline}.
}
\label{tab:quantitative_results3}
\setlength{\tabcolsep}{3.8pt}
\resizebox{\textwidth}{!}{
\begin{threeparttable}
\begin{tabular}{lcccccccccccc}
\toprule
\multirow{2}{*}{\textbf{Method}} & 
\multicolumn{6}{c}{\textbf{MipNerf360}} &
\multicolumn{6}{c}{\textbf{Scannet++}} \\

& \textbf{CLIP-T}↑ & \textbf{CLIP-D}↑ & \textbf{S.C.}↑ & \textbf{B.C.}↑ & \textbf{A.Q.}↑ & \textbf{I.Q.}↑ & \textbf{CLIP-T}↑ & \textbf{CLIP-D}↑ & \textbf{S.C.}↑ & \textbf{B.C.}↑ & \textbf{A.Q.}↑ & \textbf{I.Q.}↑ \\

\midrule
\multicolumn{13}{c}{\textbf{Relighting on Videos}} \\
RelightVid & \underline{0.2346} & \underline{0.0578} & \underline{0.7260} & \underline{0.8558} & \underline{0.5593} & \underline{61.16} & \textbf{0.2377} & \underline{0.0267} & \textbf{0.6885} & \textbf{0.8494} & \textbf{0.5688} & \textbf{67.01} \\
Lumen & 0.1865 & -0.0198 & 0.6108 & 0.8007 & 0.4180 & 45.62 & 0.2011 & -0.0624 & 0.5687 & 0.7900 & 0.4275 & 43.17 \\
\textbf{Ours} & \textbf{0.2403} & \textbf{0.0883} & \textbf{0.7386} & \textbf{0.8684} & \textbf{0.5834} & \textbf{66.93} & \underline{0.2309} & \textbf{0.0658} & \underline{0.6636} & \underline{0.8478} & \underline{0.5447} & \underline{57.08} \\
\midrule
\multicolumn{13}{c}{\textbf{Relighting on Gaussian Splatting}} \\
DGE & \underline{0.2304} & \underline{0.0810} & 0.8343 & 0.8944 & \underline{0.5888} & \textbf{69.71} & \textbf{0.2452} & \underline{0.0551} & 0.8550 & 0.9236 & 0.5213 & 39.28 \\
EditSplat & 0.2201 & 0.0363 & 0.8349 & 0.8898 & 0.5571 & 62.44 & 0.2291 & 0.0163 & \underline{0.8767} & \textbf{0.9261} & \underline{0.5544} & 43.36 \\
IN2N & 0.2264 & 0.0433 & \underline{0.8674} & 0.9118 & 0.5210 & 49.21 & 0.2142 & -0.0135 & \textbf{0.8810} & 0.9225 & 0.5309 & \underline{46.73} \\
IGS2GS & 0.2160 & 0.0422 & 0.8633 & \underline{0.9184} & 0.4769 & 39.86 & 0.2243 & 0.0195 & 0.8628 & \underline{0.9238} & 0.5079 & 38.46 \\
IGS2GS-IC & 0.2064 & 0.0251 & \textbf{0.8754} & \textbf{0.9278} & 0.4473 & 28.86 & 0.1977 & -0.0287 & 0.8715 & 0.9233 & 0.4628 & 24.82 \\
\textbf{Ours} & \textbf{0.2402} & \textbf{0.0881} & 0.8421 & 0.9050 & \textbf{0.6047} & \underline{68.81} & \underline{0.2308} & \textbf{0.0658} & 0.8629 & 0.9223 & \textbf{0.5792} & \textbf{50.62} \\
\bottomrule
\end{tabular}
\end{threeparttable}
}
\end{table}

\paragraph{Evaluation}
We conduct both qualitative and quantitative evaluations of our method. For quantitative analysis, following DGE, we evaluate CLIP score and CLIP directional score (noted as \textbf{CLIP-T} and \textbf{CLIP-D}) between rendered images and target text prompt to measure the alignment of 3D relit scenes and instruction. In addition, metrics such as PSNR, SSIM, and LPIPS are commonly used to evaluate how well a Gaussian Splatting (GS) scene fits the corresponding real-world scene. These metrics can also partially reflect the multi-view consistency of rendered images. However, in our experiments, we observed that when fine-tuning a pretrained GS scene, these metrics tend to favor models with smaller editing degrees, since less-edited supervision images make it easier for the GS to converge toward high multi-view consistency—closer to that of the original training images, which is unfair to models that produce more complex and visually refined results. Therefore, we omit these metrics in method comparison when evaluating the final relighting performance, and instead adopt VBench~\cite{vbench}, an end-to-end video quality assessment framework. We select the following indicators—\textbf{Subject Consistency}, \textbf{Background Consistency}, \textbf{Aesthetic Quality}, and \textbf{Imaging Quality}—to comprehensively evaluate the relighting results of different models. It is worth noting that since VBench is video-based, we need to provide a continuous video as input. For GS-based relighting methods, we generate a smooth camera trajectory by interpolating between the training viewpoints for each scene in the benchmark, and then render videos along these trajectories. For methods that perform relighting directly on videos, we take the relit frames corresponding to the training viewpoints, sort them by viewing direction, concatenate them into a single video, and then evaluate it using VBench.


\subsection{Comparisons with Prior Work}

Our baselines include state-of-the-art 3D editing methods such as IN2N~\cite{in2n}, DGE~\cite{dge}, and EditSplat~\cite{editsplat}, as well as video relighting approaches including RelightVid~\cite{relightvid} and Lumen~\cite{lumen}. In addition, we develop an IC-Light~\cite{iclight}-based variant of IGS2GS, which adapts IN2N to Gaussian Splatting scenes, replacing the original 2D image editor InstructPix2Pix~\cite{instructpix2pix}.

We present qualitative comparisons between our method and baselines on the IN2N, Mip-NeRF360, and ScanNet++ datasets to visually assess the effectiveness of our framework. In Fig.~\ref{fig:banner}, our approach produces multi-view consistent edits that faithfully reflect the user-provided lighting condition, while maintaining high-quality geometry and texture details across views. As shown in Tab.~\ref{tab:quantitative_results2} and Tab.~\ref{tab:quantitative_results3}, our method consistently outperforms prior approaches across all datasets and metrics. Specifically, our model achieves the highest scores while keeping the inference process in few minutes, indicating superior visual fidelity, cross-view consistency, semantic alignment and time consuming.

\begin{table}[t]
\centering
\caption{User preference study for various gaussian splatting relighting methods. We collected 29 users’ rankings of relighting results from different models along three dimensions—subjective preference, artistic aesthetics, and lighting-control consistency—and computed the average rank. }
\label{tab:preference_study}
\resizebox{\textwidth}{!}{
\begin{tabular}{lccc}
\toprule
\multirow{2}{*}{\textbf{Method}} & \multicolumn{3}{c}{Avg. Rank (\#) ↓} \\
& \textbf{Subjective Preference} & \textbf{Artistic Aesthetics} & \textbf{Lighting-Control Consistency} \\
\midrule
DGE & 2.15 & 2.23 & 2.31 \\
EditSplat & 3.23 & 3.28 & 3.48 \\
IN2N & 4.15 & 4.15 & 4.09 \\
IGS2GS & 5.45 & 5.28 & 5.05 \\
IGS2GS-IC & 4.36 & 4.26 & 4.01 \\
\textbf{Ours} & \textbf{1.57} & \textbf{1.69} & \textbf{1.87} \\
\bottomrule
\end{tabular}
}
\end{table}

Tab.~\ref{tab:preference_study} presents the preferences of 29 users for GS relighting methods shown in Fig.~\ref{fig:banner}, based on dimensions such as subjective preference, image-text matching, and lighting/artistic effects. The data indicate that our method is favored by users.

In summary, compared with prior methods, our method generates more realistic lighting and color adjustments, preserves fine-grained scene structures, and exhibits fewer artifacts in occluded or texture-rich regions.

\subsection{Ablation Study}

\begin{table}[t]
\centering
\caption{Component-wise ablation on IN2N dataset. Each component is removed progressively to analyze its contribution.}
\label{tab:ablation_components}
\resizebox{\textwidth}{!}{
\begin{tabular}{ccccccccc}
\toprule
\textbf{PAM} & \textbf{CV-Attn} & \textbf{IC-Light} & \textbf{CLIP-T}↑ & \textbf{CLIP-D}↑ & \textbf{S.C.}↑ & \textbf{B.C.}↑ & \textbf{A.Q.}↑ & \textbf{I.Q.}↑ \\
\midrule
\checkmark & \checkmark & \checkmark & 0.2580 & 0.1170 & 0.8748 & 0.9270 & 0.6043 & 57.80 \\
$\times$ & \checkmark & \checkmark & 0.2467 & 0.1115 & 0.8734 & 0.9243 & 0.6130 & 58.22 \\
$\times$ & $\times$ & \checkmark & 0.2667 & 0.1188 & 0.8631 & 0.9160 & 0.5957 & 56.36 \\
$\times$ & $\times$ & $\times$ & 0.2369 & 0.0918 & 0.8614 & 0.9016 & 0.5619 & 60.65 \\
\bottomrule
\end{tabular}
}
\end{table}

Next, we conduct an ablation study to evaluate the effectiveness of several key components in our editing pipeline: IC-Light Relighting Model, Multi-View Consistent Inference and Position-Align Module. Tab.~\ref{tab:ablation_components} shows the component-wise quantitative ablation study and then qualitative result of each component will be present.

\textbf{Component-wise Ablation.} We sequentially remove key components from our full pipeline, position-alignment module (w/o PAM), cross-view attention module (w/o CV-Attn), and IC-Light as the 2D image editor (w/o IC-Light). Table~\ref{tab:ablation_components} reports the corresponding metrics scores on the IN2N dataset. Removing PAM and CV-Attn leads to a noticeable drop in semantic alignment and multi-view consistency, which shows the effectiveness of our proprosed modules. However, we observed an abnormal increase in the CLIP score when only IC-Light is used. We speculate that this may be because the CLIP score emphasizes the overall alignment between the image and the text, while lacking training objectives related to lighting consistency. The Imaging Quality metric exhibits relatively “random’’ behavior on this task, which may be because it focuses more on image sharpness rather than lighting effects—an aspect that our proposed module is not designed to address.


\begin{table}[t]
\centering
\caption{Multi-view consistency ablation on Cross-view Attention. We evaluate PSNR, SSIM, LPIPS for renderings of relit GS scene on IN2N dataset.}
\label{tab:ablation_mv}
\begin{tabular}{l|ccc}
\toprule
Method & PSNR↑ & SSIM↑ &LPIPS↓\\
\midrule
w/o CV-Attn & 13.12 & 0.4602 & 0.4253 \\
w/ CV-Attn & \textbf{18.88} & \textbf{0.6267} & \textbf{0.2549}\\
\bottomrule
\end{tabular}
\end{table}

\newcolumntype{C}[1]{>{\centering\arraybackslash}m{#1}}

\begin{figure}[htbp]
\centering
\resizebox{\textwidth}{!}{
\setlength{\tabcolsep}{0pt}
\begin{tabular}{
  C{0.13\textwidth}C{0.13\textwidth}C{0.13\textwidth}
  @{\hskip 0.015\textwidth} 
  C{0.13\textwidth}C{0.13\textwidth}C{0.13\textwidth}
  @{\hskip 0.015\textwidth} 
  C{0.13\textwidth}C{0.13\textwidth}C{0.13\textwidth}
  @{\hskip 0.015\textwidth} 
  m{0.03\textwidth} 
}
\hline
\multicolumn{3}{c}{\textbf{inputs} / \textbf{instruction}} & \multicolumn{3}{c}{\textbf{relit images} } & \multicolumn{3}{c}{GS renders} & \\


\includegraphics[width=0.13\textwidth]{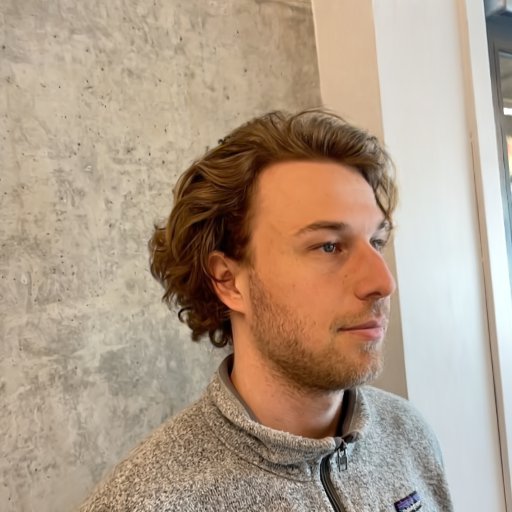} &
\includegraphics[width=0.13\textwidth]{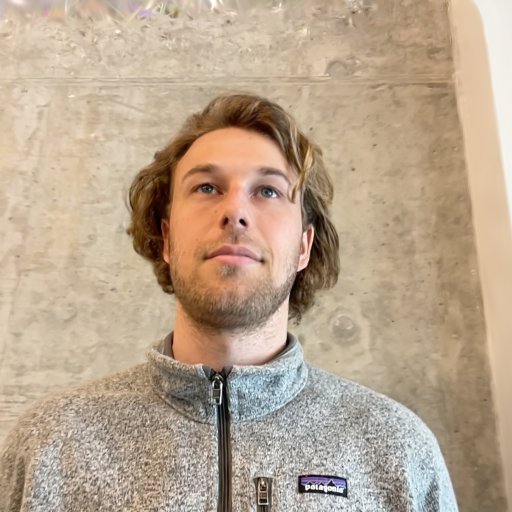} &
\includegraphics[width=0.13\textwidth]{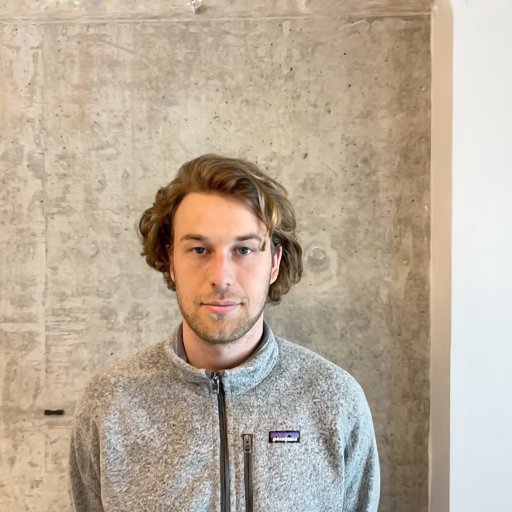} &
\includegraphics[width=0.13\textwidth]{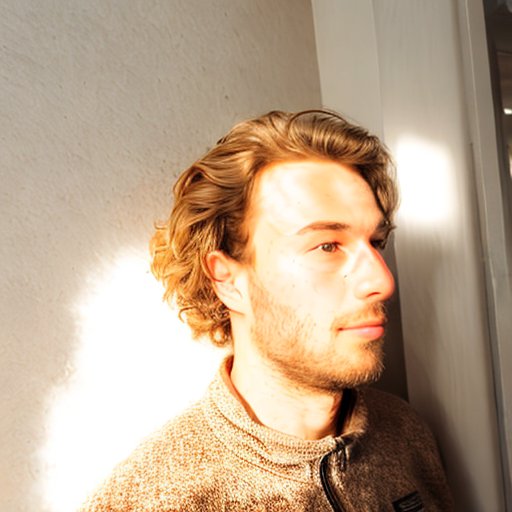} &
\includegraphics[width=0.13\textwidth]{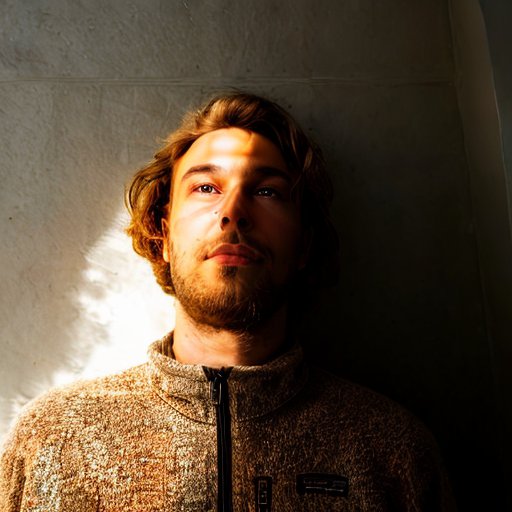} &
\includegraphics[width=0.13\textwidth]{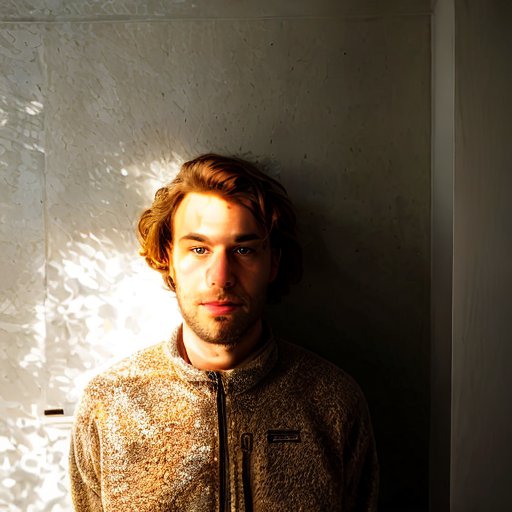} &
\includegraphics[width=0.13\textwidth]{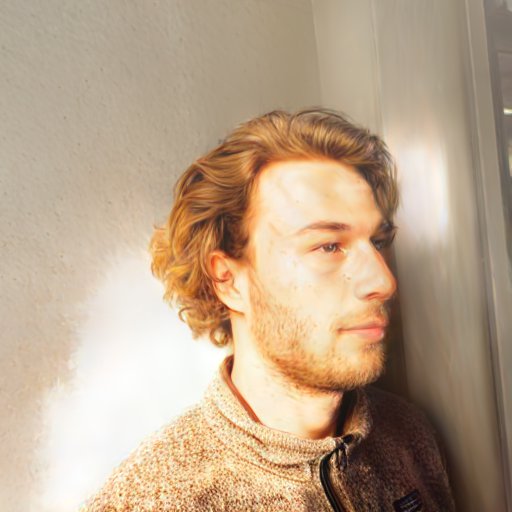} &
\includegraphics[width=0.13\textwidth]{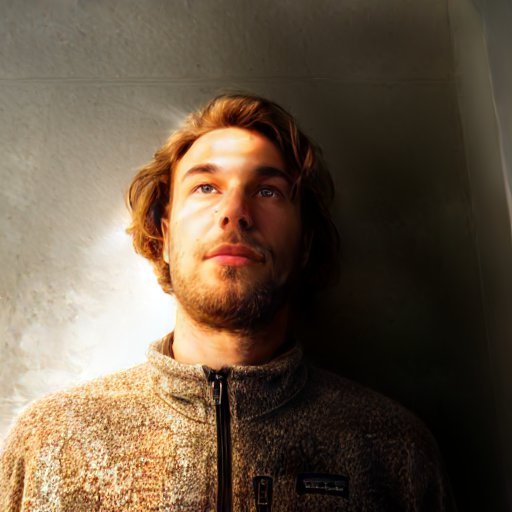} &
\includegraphics[width=0.13\textwidth]{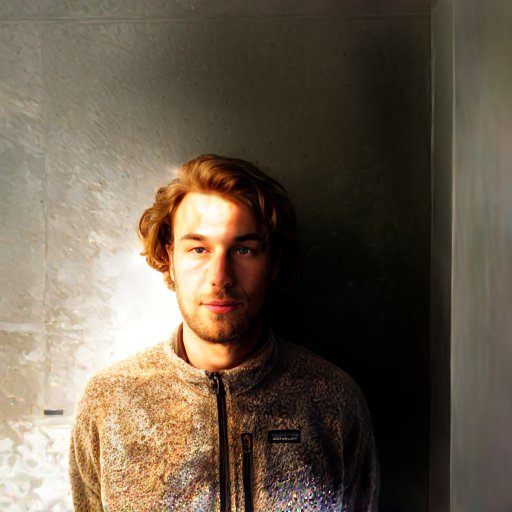} &
\rotatebox[origin=c]{90}{\raisebox{2.5em}{\tiny{MV-ICLight}}}\\

 
\multicolumn{3}{C{0.39\textwidth}}{
  \centering\footnotesize\textit{
    "detailed face, sunshine, indoor, warm atmosphere"
  }
} &
\includegraphics[width=0.13\textwidth]{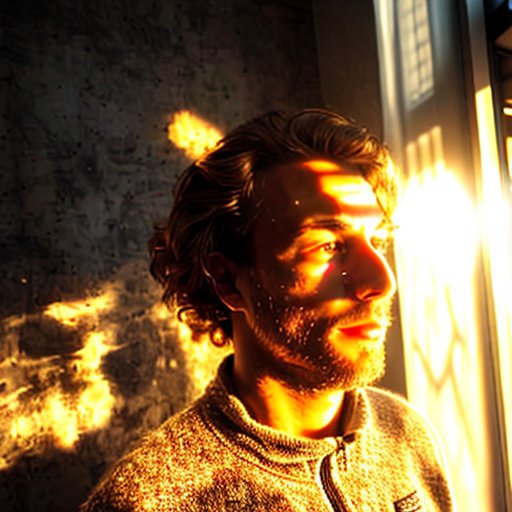} &
\includegraphics[width=0.13\textwidth]{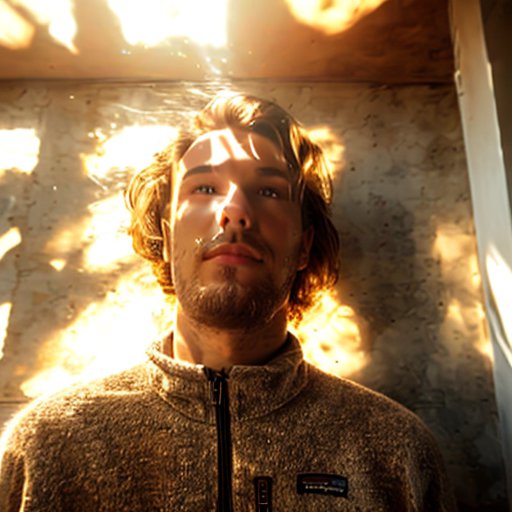} &
\includegraphics[width=0.13\textwidth]{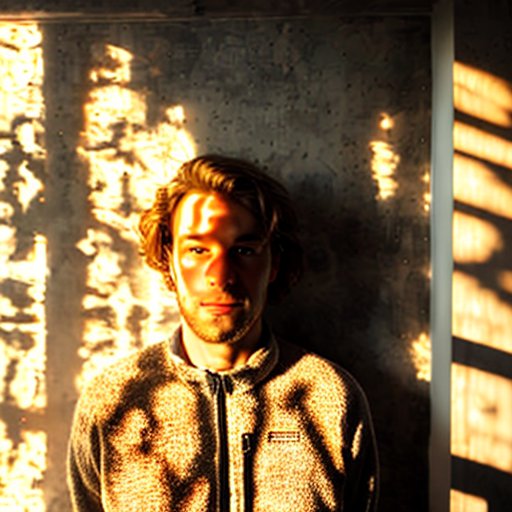} &
\includegraphics[width=0.13\textwidth]{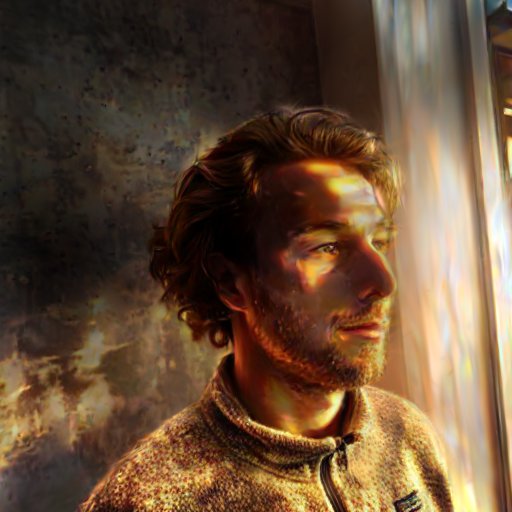} &
\includegraphics[width=0.13\textwidth]{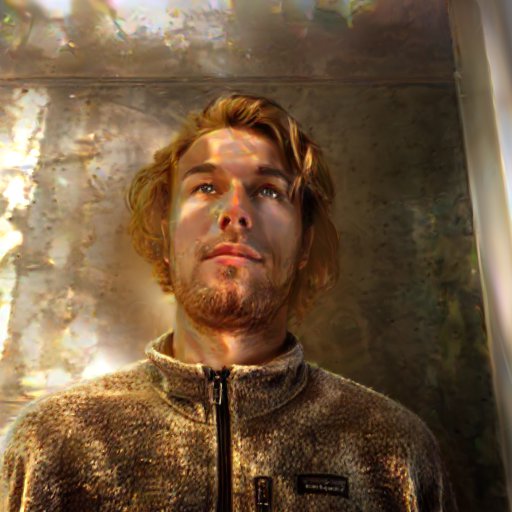} &
\includegraphics[width=0.13\textwidth]{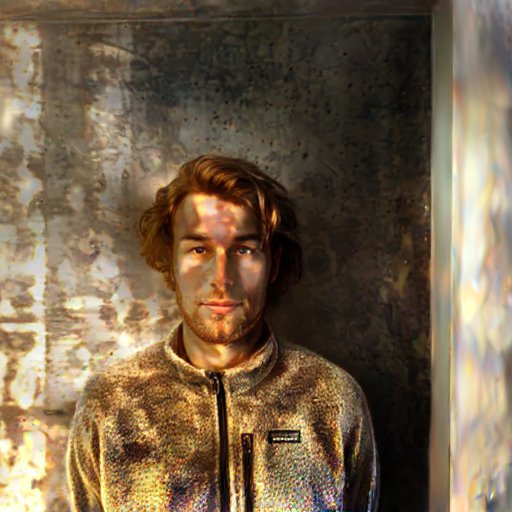} &
\rotatebox[origin=c]{90}{\raisebox{1.8em}{\tiny{IC-Light}}}\\

\includegraphics[width=0.13\textwidth]{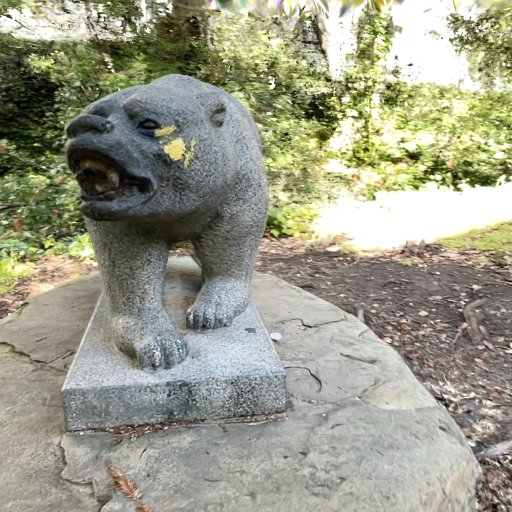} &
\includegraphics[width=0.13\textwidth]{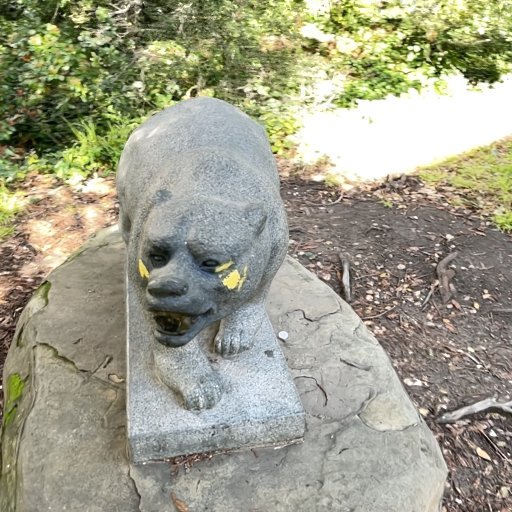} &
\includegraphics[width=0.13\textwidth]{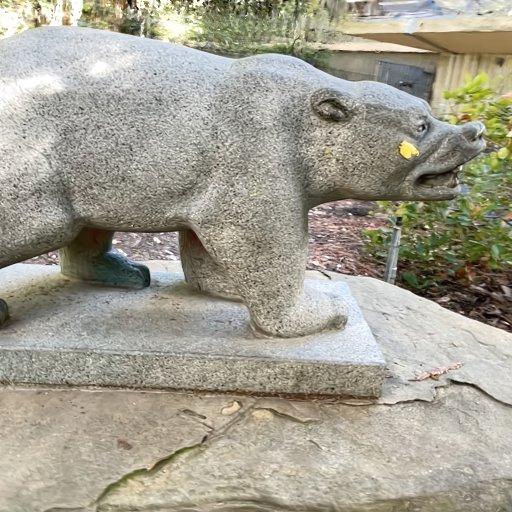} &
\includegraphics[width=0.13\textwidth]{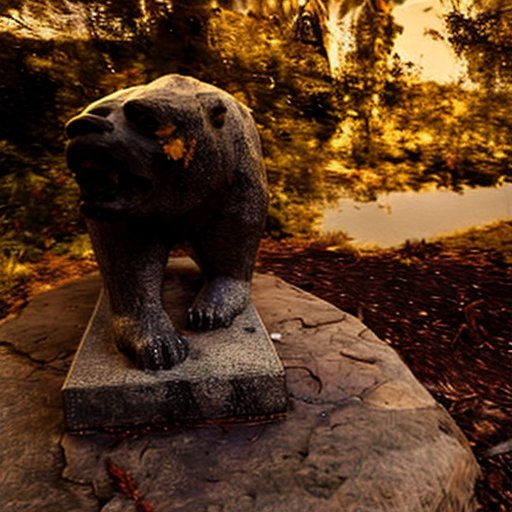} &
\includegraphics[width=0.13\textwidth]{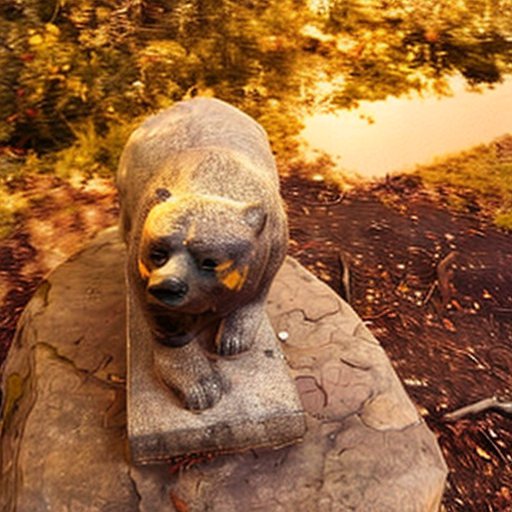} &
\includegraphics[width=0.13\textwidth]{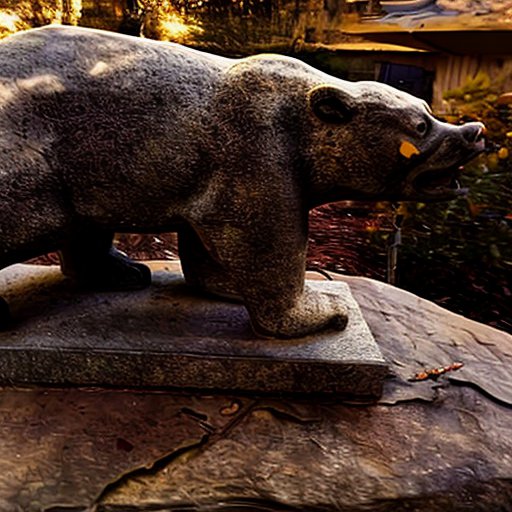} &
\includegraphics[width=0.13\textwidth]{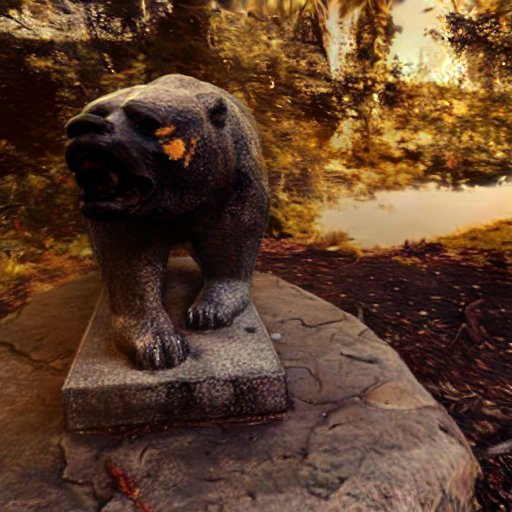} &
\includegraphics[width=0.13\textwidth]{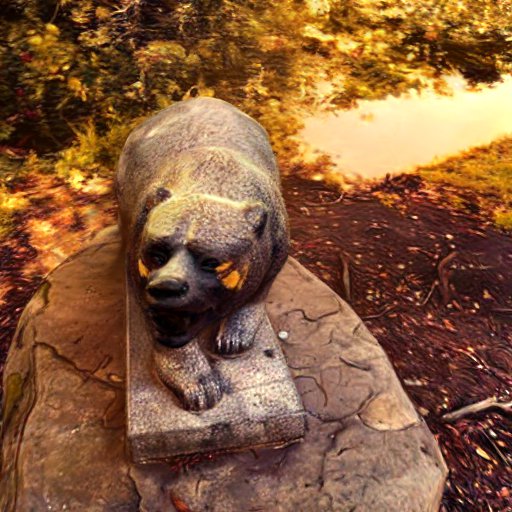} &
\includegraphics[width=0.13\textwidth]{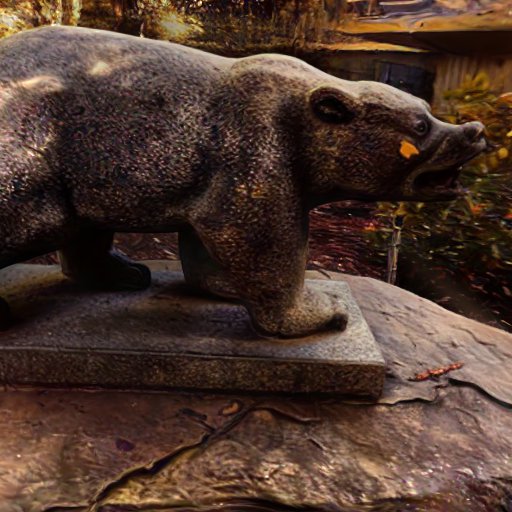} &
\rotatebox[origin=c]{90}{\raisebox{2.5em}{\tiny{MV-ICLight}}}\\

\multicolumn{3}{C{0.39\textwidth}}{
  \centering\footnotesize\textit{
    "detailed bear statue, marble, twilight, golden autumn forest, sunset glory"
  }
} &
\includegraphics[width=0.13\textwidth]{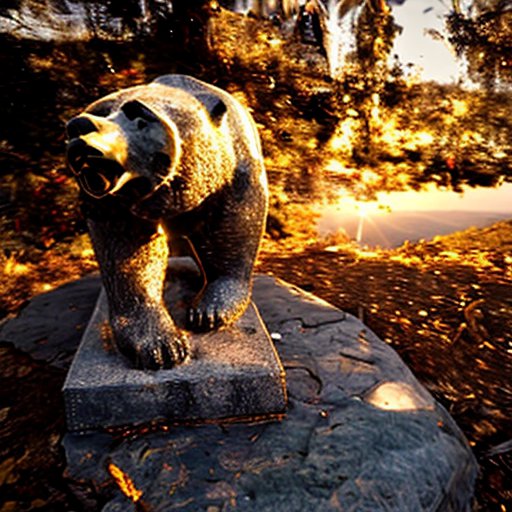} &
\includegraphics[width=0.13\textwidth]{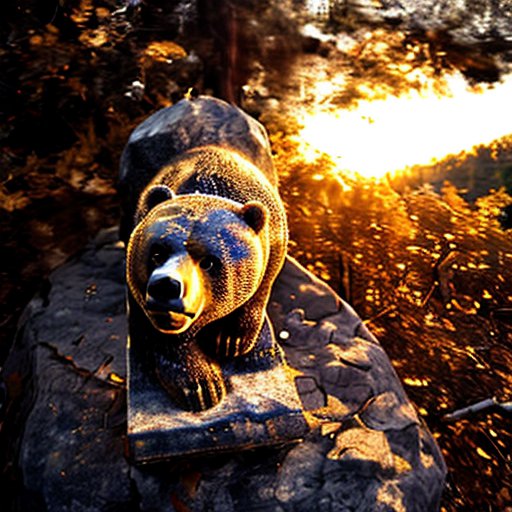} &
\includegraphics[width=0.13\textwidth]{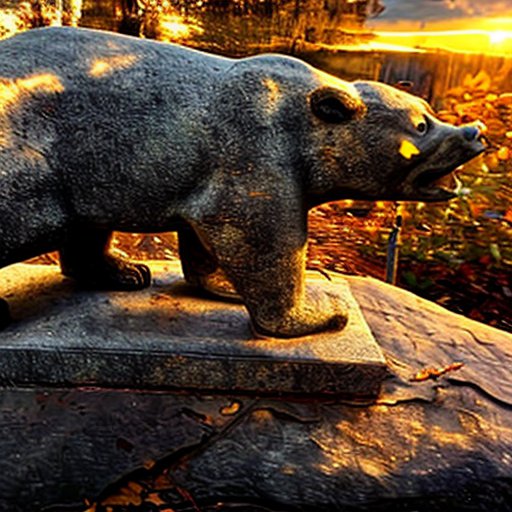} &
\includegraphics[width=0.13\textwidth]{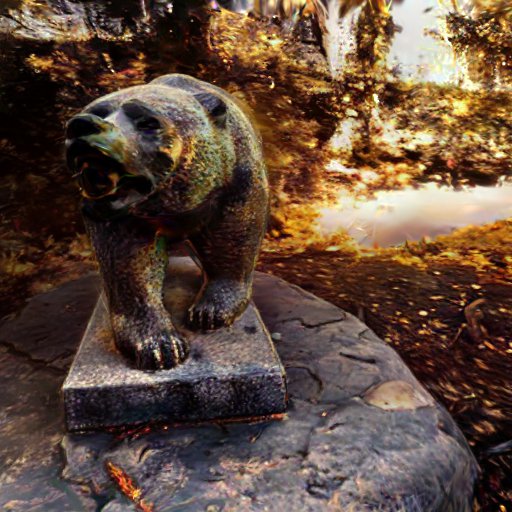} &
\includegraphics[width=0.13\textwidth]{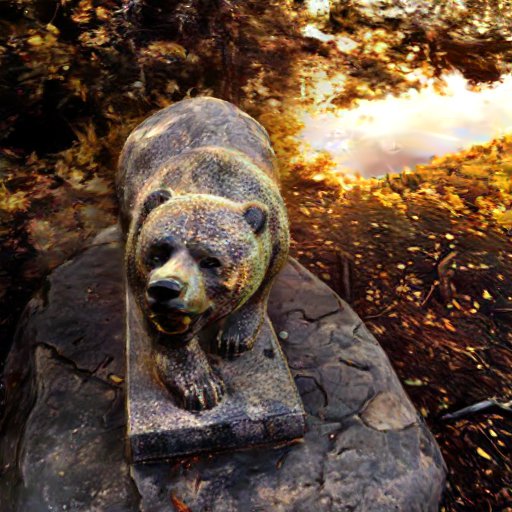} &
\includegraphics[width=0.13\textwidth]{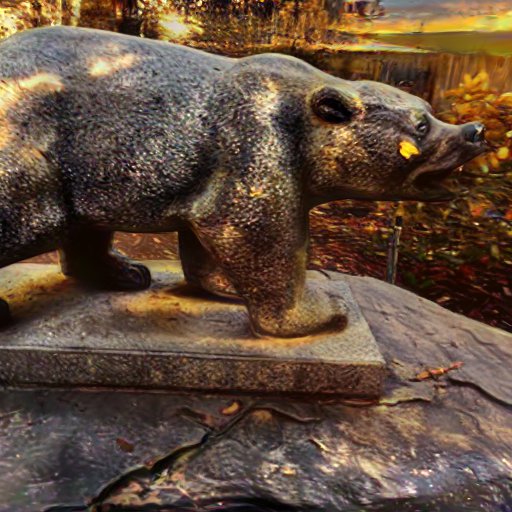} &
\rotatebox[origin=c]{90}{\raisebox{1.8em}{\tiny{IC-Light}}}\\

\includegraphics[width=0.13\textwidth]{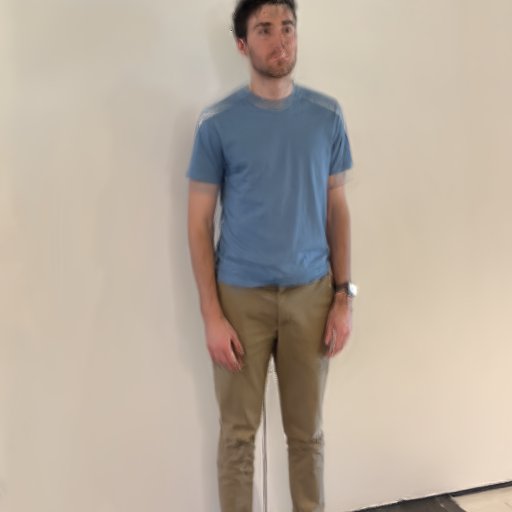} &
\includegraphics[width=0.13\textwidth]{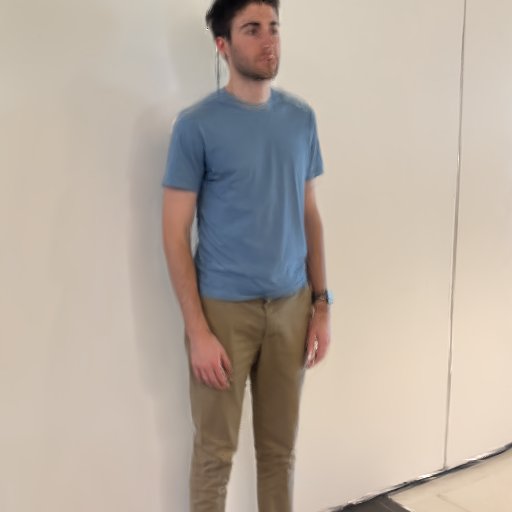} &
\includegraphics[width=0.13\textwidth]{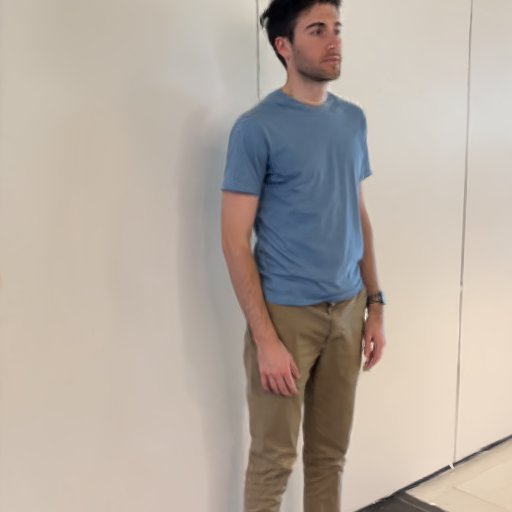} &
\includegraphics[width=0.13\textwidth]{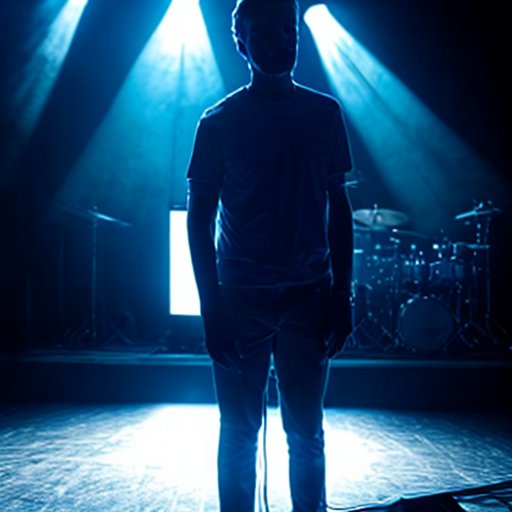} &
\includegraphics[width=0.13\textwidth]{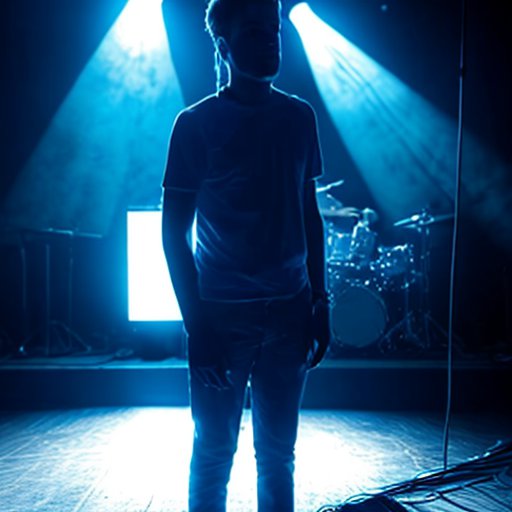} &
\includegraphics[width=0.13\textwidth]{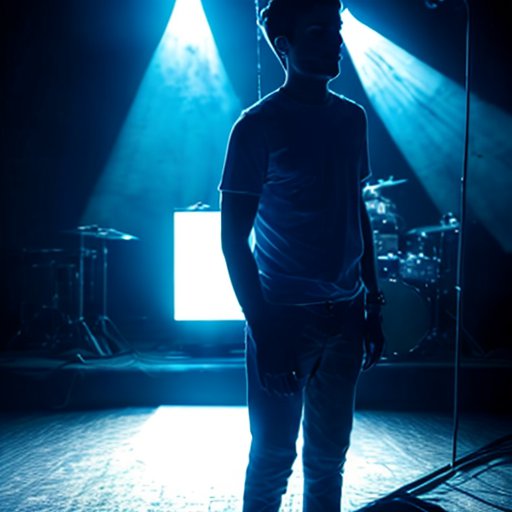} &
\includegraphics[width=0.13\textwidth]{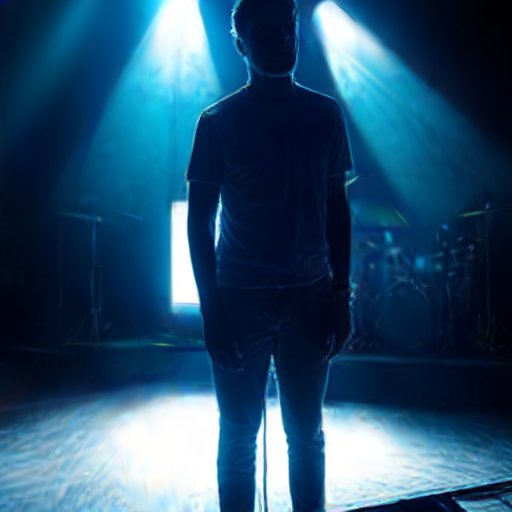} &
\includegraphics[width=0.13\textwidth]{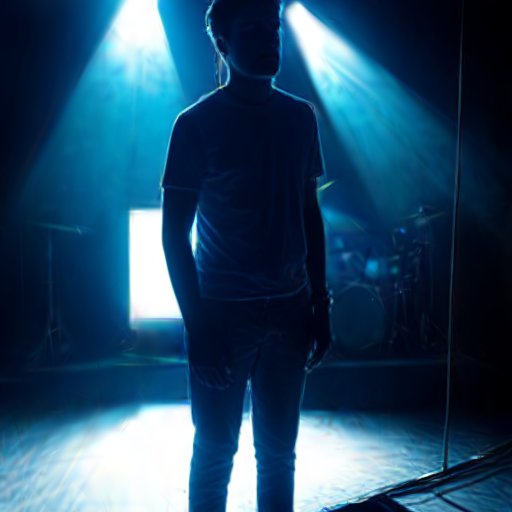} &
\includegraphics[width=0.13\textwidth]{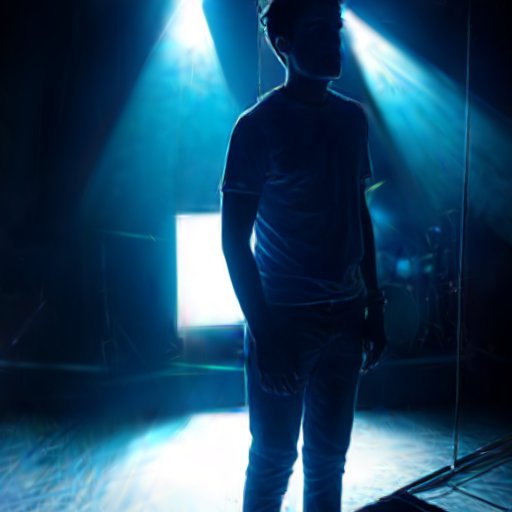} &
\rotatebox[origin=c]{90}{\raisebox{2.5em}{\tiny{MV-ICLight}}}\\

\multicolumn{3}{C{0.39\textwidth}}{
  \centering\footnotesize\textit{
    "detailed clear face, cool tunes, stage lighting, blue spotlight"
  }
} &
\includegraphics[width=0.13\textwidth]{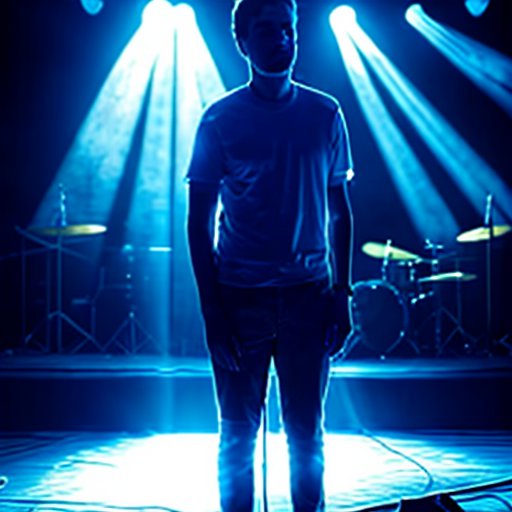} &
\includegraphics[width=0.13\textwidth]{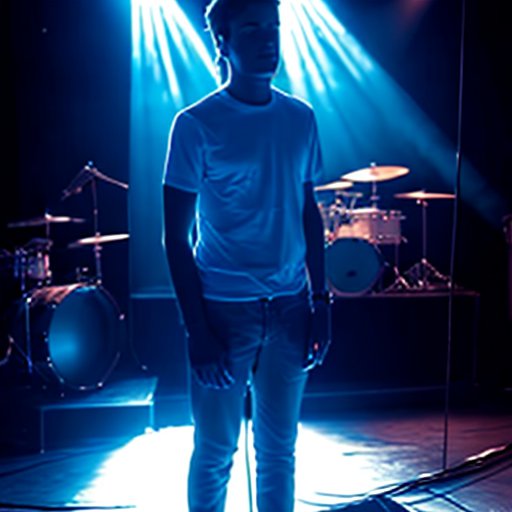} &
\includegraphics[width=0.13\textwidth]{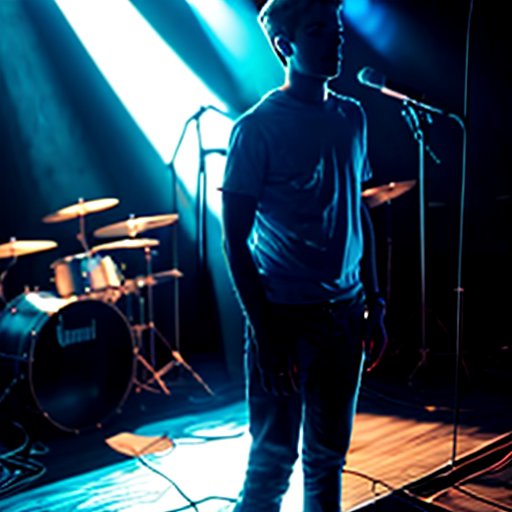} &
\includegraphics[width=0.13\textwidth]{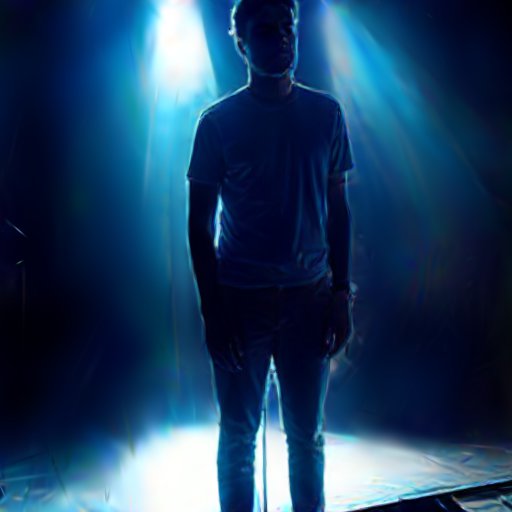} &
\includegraphics[width=0.13\textwidth]{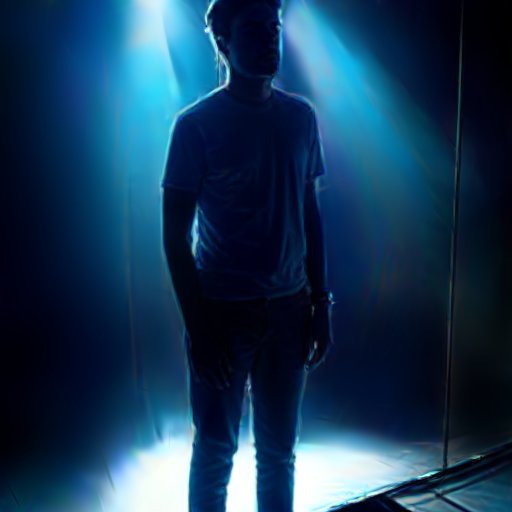} &
\includegraphics[width=0.13\textwidth]{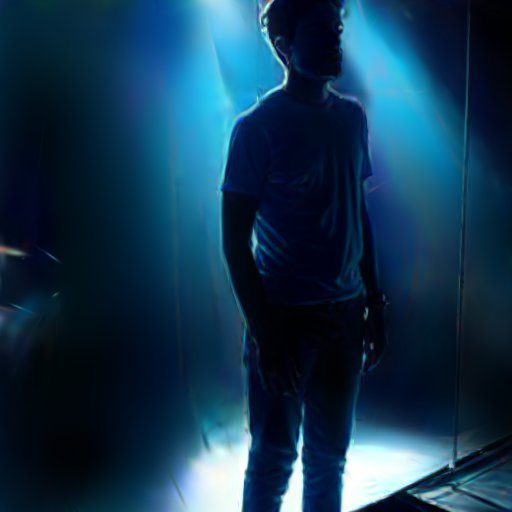} &
\rotatebox[origin=c]{90}{\raisebox{1.8em}{\tiny{IC-Light}}}\\

\includegraphics[width=0.13\textwidth]{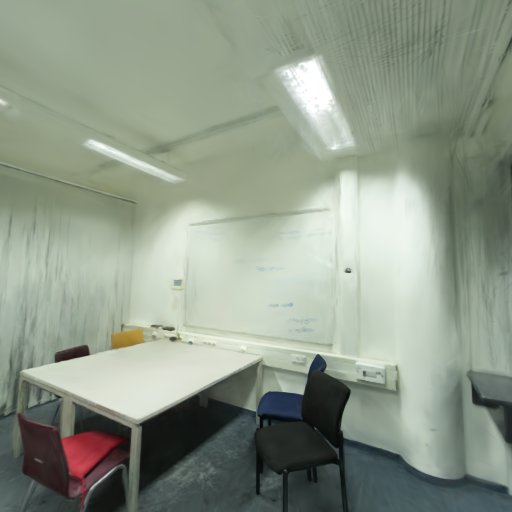} &
\includegraphics[width=0.13\textwidth]{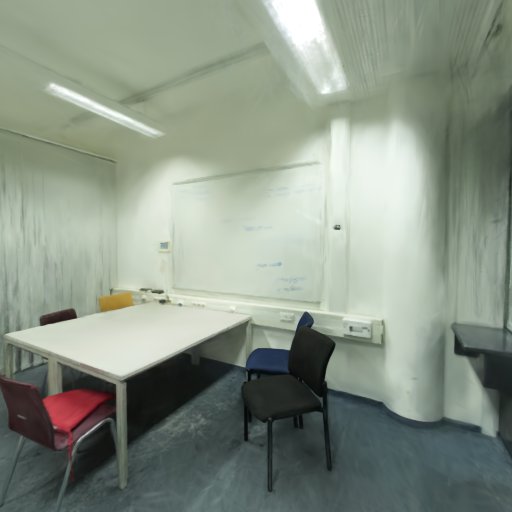} &
\includegraphics[width=0.13\textwidth]{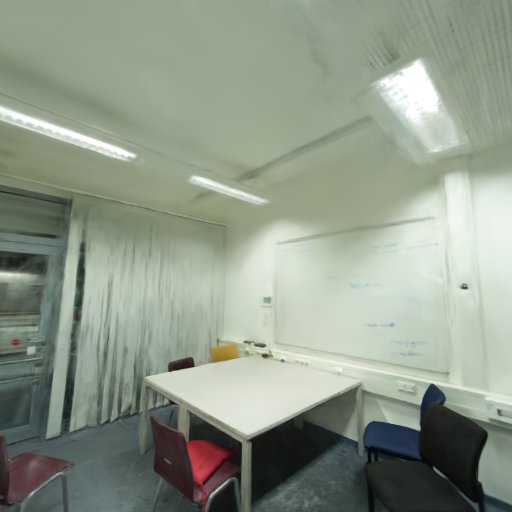} &
\includegraphics[width=0.13\textwidth]{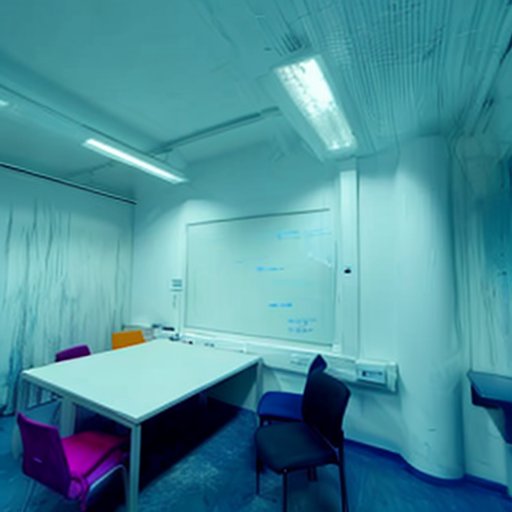} &
\includegraphics[width=0.13\textwidth]{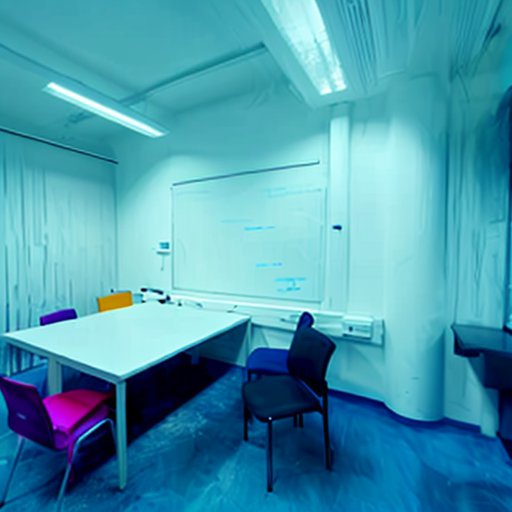} &
\includegraphics[width=0.13\textwidth]{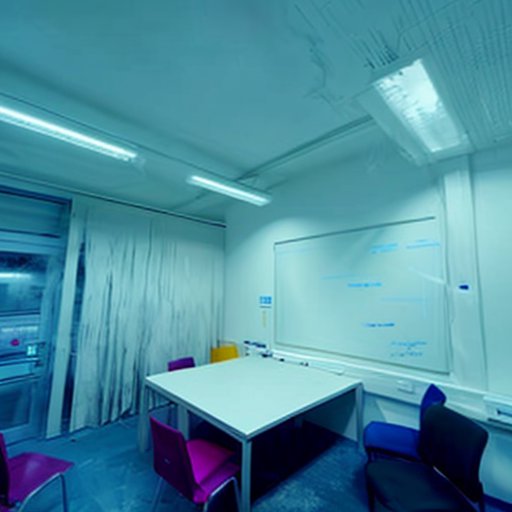} &
\includegraphics[width=0.13\textwidth]{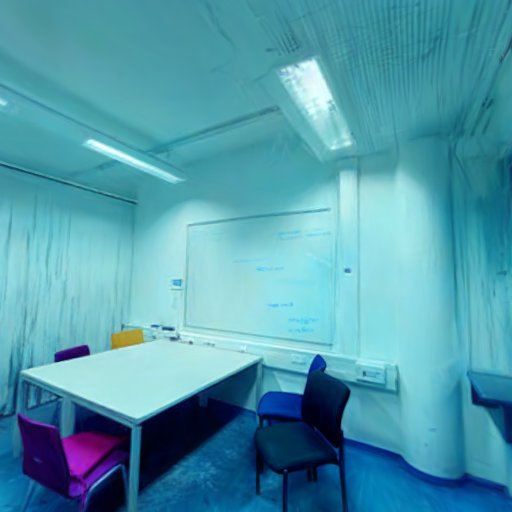} &
\includegraphics[width=0.13\textwidth]{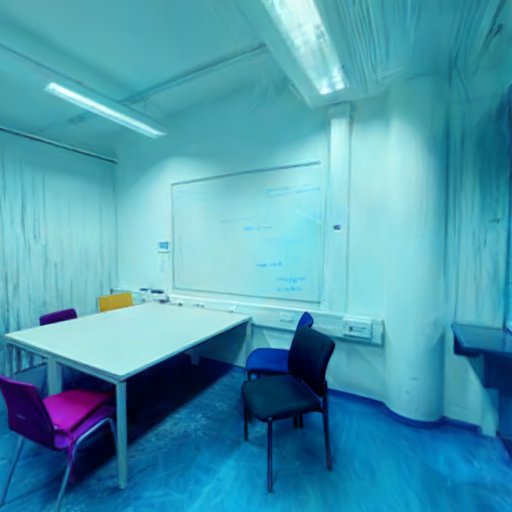} &
\includegraphics[width=0.13\textwidth]{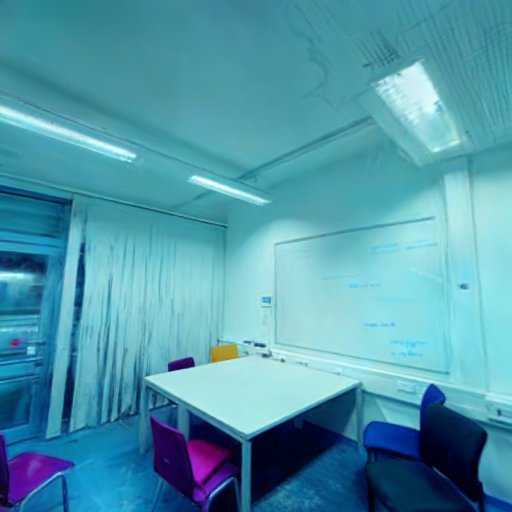} &
\rotatebox[origin=c]{90}{\raisebox{2.5em}{\tiny{MV-ICLight}}}\\

\multicolumn{3}{C{0.39\textwidth}}{
  \centering\footnotesize\textit{
    "office desk, indoor, night scene, cool fluorescent light"
  }
} &
\includegraphics[width=0.13\textwidth]{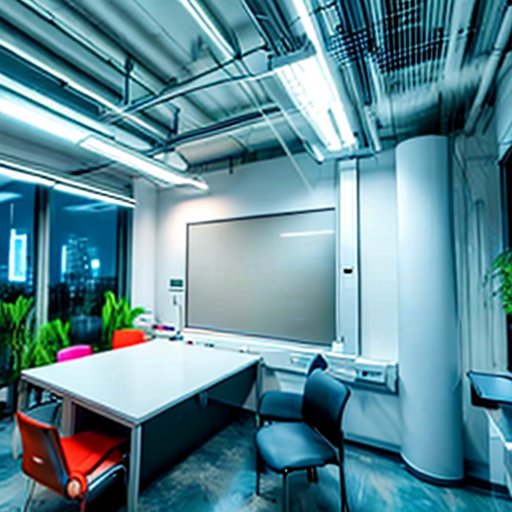} &
\includegraphics[width=0.13\textwidth]{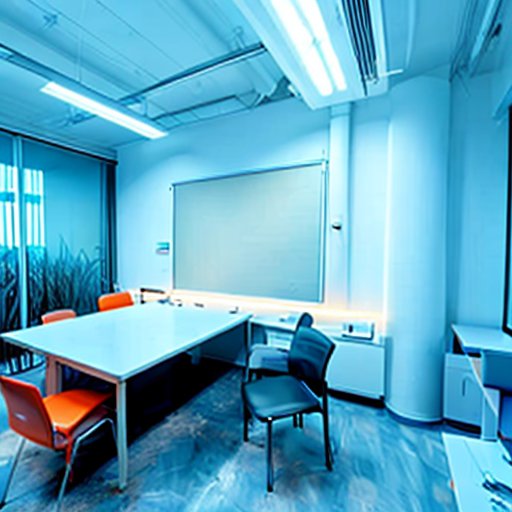} &
\includegraphics[width=0.13\textwidth]{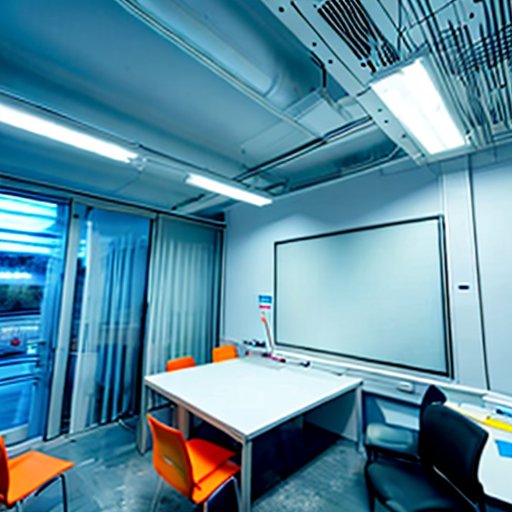} &
\includegraphics[width=0.13\textwidth]{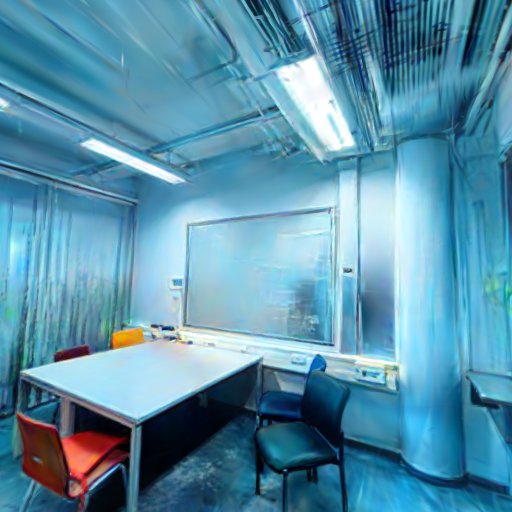} &
\includegraphics[width=0.13\textwidth]{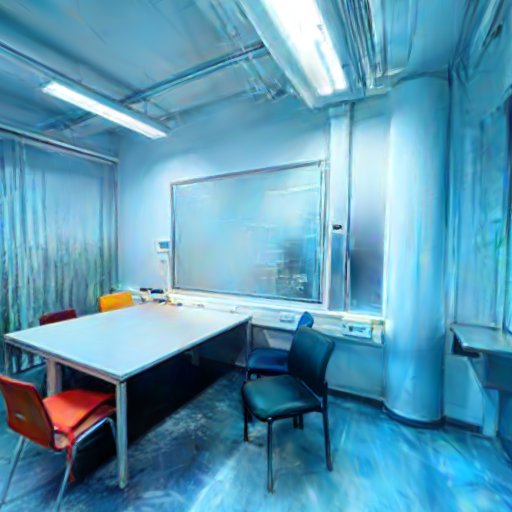} &
\includegraphics[width=0.13\textwidth]{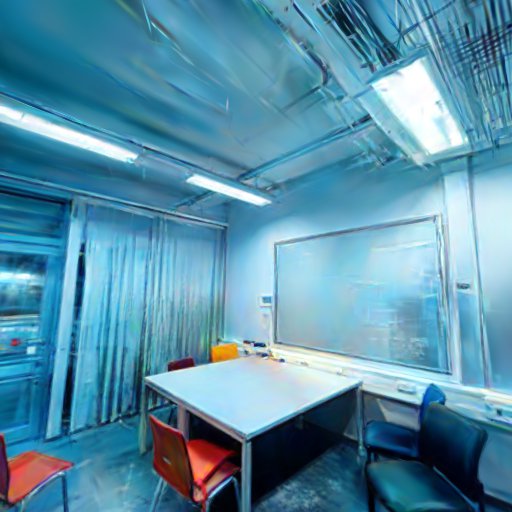} &
\rotatebox[origin=c]{90}{\raisebox{1.8em}{\tiny{IC-Light}}}\\

\includegraphics[width=0.13\textwidth]{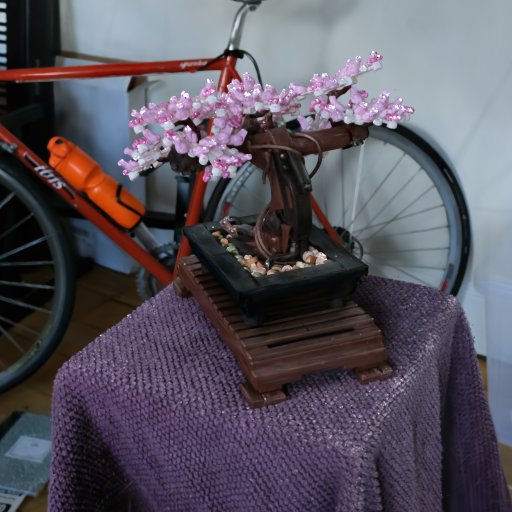} &
\includegraphics[width=0.13\textwidth]{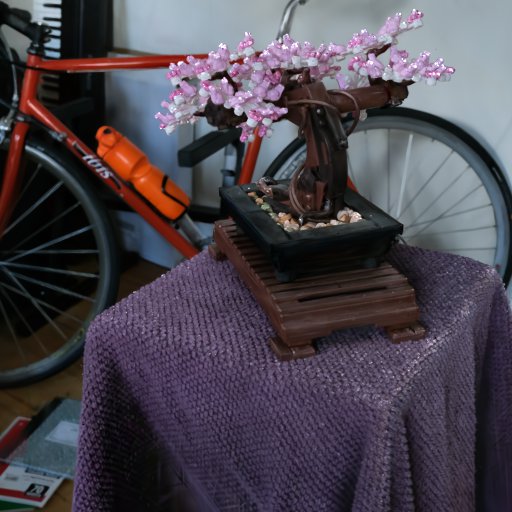} &
\includegraphics[width=0.13\textwidth]{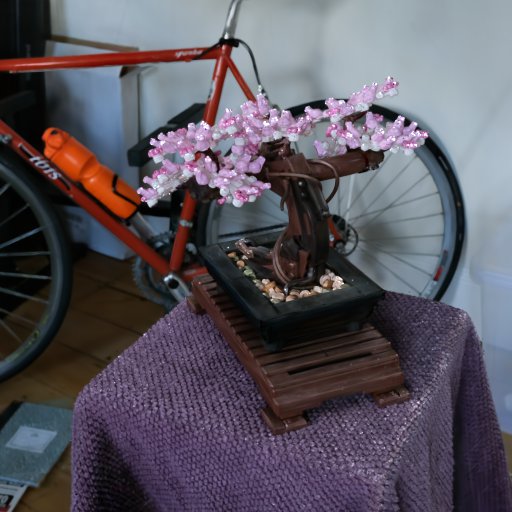} &
\includegraphics[width=0.13\textwidth]{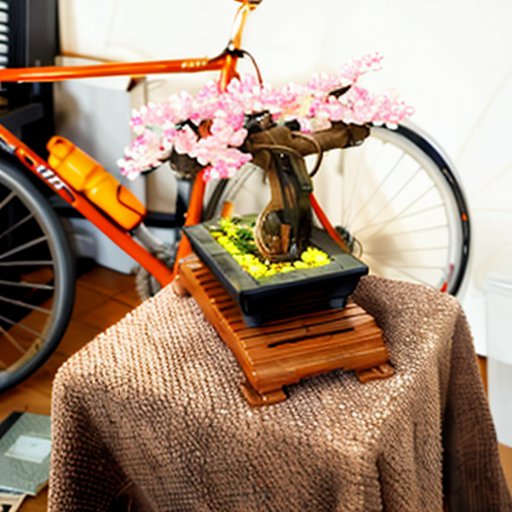} &
\includegraphics[width=0.13\textwidth]{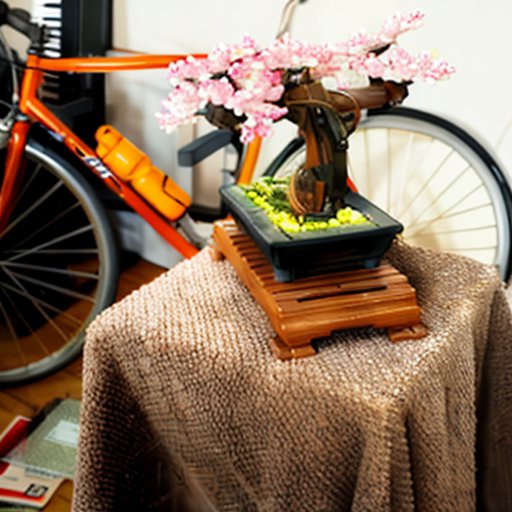} &
\includegraphics[width=0.13\textwidth]{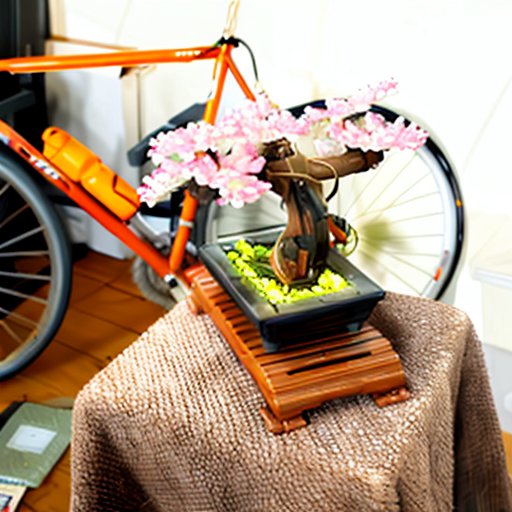} &
\includegraphics[width=0.13\textwidth]{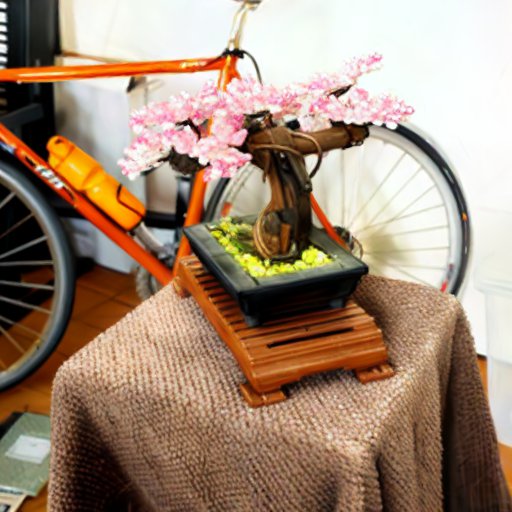} &
\includegraphics[width=0.13\textwidth]{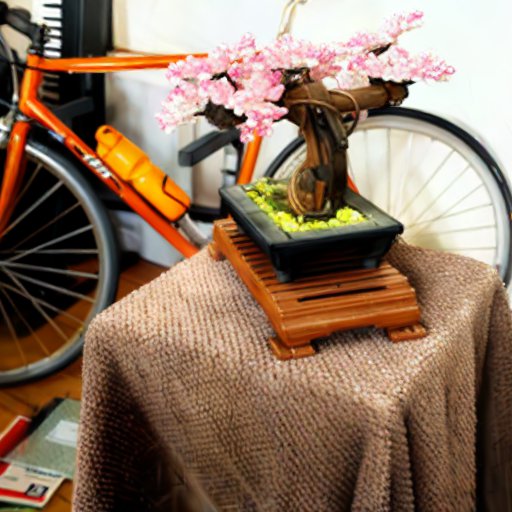} &
\includegraphics[width=0.13\textwidth]{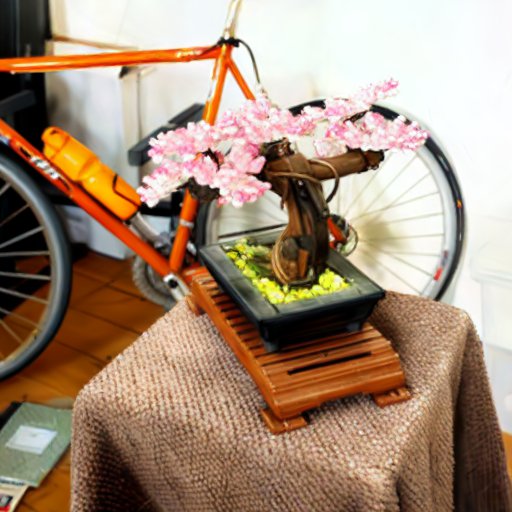} &
\rotatebox[origin=c]{90}{\raisebox{2.5em}{\tiny{MV-ICLight}}}\\

\multicolumn{3}{C{0.39\textwidth}}{
  \centering\footnotesize\textit{
    "bonsai tree, indoor, warm desk lamp illumination, cozy atmosphere"
  }
} &
\includegraphics[width=0.13\textwidth]{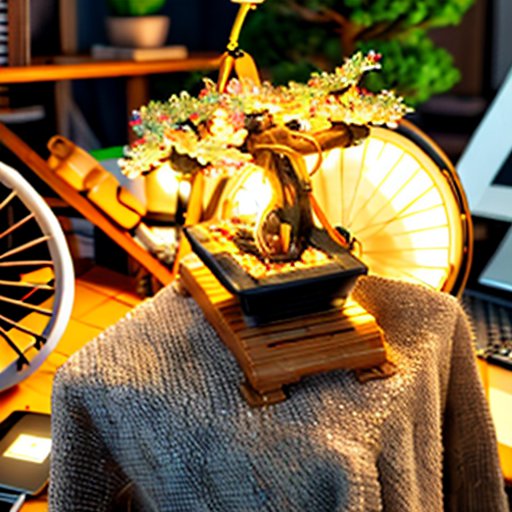} &
\includegraphics[width=0.13\textwidth]{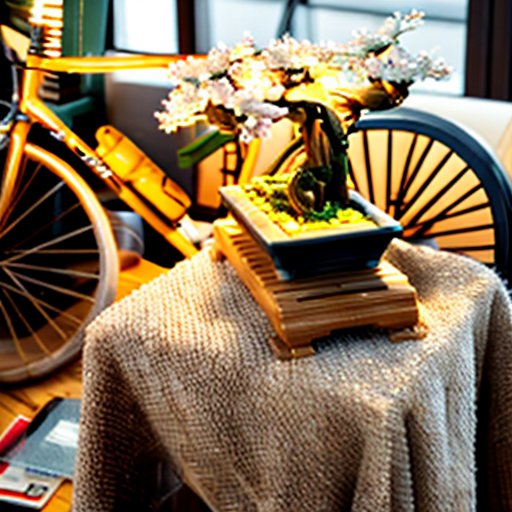} &
\includegraphics[width=0.13\textwidth]{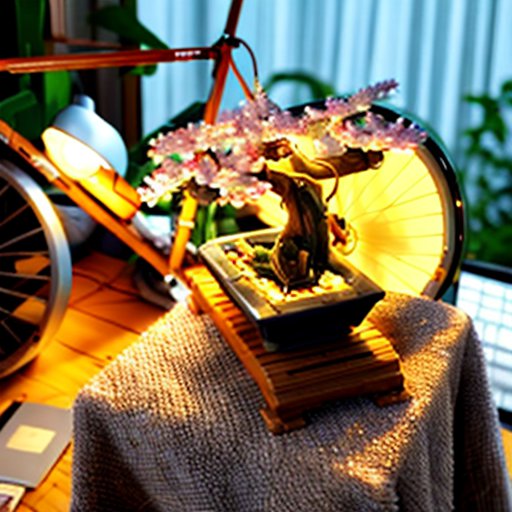} &
\includegraphics[width=0.13\textwidth]{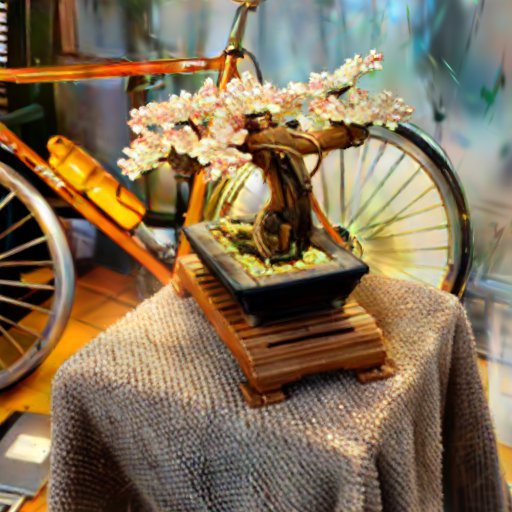} &
\includegraphics[width=0.13\textwidth]{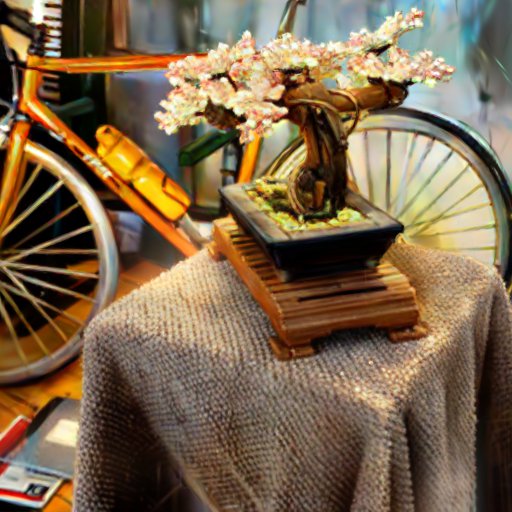} &
\includegraphics[width=0.13\textwidth]{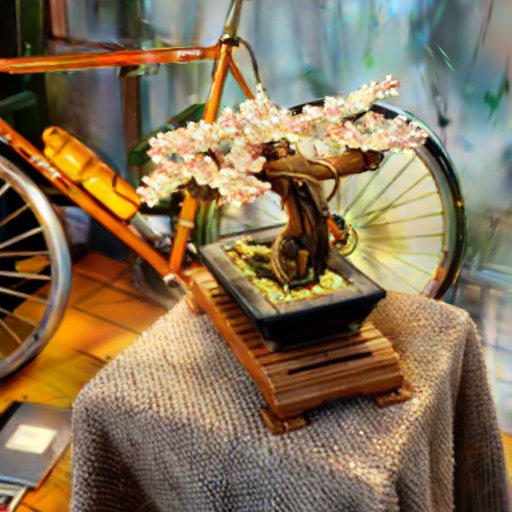} &
\rotatebox[origin=c]{90}{\raisebox{1.8em}{\tiny{IC-Light}}}\\

\hline
\end{tabular}
}
\caption{Ablation study of multi-view consistency. For each scene, first row shows the relit images and renders from fine-tuned GS scene with our \textbf{MV-ICLight}, while second row shows the ones with vanilla IC-Light. Our \textbf{MV-ICLight} shows a great multi-view consistency between sampled perspectives which contributes to divergency of GS finetuning. On the other hand, vanilla IC-Light without any multi-view consistency constrain generates diverse relit images, which results a degrade and blurry GS rendering.}
\label{fig:mv_ablation}
\end{figure}

\textbf{Relighting without Multi-View Consistency.} To evaluate the effectiveness of our multi-view consistent relighting, we replace it with independent per-view relighting for the same set of views and use those results to fine-tune Gaussians. As shown in Tab.~\ref{tab:ablation_mv}, incorporating multi-view consistency significantly improves reconstruction metrics such as PSNR, SSIM and LPIPS, demonstrating the advantage of maintaining cross-view coherence. Furthermore, Fig.~\ref{fig:mv_ablation} illustrates intermediate 2D relighting results with and without multi-view consistency. It can be clearly observed that consistent relighting leads to visually coherent appearances across views, while independent relighting causes noticeable discrepancies among them.

\newcolumntype{C}[1]{>{\centering\arraybackslash}m{#1}}

\begin{figure}[htbp]
\centering
\resizebox{\textwidth}{!}{
\setlength{\tabcolsep}{0pt}
\begin{tabular}{
  C{0.3\textwidth}
  @{\hskip 0.01\textwidth} 
  C{0.13\textwidth}C{0.13\textwidth}C{0.13\textwidth}
  @{\hskip 0.01\textwidth} 
  C{0.13\textwidth}C{0.13\textwidth}C{0.13\textwidth}
  @{\hskip 0.01\textwidth} 
  C{0.13\textwidth}C{0.13\textwidth}C{0.13\textwidth}
}
\hline
\multicolumn{1}{c}{\textbf{instruction}} & \multicolumn{3}{c}{\textbf{inputs} / \textbf{init latents}} & \multicolumn{3}{c}{\textbf{with PAM}} & \multicolumn{3}{c}{\textbf{w/o PAM}} \\

{\centering\footnotesize\textit{"detailed face, sunshine, indoor, warm atmosphere,"}} &

\includegraphics[width=0.13\textwidth]{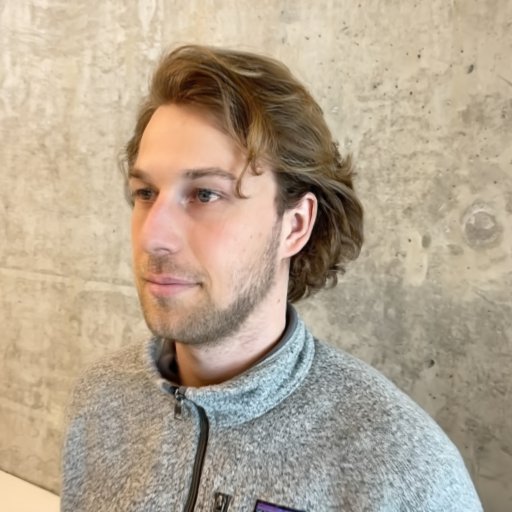} &
\includegraphics[width=0.13\textwidth]{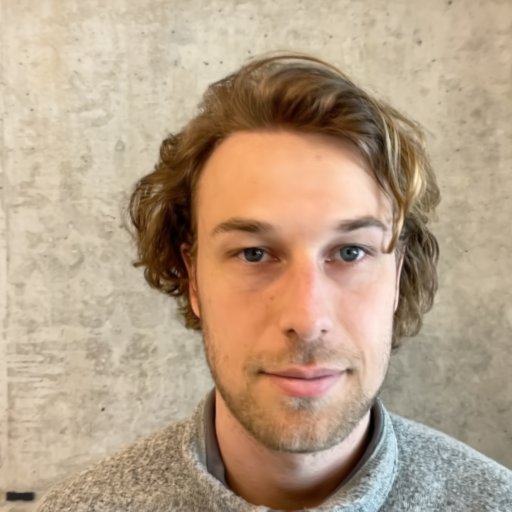} &
\includegraphics[width=0.13\textwidth]{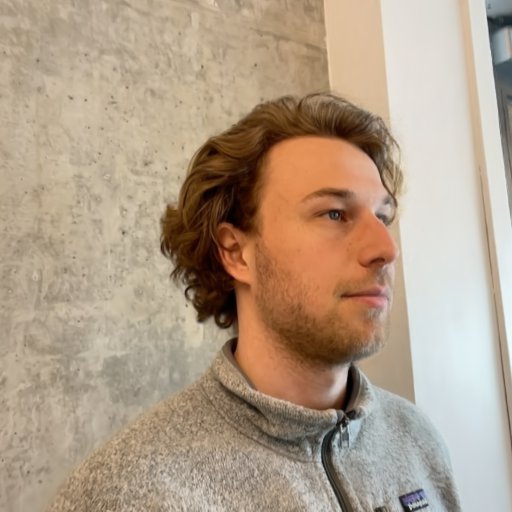} &
\includegraphics[width=0.13\textwidth]{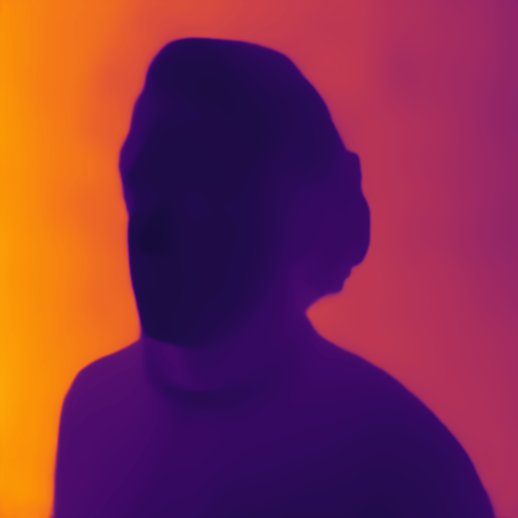} &
\includegraphics[width=0.13\textwidth]{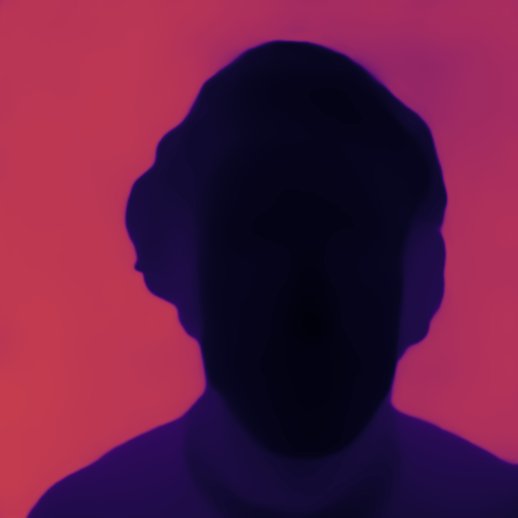} &
\includegraphics[width=0.13\textwidth]{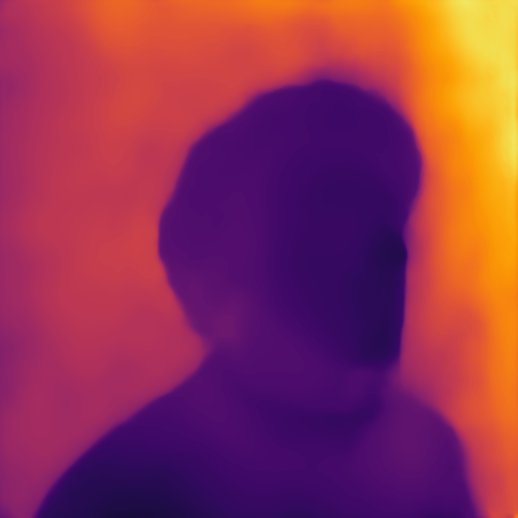} &
\includegraphics[width=0.13\textwidth]{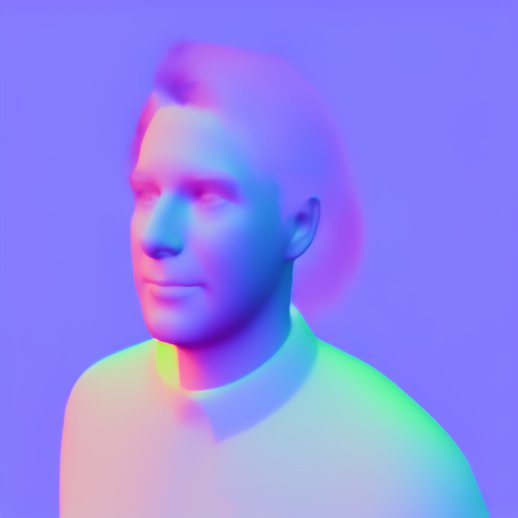} &
\includegraphics[width=0.13\textwidth]{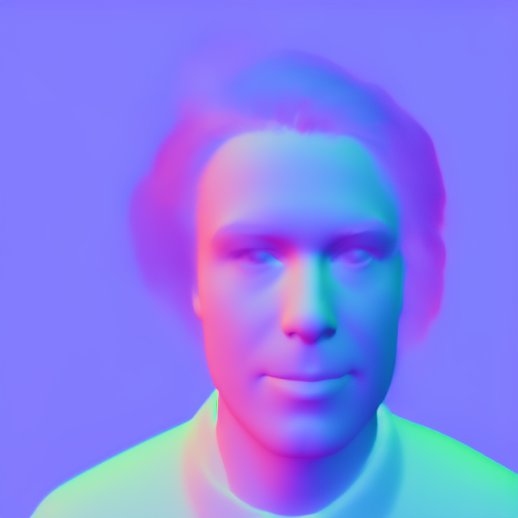} &
\includegraphics[width=0.13\textwidth]{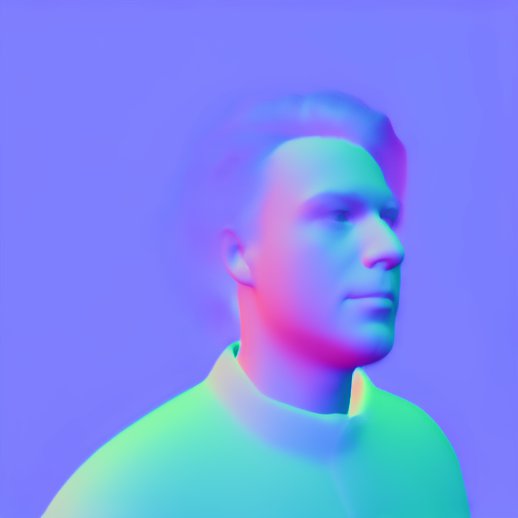} \\


{\centering\footnotesize\textit{"..., light from \textbf{left}"}} & 
\includegraphics[width=0.13\textwidth]{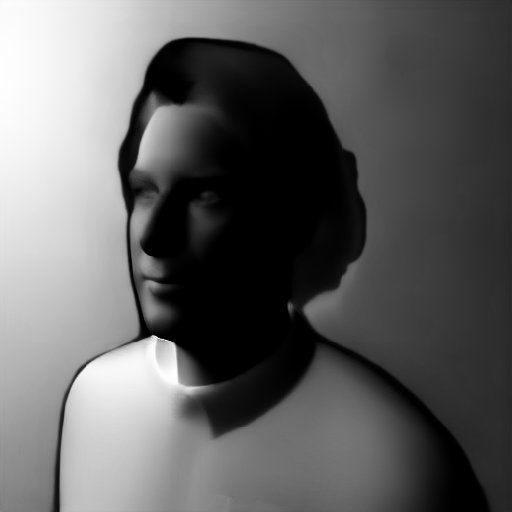} &
\includegraphics[width=0.13\textwidth]{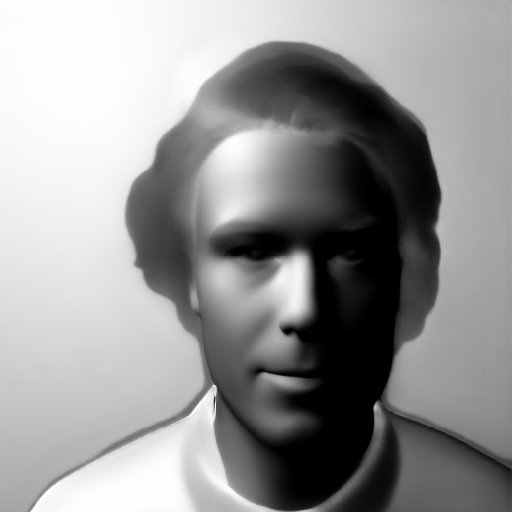} &
\includegraphics[width=0.13\textwidth]{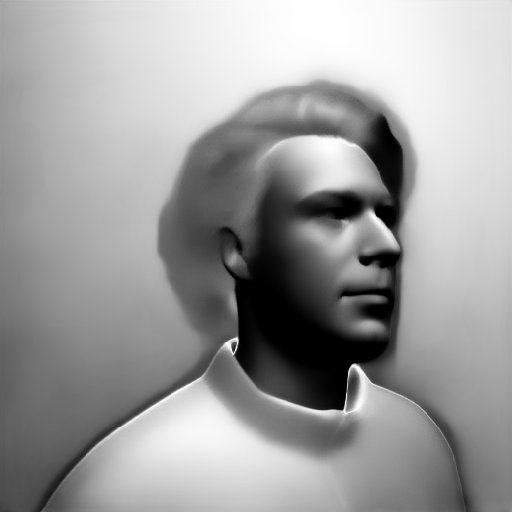} &
\includegraphics[width=0.13\textwidth]{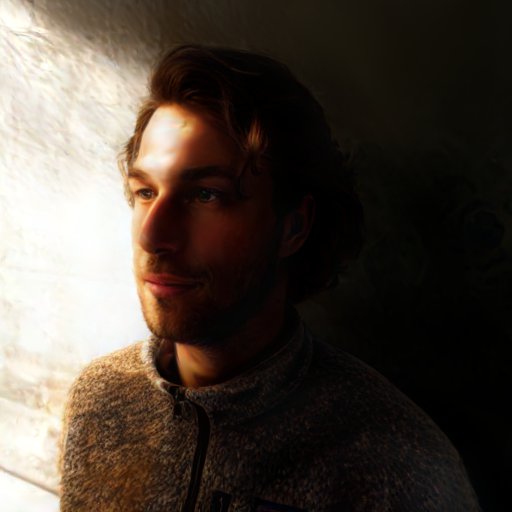} &
\includegraphics[width=0.13\textwidth]{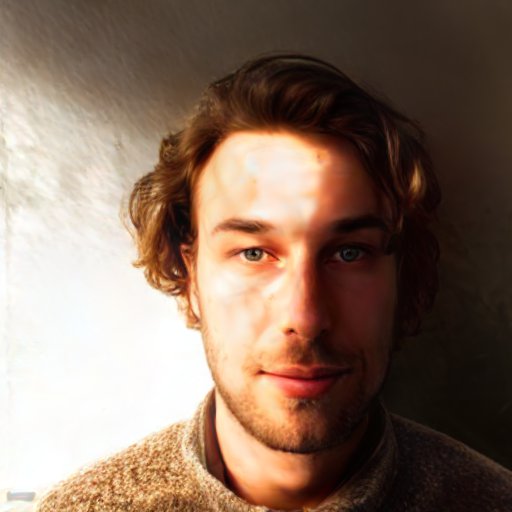} &
\includegraphics[width=0.13\textwidth]{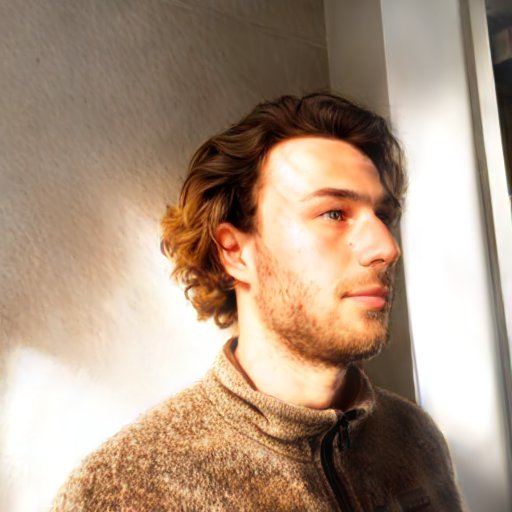} &
\includegraphics[width=0.13\textwidth]{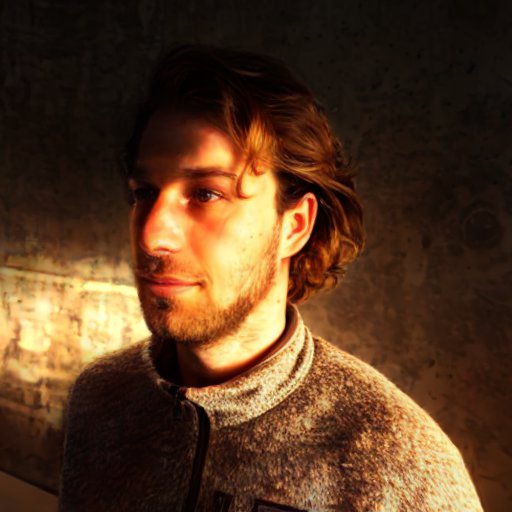} &
\includegraphics[width=0.13\textwidth]{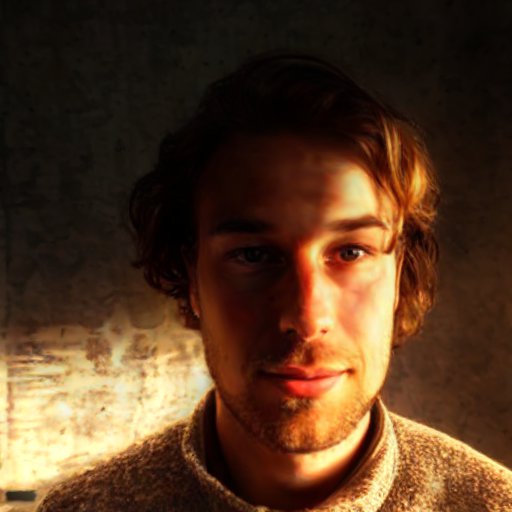} &
\includegraphics[width=0.13\textwidth]{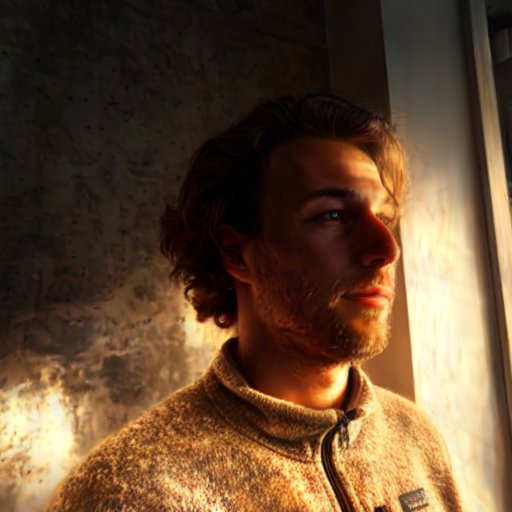} \\

{\centering\footnotesize\textit{"..., light from \textbf{right}"}} & 
\includegraphics[width=0.13\textwidth]{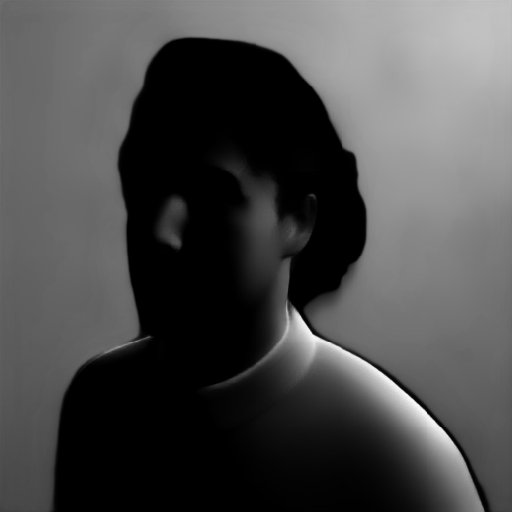} &
\includegraphics[width=0.13\textwidth]{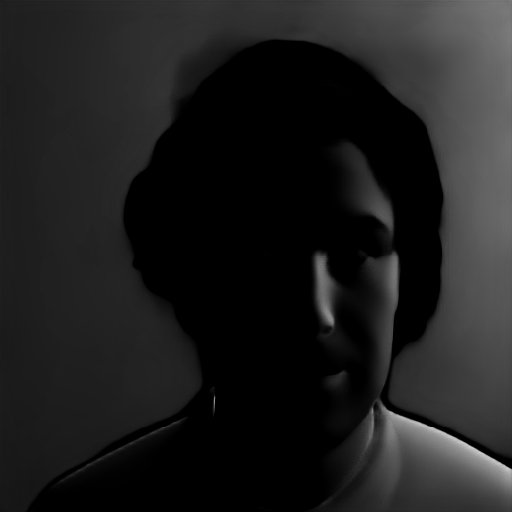} &
\includegraphics[width=0.13\textwidth]{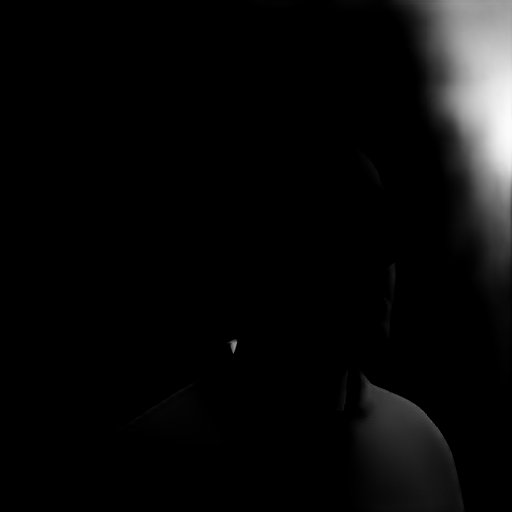} &
\includegraphics[width=0.13\textwidth]{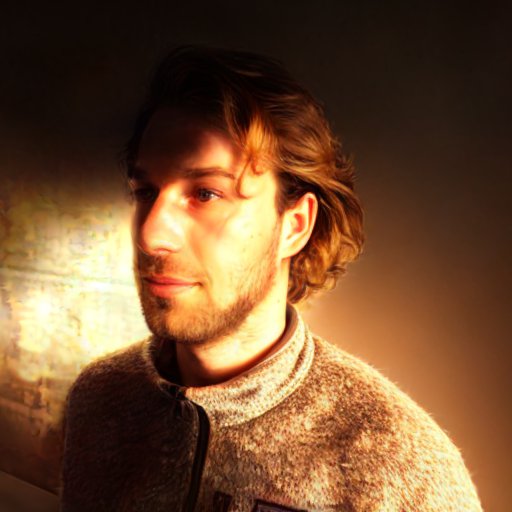} &
\includegraphics[width=0.13\textwidth]{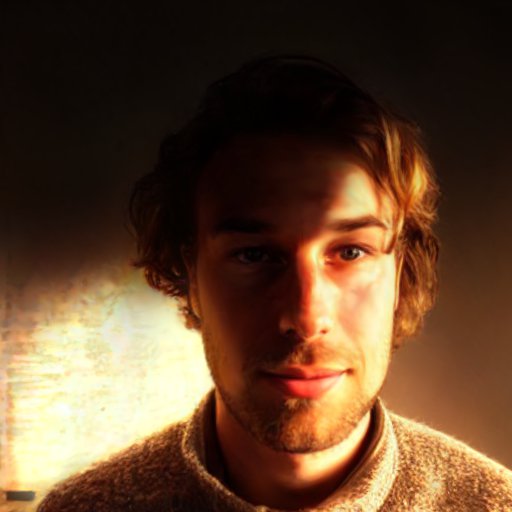} &
\includegraphics[width=0.13\textwidth]{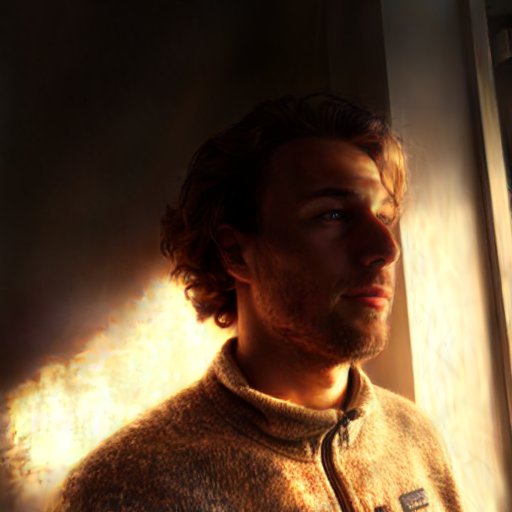} &
\includegraphics[width=0.13\textwidth]{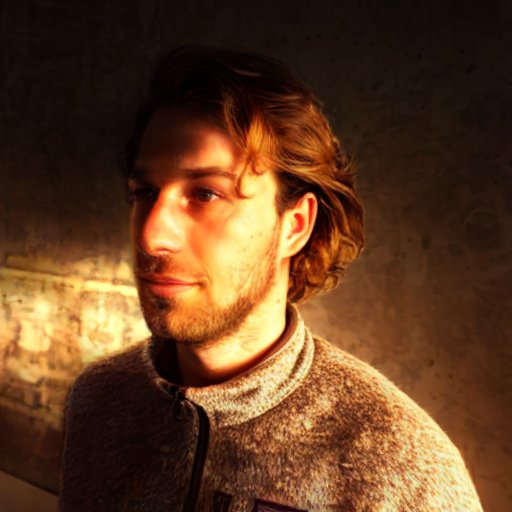} &
\includegraphics[width=0.13\textwidth]{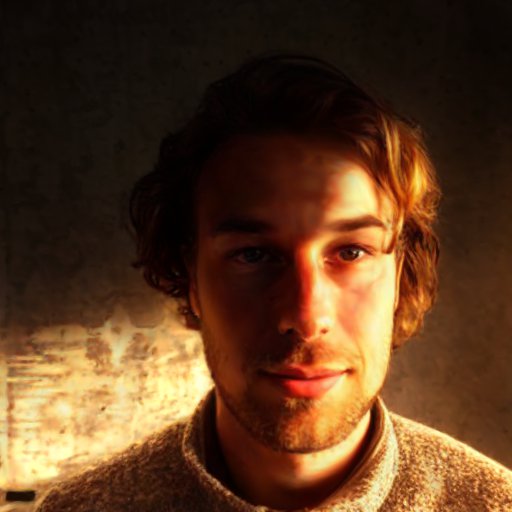} &
\includegraphics[width=0.13\textwidth]{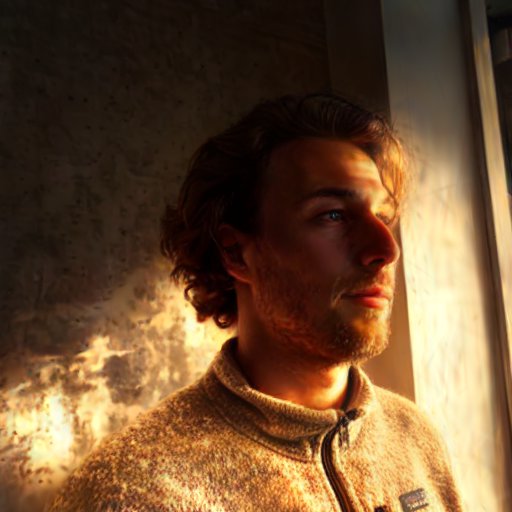} \\

{\centering\footnotesize\textit{"detailed bear statue, marble, twilight, golden autumn forest,"}} &

\includegraphics[width=0.13\textwidth]{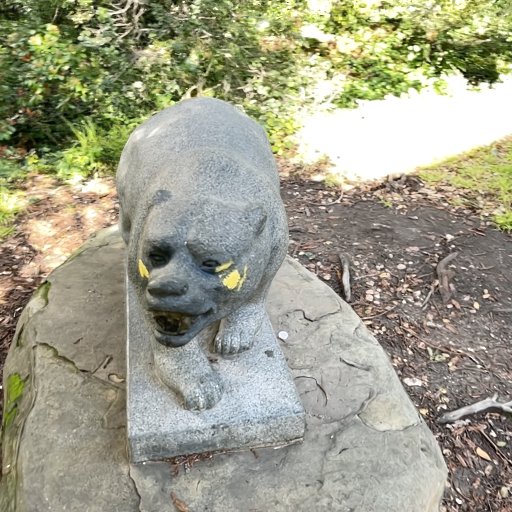} &
\includegraphics[width=0.13\textwidth]{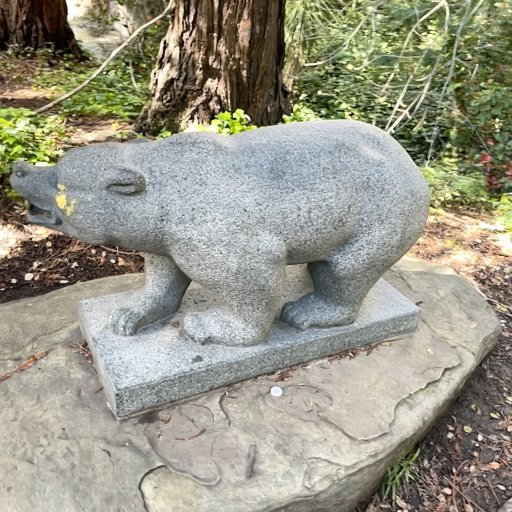} &
\includegraphics[width=0.13\textwidth]{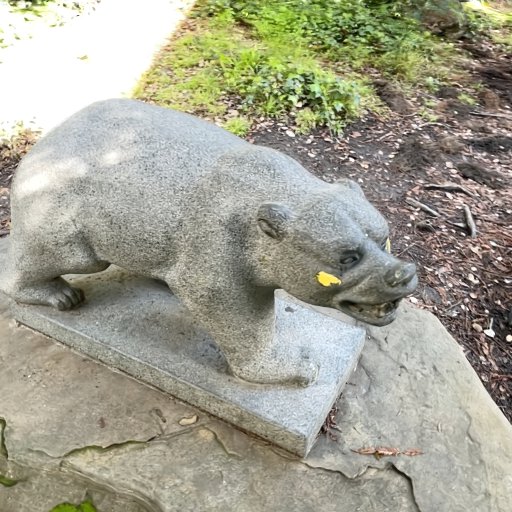} &
\includegraphics[width=0.13\textwidth]{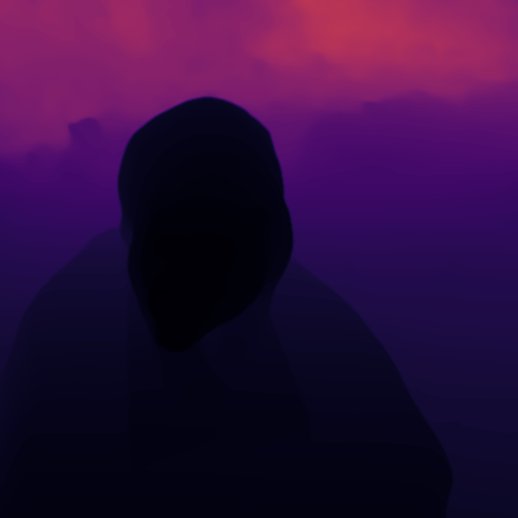} &
\includegraphics[width=0.13\textwidth]{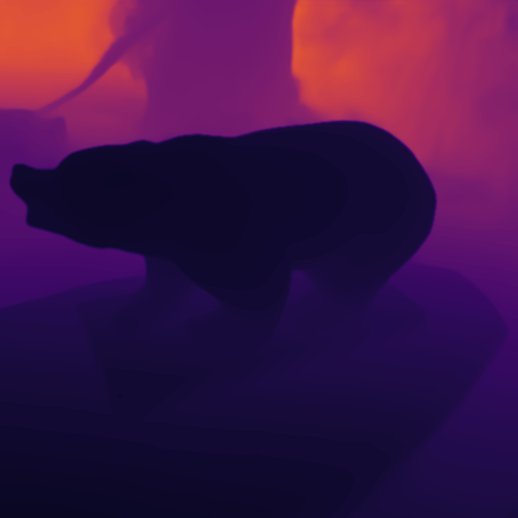} &
\includegraphics[width=0.13\textwidth]{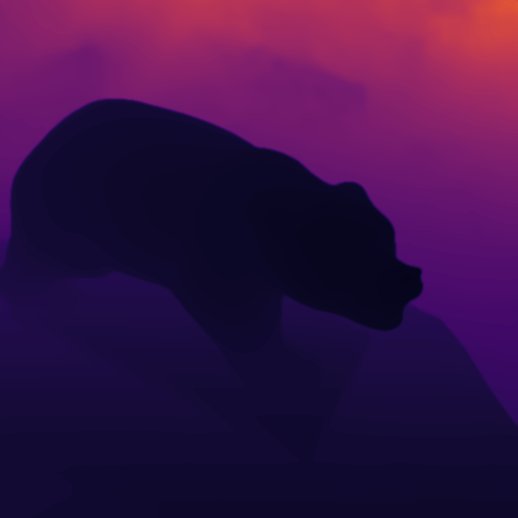} &
\includegraphics[width=0.13\textwidth]{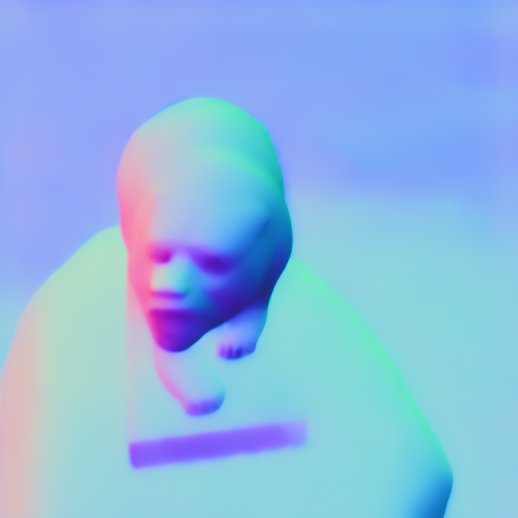} &
\includegraphics[width=0.13\textwidth]{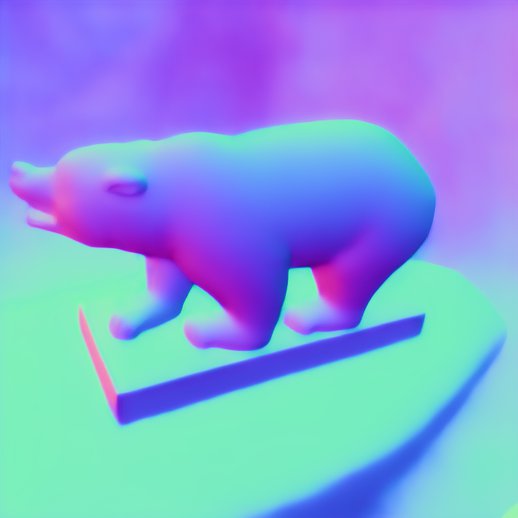} &
\includegraphics[width=0.13\textwidth]{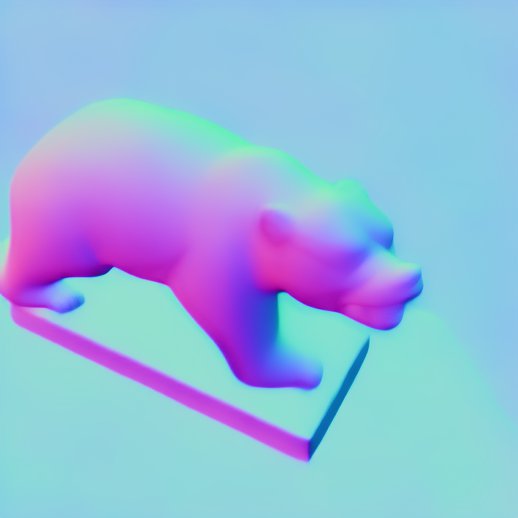} \\

{\centering\footnotesize\textit{"..., sunset glory from \textbf{left} side"}} & 
\includegraphics[width=0.13\textwidth]{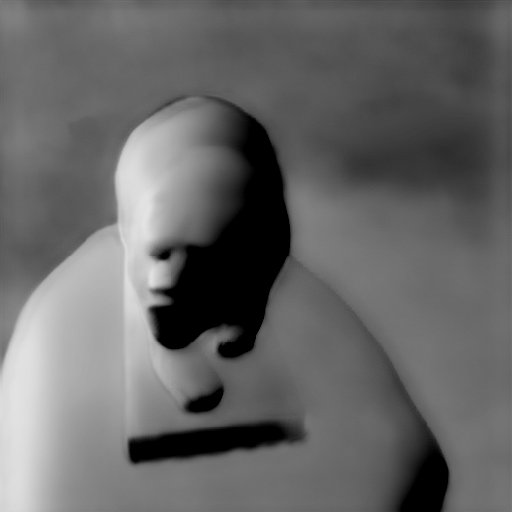} &
\includegraphics[width=0.13\textwidth]{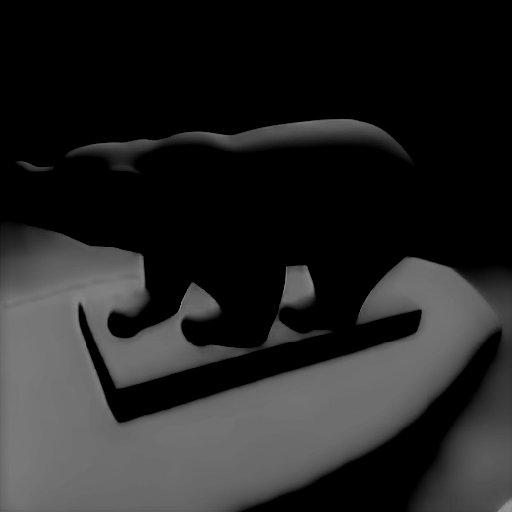} &
\includegraphics[width=0.13\textwidth]{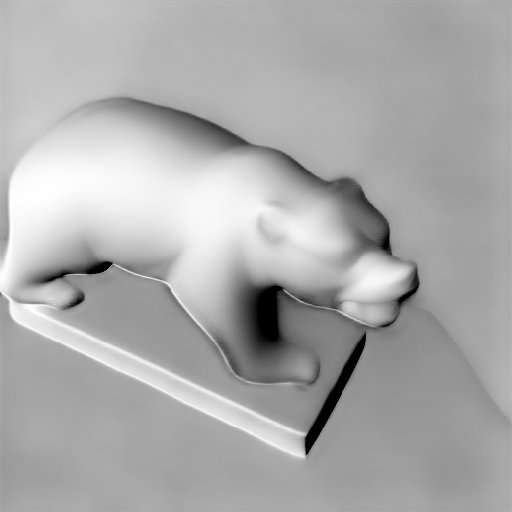} &
\includegraphics[width=0.13\textwidth]{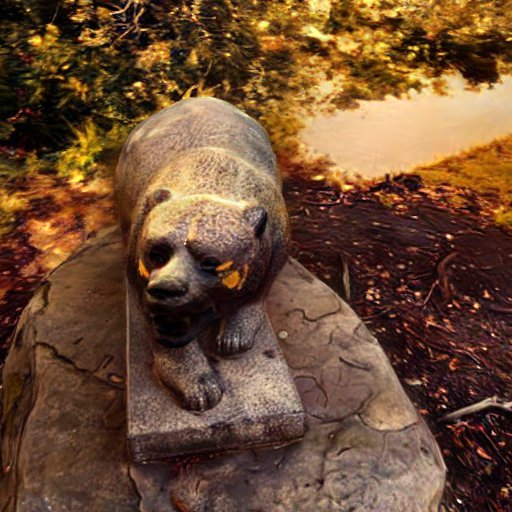} &
\includegraphics[width=0.13\textwidth]{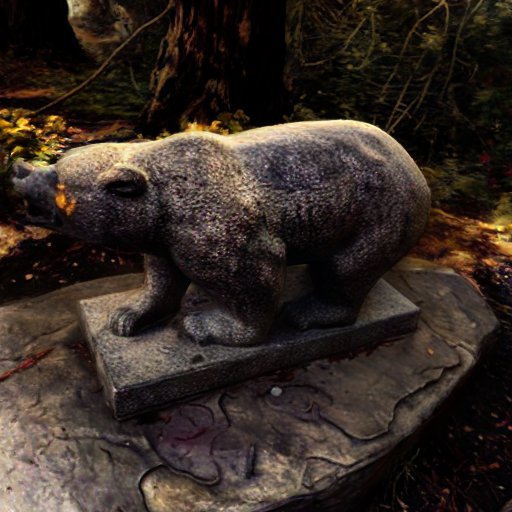} &
\includegraphics[width=0.13\textwidth]{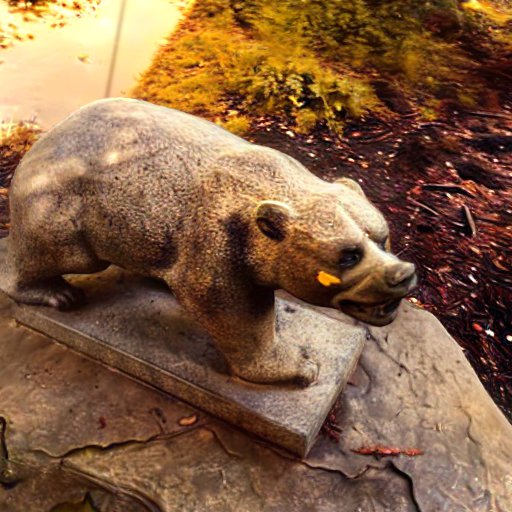} &
\includegraphics[width=0.13\textwidth]{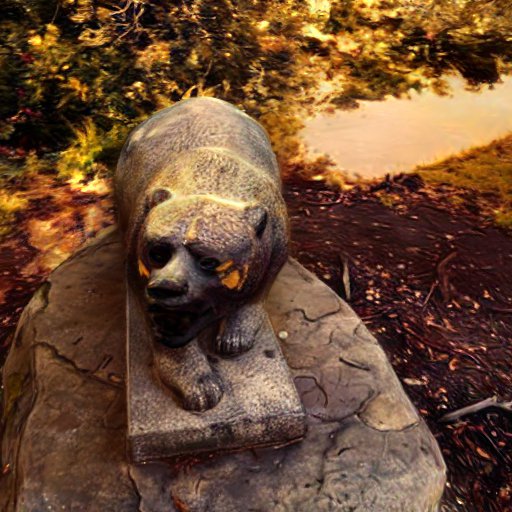} &
\includegraphics[width=0.13\textwidth]{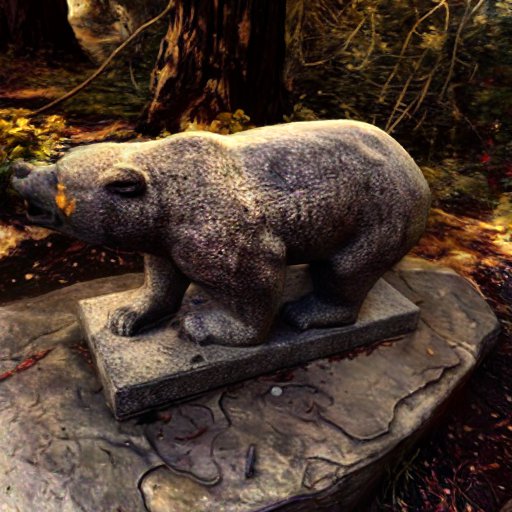} &
\includegraphics[width=0.13\textwidth]{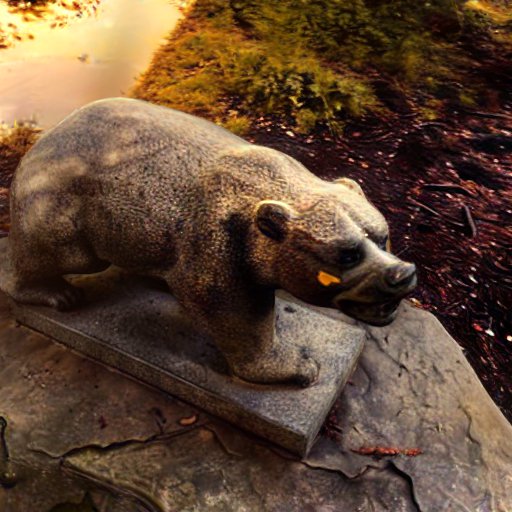} \\

{\centering\footnotesize\textit{"..., sunset glory from \textbf{right} side"}} & 
\includegraphics[width=0.13\textwidth]{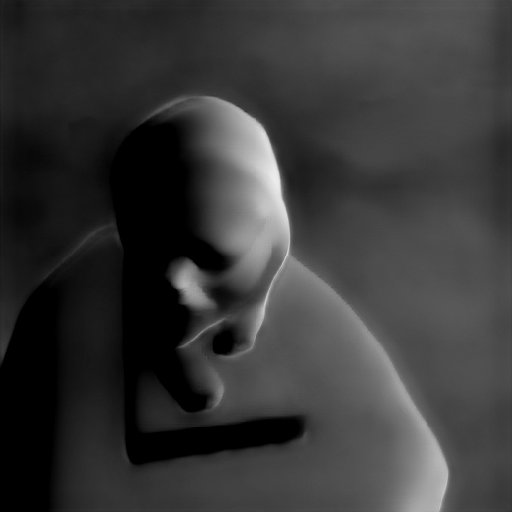} &
\includegraphics[width=0.13\textwidth]{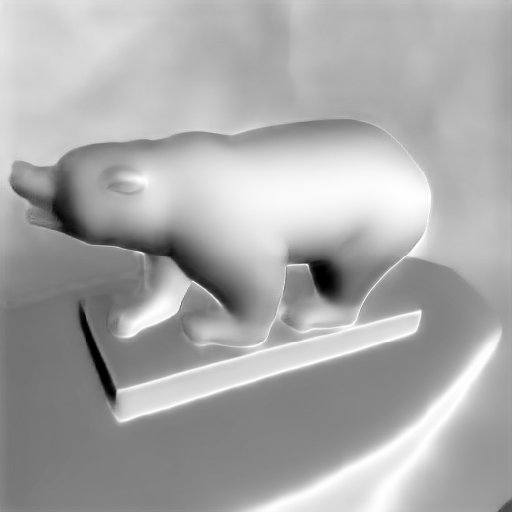} &
\includegraphics[width=0.13\textwidth]{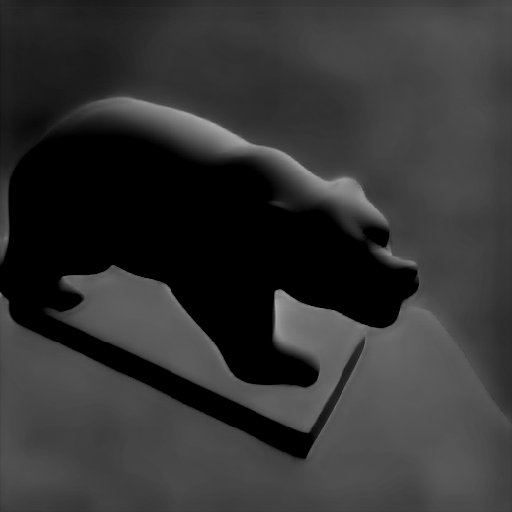} &
\includegraphics[width=0.13\textwidth]{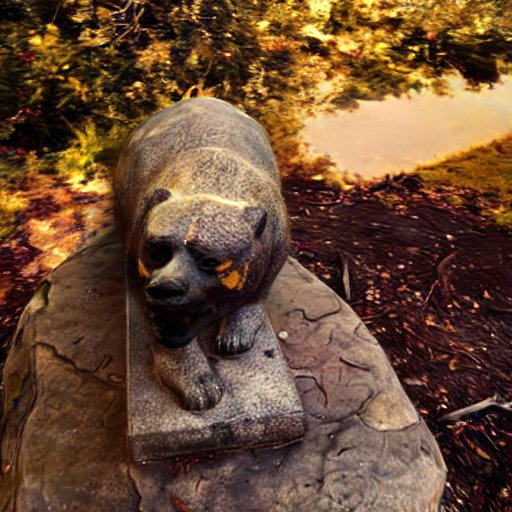} &
\includegraphics[width=0.13\textwidth]{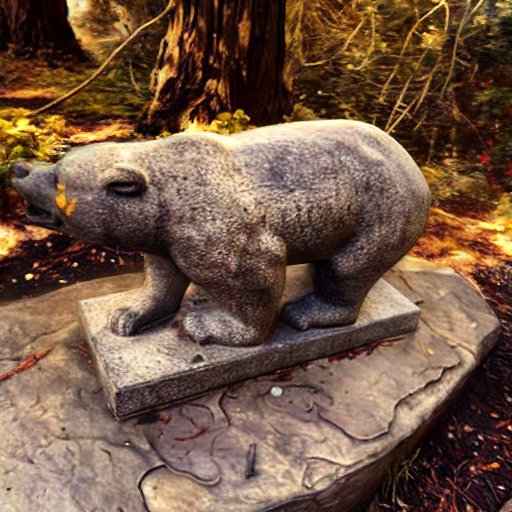} &
\includegraphics[width=0.13\textwidth]{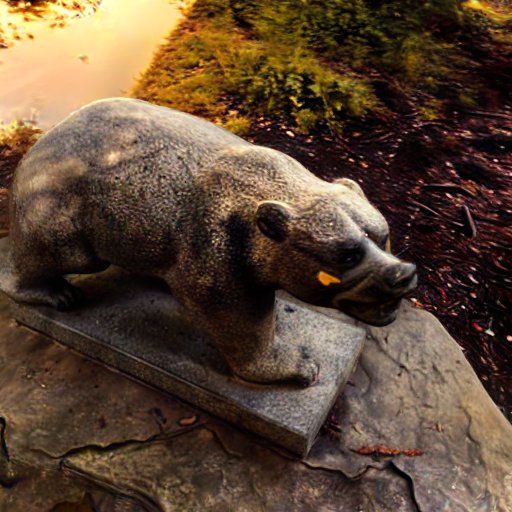} &
\includegraphics[width=0.13\textwidth]{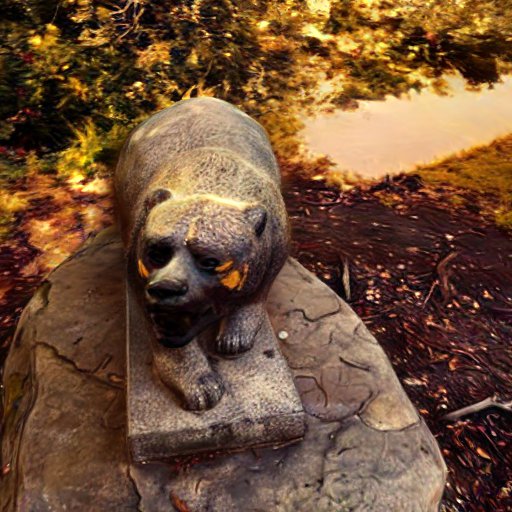} &
\includegraphics[width=0.13\textwidth]{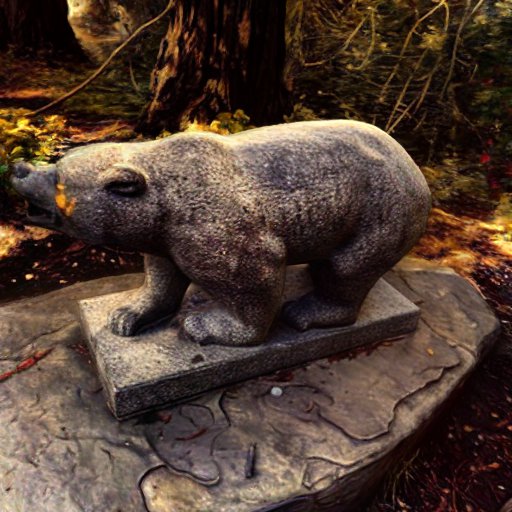} &
\includegraphics[width=0.13\textwidth]{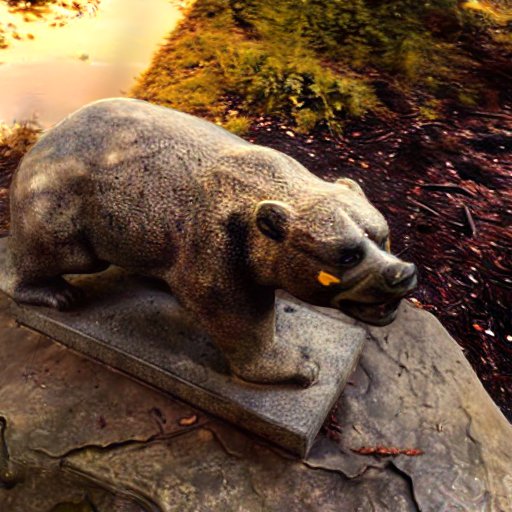} \\

{\centering\footnotesize\textit{"young man, detailed face, natural lighting, outdoor, warm,"}} &

\includegraphics[width=0.13\textwidth]{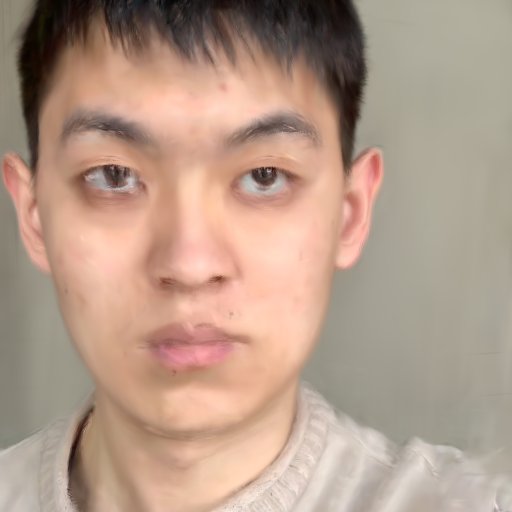} &
\includegraphics[width=0.13\textwidth]{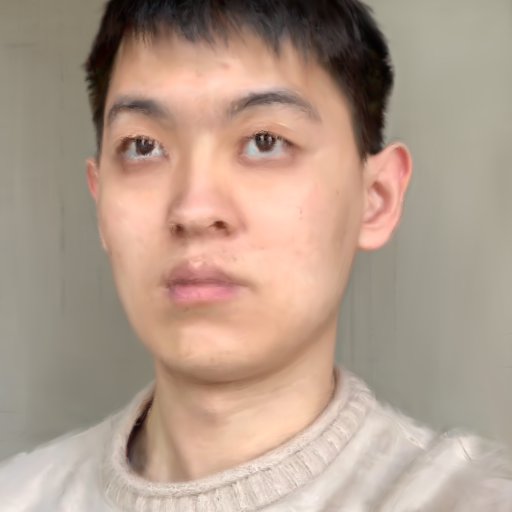} &
\includegraphics[width=0.13\textwidth]{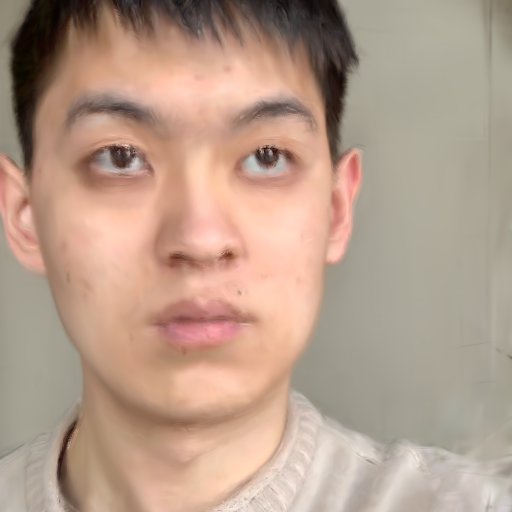} &
\includegraphics[width=0.13\textwidth]{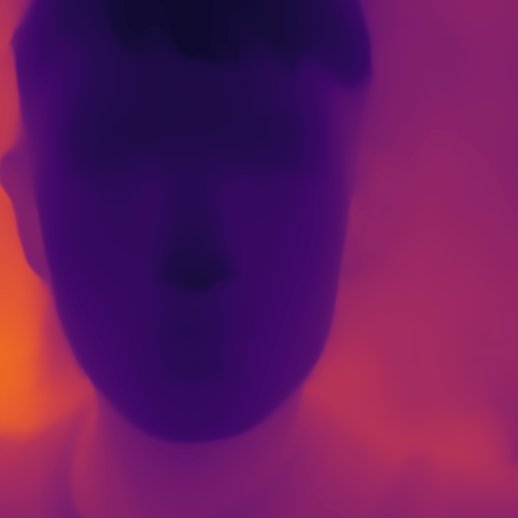} &
\includegraphics[width=0.13\textwidth]{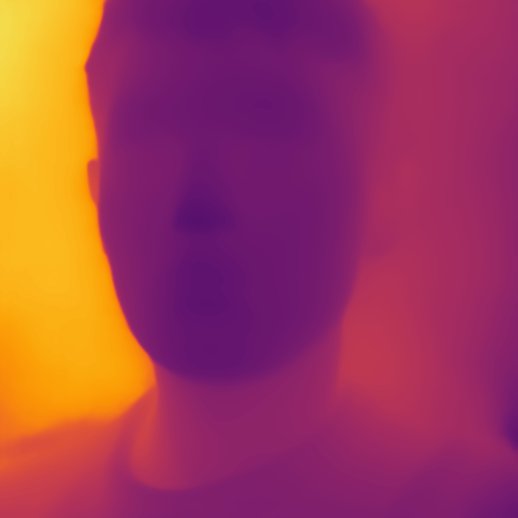} &
\includegraphics[width=0.13\textwidth]{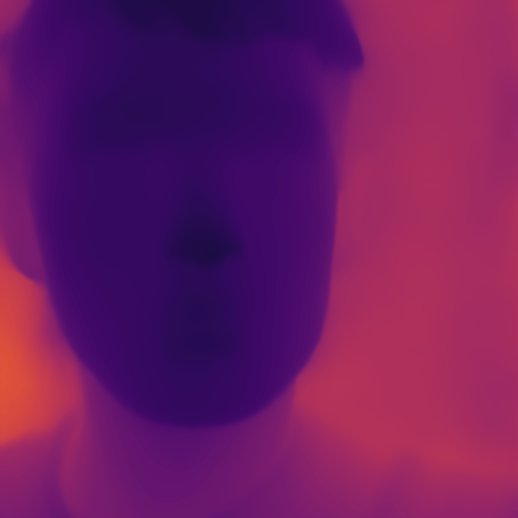} &
\includegraphics[width=0.13\textwidth]{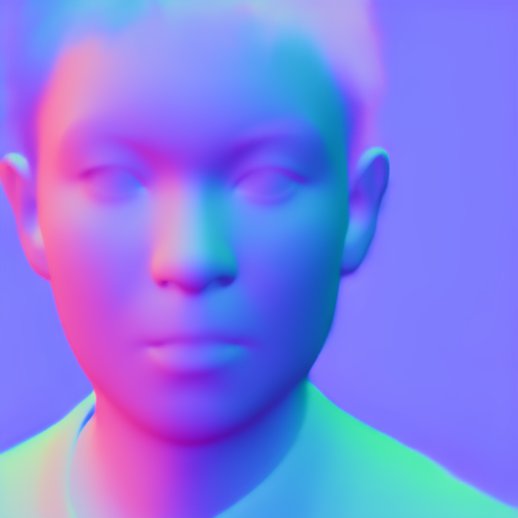} &
\includegraphics[width=0.13\textwidth]{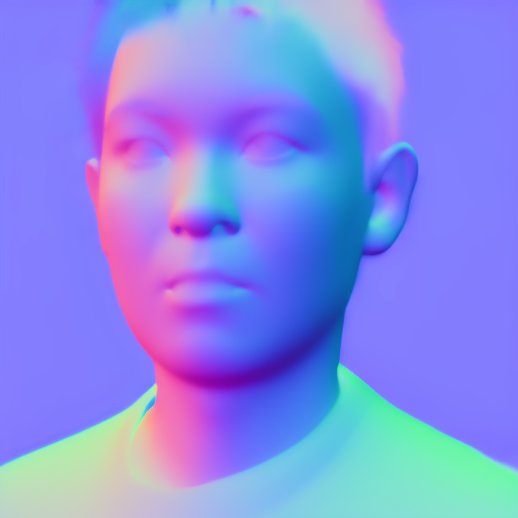} &
\includegraphics[width=0.13\textwidth]{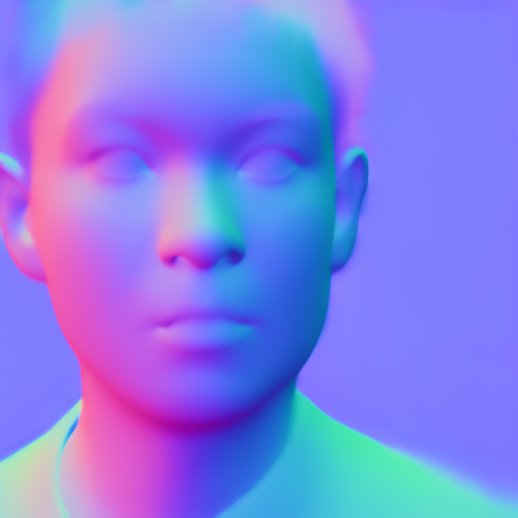} \\

{\centering\footnotesize\textit{"..., light from the \textbf{top}"}} & 
\includegraphics[width=0.13\textwidth]{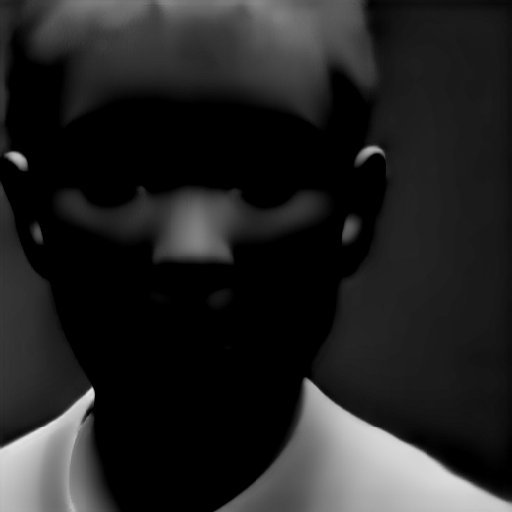} &
\includegraphics[width=0.13\textwidth]{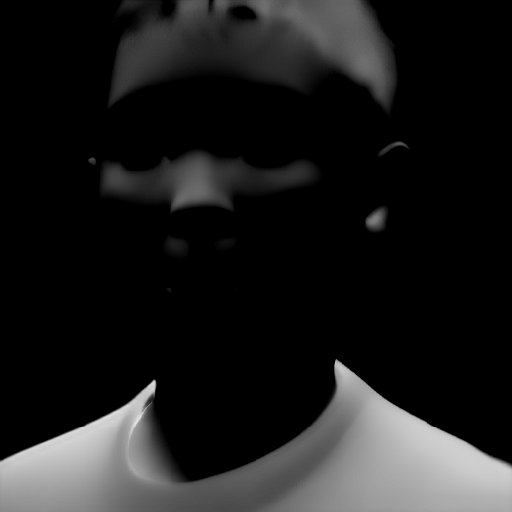} &
\includegraphics[width=0.13\textwidth]{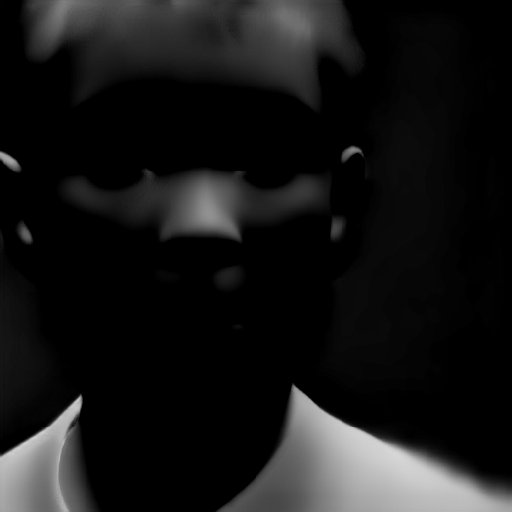} &
\includegraphics[width=0.13\textwidth]{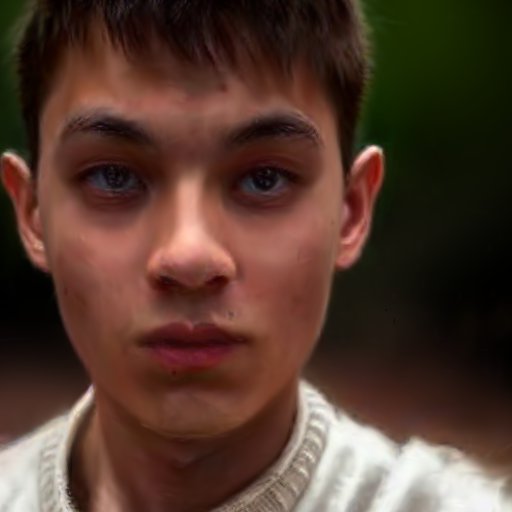} &
\includegraphics[width=0.13\textwidth]{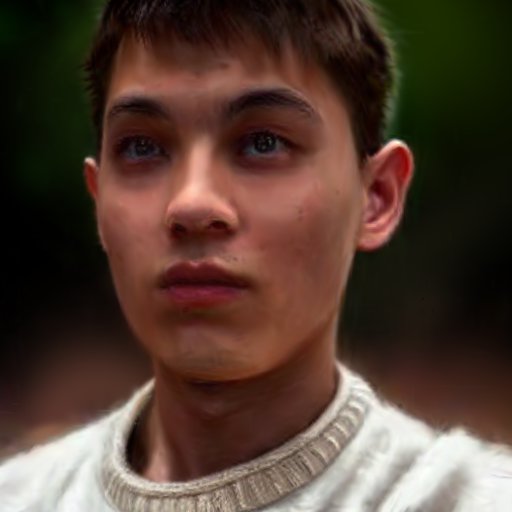} &
\includegraphics[width=0.13\textwidth]{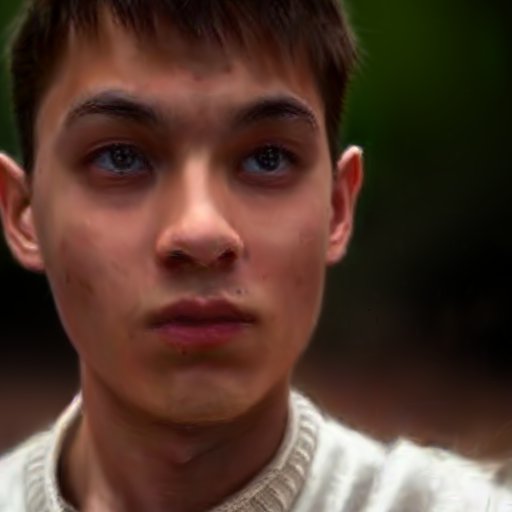} &
\includegraphics[width=0.13\textwidth]{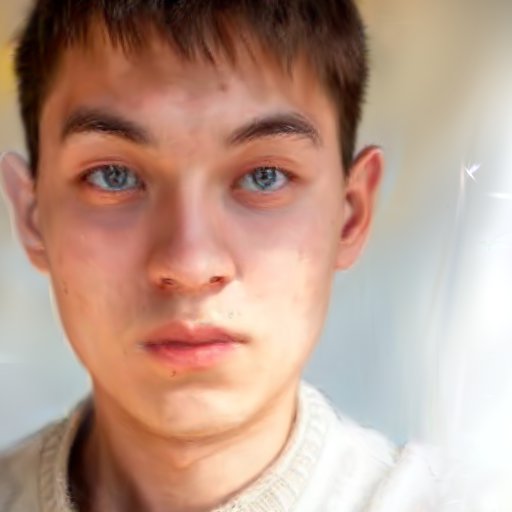} &
\includegraphics[width=0.13\textwidth]{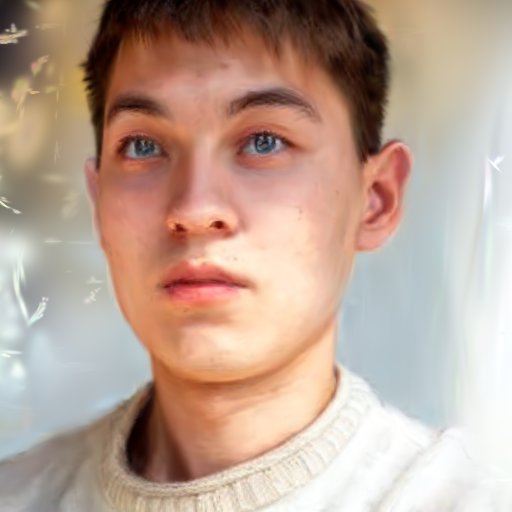} &
\includegraphics[width=0.13\textwidth]{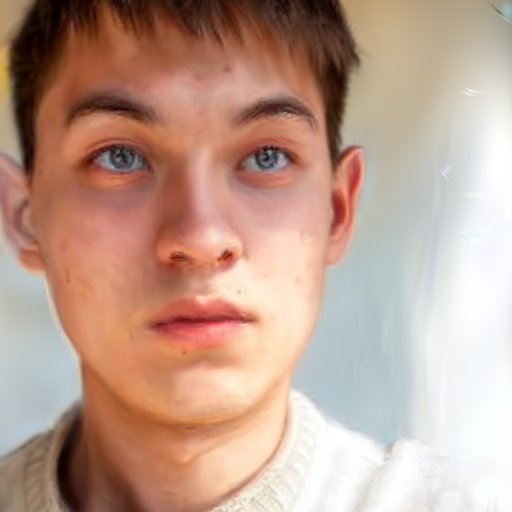} \\

{\centering\footnotesize\textit{"..., light from the \textbf{bottom}"}} & 
\includegraphics[width=0.13\textwidth]{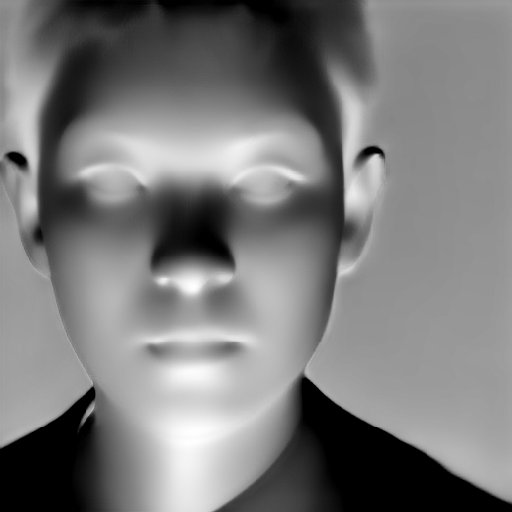} &
\includegraphics[width=0.13\textwidth]{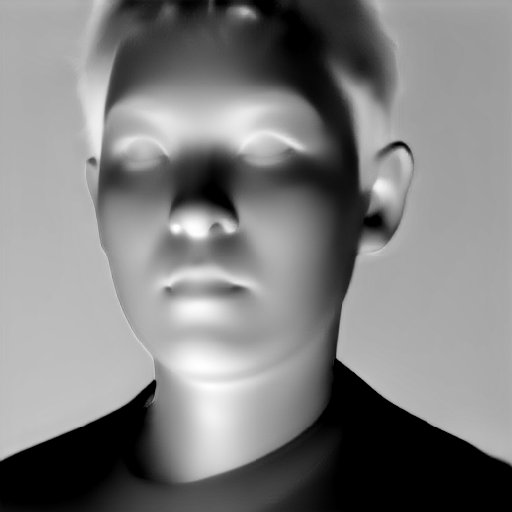} &
\includegraphics[width=0.13\textwidth]{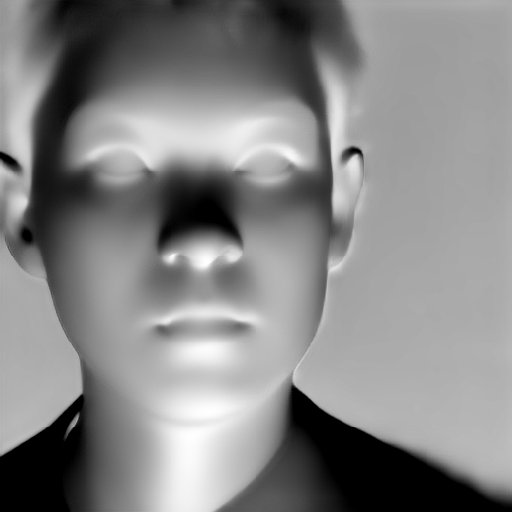} &
\includegraphics[width=0.13\textwidth]{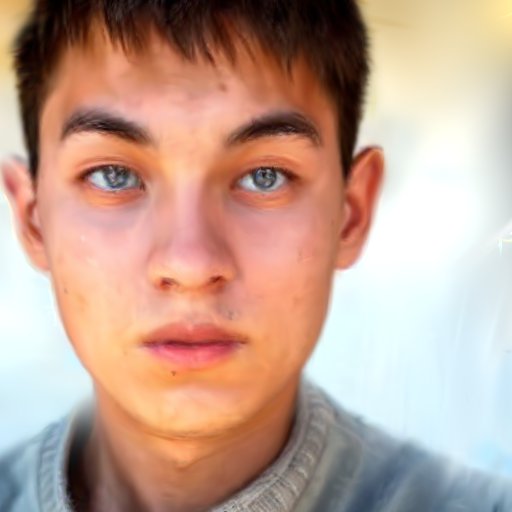} &
\includegraphics[width=0.13\textwidth]{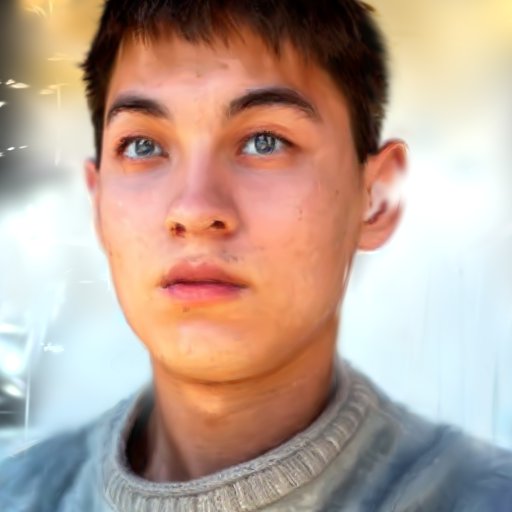} &
\includegraphics[width=0.13\textwidth]{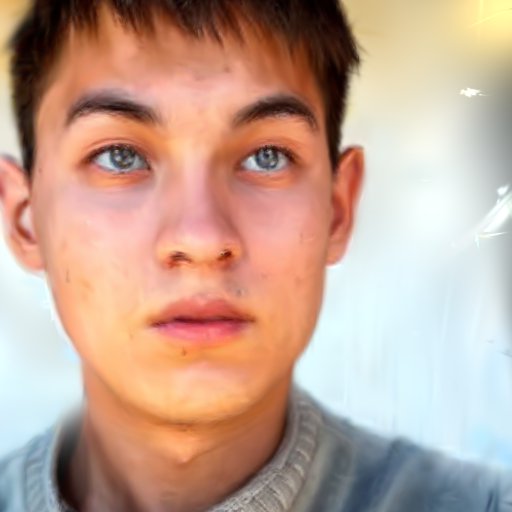} &
\includegraphics[width=0.13\textwidth]{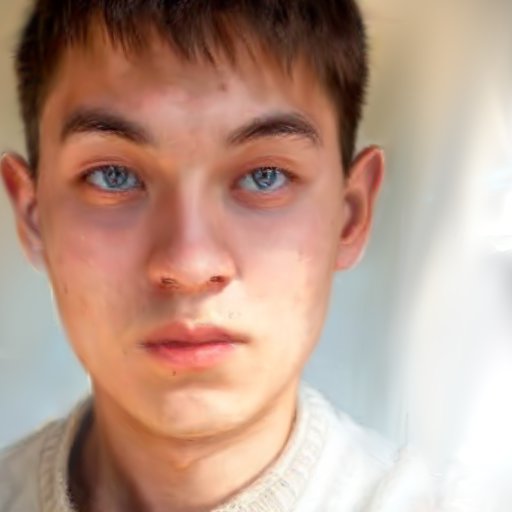} &
\includegraphics[width=0.13\textwidth]{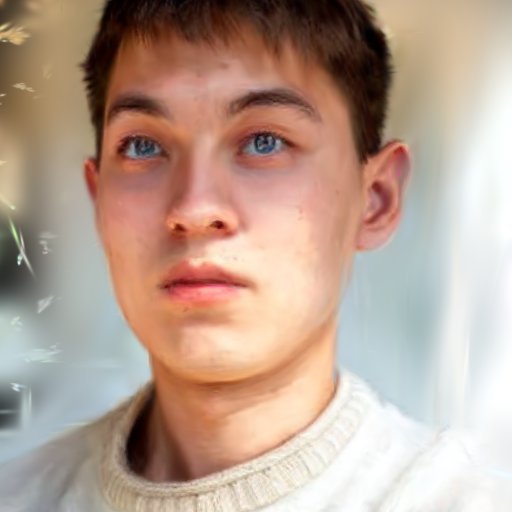} &
\includegraphics[width=0.13\textwidth]{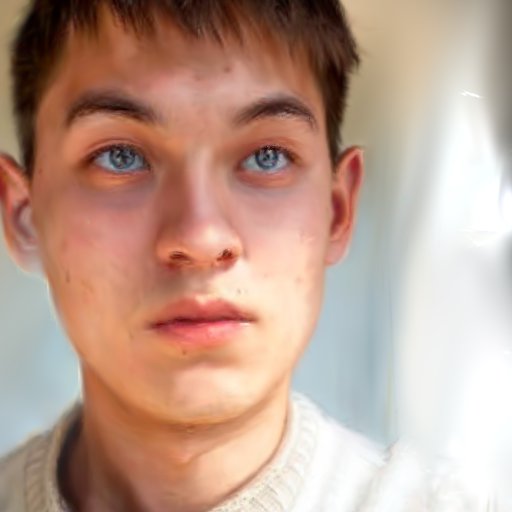} \\

\hline
\end{tabular}
}
\caption{Visualization of PAM component. For each scene, first row shows the input images and common instruction, as well as the depths and normals estimated from pre-trained models. Next, each line represents supplementary instructions for different light directions, sequentially displaying the corresponding initial latents and the relighting results of whether the PAM component is used.}
\label{fig:pam_ablation}
\end{figure}

\textbf{Effectiveness of Position-Align Module.} We present the relighting results of different methods when the input instructions include lighting position information, as shown in Fig.~\ref{fig:pam_ablation}. Experimental results demonstrate that, with the help of the PAM module, our method can more accurately capture the lighting direction described in the editing instruction, thereby producing relighting results that are more faithful to the user’s intent.

\begin{figure}[!t]  
    \centering
    \includegraphics[width=0.9\linewidth]{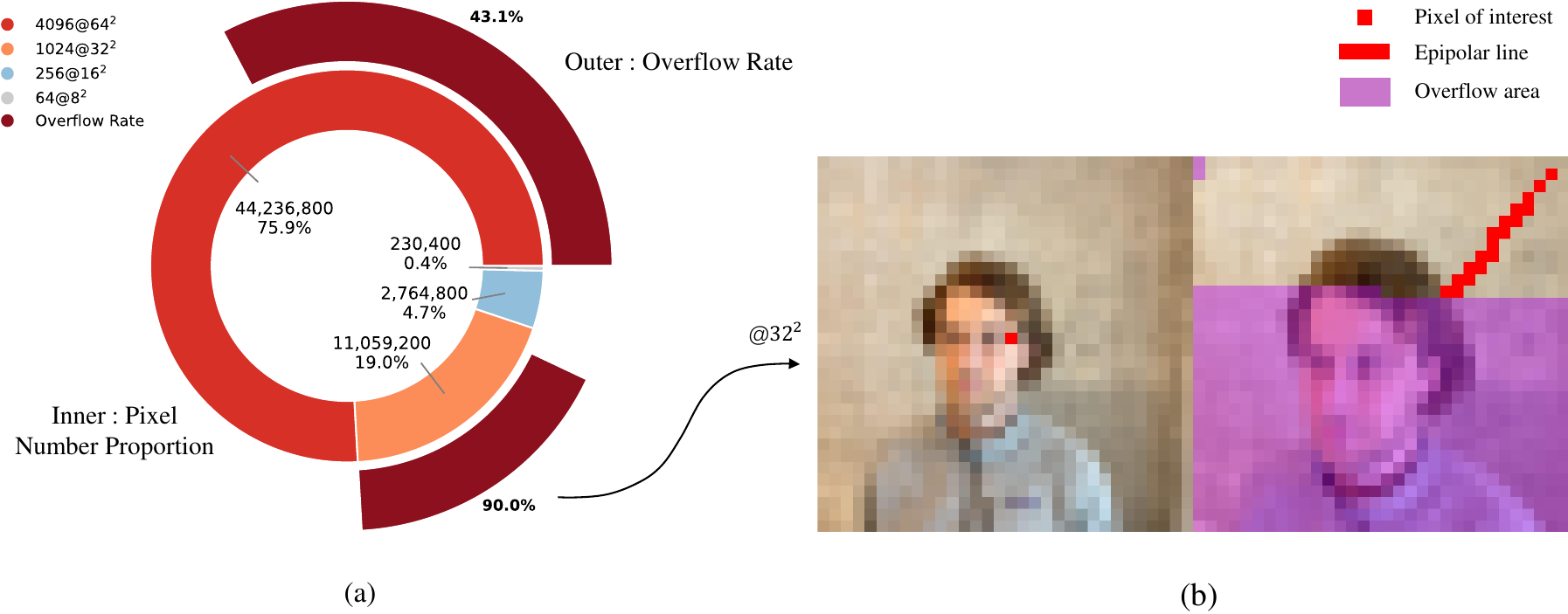}  
    \caption{(a) Statistics of overflow occurrences across different resolutions (UNet layers); (b) Visualization of how overflow disrupts the epipolar constraints (at $32^2$ resolution)}
    \label{fig:epipolar_invalid}  
\end{figure}

\textbf{Fundamental Matrix Normalization.} In practice, we found that the epipolar constraint frequently fails during UNet inference, which is typically caused by numerical overflow. To further investigate the cause, we take the face scene from IN2N as an example. Fig.~\ref{fig:epipolar_invalid} (a) shows the proportion of pixels across different UNet layers (i.e., different resolutions) and the ratio of those affected by overflow. We observe that nearly half of the pixels experience numerical overflow, severely disrupting the process of finding the maximum value along the epipolar line. As shown in Fig.~\ref{fig:epipolar_invalid} (b), regions with numerical instability cause the search range of the epipolar constraint to shrink, leading to suboptimal matches, and in more severe cases, the matching process may degenerate into global matching. Fortunately, we discovered that the overflow issue originates from the lack of normalization in the fundamental matrix computed by PyTorch, resulting in excessively large values (up to the order of $10^6$) in the matrix $F$. Tab.~\ref{tab:ablation_epipolar} reports the average number of UNet layers experiencing overflow per inference before and after normalization, as well as the relighting results in terms of PSNR, SSIM and LPIPS. The experiments demonstrate that our improved normalized fundamental matrix significantly enhances the numerical stability of the inference process and improves the multi-view consistency of the editing results.

\begin{table}[ht]
\centering
\caption{Ablation on epipolar constraint normalization of fundamental matrix. We evaluate on face scene from IN2N dataset.}
\label{tab:ablation_epipolar}
\begin{tabular}{l|cccc}
\toprule
Method & Avg. Overflow Rate↓ & PSNR↑ & SSIM↑ &LPIPS↓\\
\midrule
w/o normalization & 49.82\% & 20.81 & 0.7312 & 0.1621 \\
w/ normalization & \textbf{0.00\%} & \textbf{21.55} & \textbf{0.7328} & \textbf{0.1599}\\
\bottomrule
\end{tabular}
\end{table}

These ablation studies collectively demonstrate the importance of our proposed modules and design choices in achieving semantically faithful and geometrically coherent 3D scene edits.

\section{Limitations and Future Work}
\label{sec:limit}

There are several limitations to GS-Light. First, our reliance on off-the-shelf geometry / normal estimators means that if depth or normal maps are inaccurate (e.g. due to occlusions, reflective or transparent surfaces), the lighting fusion and shadow estimation may fail or artifacts may arise. Second, the projection back into the 3DGS representation may leave uncovered regions (views or surfaces not well seen in images), leading to inconsistency or blurring. Third, strongly specular or anisotropic materials are difficult to handle in inference only pipelines without BRDF fitting.

Future work could include integrating lightweight material priors or BRDF estimation; extending LVLM prior extraction to more complex lighting (multiple lights, colored ambient, etc.); better handling of occlusion and shadows via differentiable visibility; possibly allowing optional per-scene fine-tuning when higher fidelity is required.

\section{Conclusion}
\label{sec:concl}

We presented GS-Light, a training-efficient, text-guided, position-aware method for scene relighting in Gaussian Splatting representations. By combining prompt-derived lighting priors and view-consistency constraints, our pipeline generates multi-view coherent relit images and relit 3D scenes, especially in lighting directions, outperforming several baselines while operating purely at inference time. We believe GS-Light provides a useful step toward more accessible, controllable, and consistent relighting for 3D content.


\bibliographystyle{elsarticle-num}
\bibliography{references}

@article{nerf,
  title={Nerf: Representing scenes as neural radiance fields for view synthesis},
  author={Mildenhall, Ben and Srinivasan, Pratul P and Tancik, Matthew and Barron, Jonathan T and Ramamoorthi, Ravi and Ng, Ren},
  journal={Communications of the ACM},
  volume={65},
  number={1},
  pages={99--106},
  year={2021},
  publisher={ACM New York, NY, USA}
}

@article{3dgs,
  title={3D Gaussian splatting for real-time radiance field rendering.},
  author={Kerbl, Bernhard and Kopanas, Georgios and Leimk{\"u}hler, Thomas and Drettakis, George},
  journal={ACM Trans. Graph.},
  volume={42},
  number={4},
  pages={139--1},
  year={2023}
}

@misc{gpt5,
  title        = {GPT-5 System Card},
  author       = {{OpenAI}},
  year         = {2025},
  howpublished = {\url{https://cdn.openai.com/gpt-5-system-card.pdf}},
}

@inproceedings{sfm,
  title={Structure-from-motion revisited},
  author={Schonberger, Johannes L and Frahm, Jan-Michael},
  booktitle={Proceedings of the IEEE conference on computer vision and pattern recognition},
  pages={4104--4113},
  year={2016}
}

@inproceedings{vbench,
  title={Vbench: Comprehensive benchmark suite for video generative models},
  author={Huang, Ziqi and He, Yinan and Yu, Jiashuo and Zhang, Fan and Si, Chenyang and Jiang, Yuming and Zhang, Yuanhan and Wu, Tianxing and Jin, Qingyang and Chanpaisit, Nattapol and others},
  booktitle={Proceedings of the IEEE/CVF Conference on Computer Vision and Pattern Recognition},
  pages={21807--21818},
  year={2024}
}

@inproceedings{mip360,
  title={Mip-nerf 360: Unbounded anti-aliased neural radiance fields},
  author={Barron, Jonathan T and Mildenhall, Ben and Verbin, Dor and Srinivasan, Pratul P and Hedman, Peter},
  booktitle={Proceedings of the IEEE/CVF conference on computer vision and pattern recognition},
  pages={5470--5479},
  year={2022}
}

@inproceedings{scannetpp,
  title={Scannet++: A high-fidelity dataset of 3d indoor scenes},
  author={Yeshwanth, Chandan and Liu, Yueh-Cheng and Nie{\ss}ner, Matthias and Dai, Angela},
  booktitle={Proceedings of the IEEE/CVF International Conference on Computer Vision},
  pages={12--22},
  year={2023}
}

@article{qwen25vl,
  title={Qwen2. 5-vl technical report},
  author={Bai, Shuai and Chen, Keqin and Liu, Xuejing and Wang, Jialin and Ge, Wenbin and Song, Sibo and Dang, Kai and Wang, Peng and Wang, Shijie and Tang, Jun and others},
  journal={arXiv preprint arXiv:2502.13923},
  year={2025}
}

@inproceedings{groundingdino,
  title={Grounding dino: Marrying dino with grounded pre-training for open-set object detection},
  author={Liu, Shilong and Zeng, Zhaoyang and Ren, Tianhe and Li, Feng and Zhang, Hao and Yang, Jie and Jiang, Qing and Li, Chunyuan and Yang, Jianwei and Su, Hang and others},
  booktitle={European conference on computer vision},
  pages={38--55},
  year={2024},
  organization={Springer}
}

@inproceedings{vggt,
  title={Vggt: Visual geometry grounded transformer},
  author={Wang, Jianyuan and Chen, Minghao and Karaev, Nikita and Vedaldi, Andrea and Rupprecht, Christian and Novotny, David},
  booktitle={Proceedings of the Computer Vision and Pattern Recognition Conference},
  pages={5294--5306},
  year={2025}
}

@article{sam2,
  title={SAM 2: Segment Anything in Images and Videos},
  author={Ravi, Nikhila and Gabeur, Valentin and Hu, Yuan-Ting and Hu, Ronghang and Ryali, Chaitanya and Ma, Tengyu and Khedr, Haitham and R{\"a}dle, Roman and Rolland, Chloe and Gustafson, Laura and Mintun, Eric and Pan, Junting and Alwala, Kalyan Vasudev and Carion, Nicolas and Wu, Chao-Yuan and Girshick, Ross and Doll{\'a}r, Piotr and Feichtenhofer, Christoph},
  journal={arXiv preprint arXiv:2408.00714},
  url={https://arxiv.org/abs/2408.00714},
  year={2024}
}

@incollection{phong,
  title={Illumination for computer generated pictures},
  author={Phong, Bui Tuong},
  booktitle={Seminal graphics: pioneering efforts that shaped the field},
  pages={95--101},
  year={1998}
}

@inproceedings{lightstage,
  title={Acquiring the reflectance field of a human face},
  author={Debevec, Paul and Hawkins, Tim and Tchou, Chris and Duiker, Haarm-Pieter and Sarokin, Westley and Sagar, Mark},
  booktitle={Proceedings of the 27th annual conference on Computer graphics and interactive techniques},
  pages={145--156},
  year={2000}
}

@article{deep_iamge_based_relight,
  title={Deep image-based relighting from optimal sparse samples},
  author={Xu, Zexiang and Sunkavalli, Kalyan and Hadap, Sunil and Ramamoorthi, Ravi},
  journal={ACM Transactions on Graphics (ToG)},
  volume={37},
  number={4},
  pages={1--13},
  year={2018},
  publisher={ACM New York, NY, USA}
}

@article{single_portrait_relight,
  title={Single image portrait relighting.},
  author={Sun, Tiancheng and Barron, Jonathan T and Tsai, Yun-Ta and Xu, Zexiang and Yu, Xueming and Fyffe, Graham and Rhemann, Christoph and Busch, Jay and Debevec, Paul E and Ramamoorthi, Ravi},
  journal={ACM Trans. Graph.},
  volume={38},
  number={4},
  pages={79--1},
  year={2019}
}

@inproceedings{physics_guided_relight,
  title={Learning physics-guided face relighting under directional light},
  author={Nestmeyer, Thomas and Lalonde, Jean-Fran{\c{c}}ois and Matthews, Iain and Lehrmann, Andreas},
  booktitle={Proceedings of the IEEE/CVF Conference on Computer Vision and Pattern Recognition},
  pages={5124--5133},
  year={2020}
}

@inproceedings{switchlight,
  title={Switchlight: Co-design of physics-driven architecture and pre-training framework for human portrait relighting},
  author={Kim, Hoon and Jang, Minje and Yoon, Wonjun and Lee, Jisoo and Na, Donghyun and Woo, Sanghyun},
  booktitle={Proceedings of the IEEE/CVF Conference on Computer Vision and Pattern Recognition},
  pages={25096--25106},
  year={2024}
}

@article{portrait_transfer,
  title={Portrait lighting transfer using a mass transport approach},
  author={Shu, Zhixin and Hadap, Sunil and Shechtman, Eli and Sunkavalli, Kalyan and Paris, Sylvain and Samaras, Dimitris},
  journal={ACM Transactions on Graphics (TOG)},
  volume={36},
  number={4},
  pages={1},
  year={2017},
  publisher={ACM New York, NY, USA}
}

@inproceedings{sfsnet,
  title={Sfsnet: Learning shape, reflectance and illuminance of facesin the wild'},
  author={Sengupta, Soumyadip and Kanazawa, Angjoo and Castillo, Carlos D and Jacobs, David W},
  booktitle={Proceedings of the IEEE conference on computer vision and pattern recognition},
  pages={6296--6305},
  year={2018}
}

@inproceedings{iclight,
  title={Scaling in-the-wild training for diffusion-based illumination harmonization and editing by imposing consistent light transport},
  author={Zhang, Lvmin and Rao, Anyi and Agrawala, Maneesh},
  booktitle={The Thirteenth International Conference on Learning Representations},
  year={2025}
}

@inproceedings{relight_harmoni,
  title={Relightful harmonization: Lighting-aware portrait background replacement},
  author={Ren, Mengwei and Xiong, Wei and Yoon, Jae Shin and Shu, Zhixin and Zhang, Jianming and Jung, HyunJoon and Gerig, Guido and Zhang, He},
  booktitle={Proceedings of the IEEE/CVF Conference on Computer Vision and Pattern Recognition},
  pages={6452--6462},
  year={2024}
}

@article{relightvid,
  title={RelightVid: Temporal-consistent diffusion model for video relighting},
  author={Fang, Ye and Sun, Zeyi and Zhang, Shangzhan and Wu, Tong and Xu, Yinghao and Zhang, Pan and Wang, Jiaqi and Wetzstein, Gordon and Lin, Dahua},
  journal={arXiv preprint arXiv:2501.16330},
  year={2025}
}

@article{lumen,
  title={Lumen: Consistent Video Relighting and Harmonious Background Replacement with Video Generative Models},
  author={Zeng, Jianshu and Liu, Yuxuan and Feng, Yutong and Miao, Chenxuan and Gao, Zixiang and Qu, Jiwang and Zhang, Jianzhang and Wang, Bin and Yuan, Kun},
  journal={arXiv preprint arXiv:2508.12945},
  year={2025}
}

@article{dm,
  title={Diffusion models beat gans on image synthesis},
  author={Dhariwal, Prafulla and Nichol, Alexander},
  journal={Advances in neural information processing systems},
  volume={34},
  pages={8780--8794},
  year={2021}
}

@inproceedings{ldm,
  title={High-resolution image synthesis with latent diffusion models},
  author={Rombach, Robin and Blattmann, Andreas and Lorenz, Dominik and Esser, Patrick and Ommer, Bj{\"o}rn},
  booktitle={Proceedings of the IEEE/CVF conference on computer vision and pattern recognition},
  pages={10684--10695},
  year={2022}
}

@article{ddim,
  title={Denoising diffusion implicit models},
  author={Song, Jiaming and Meng, Chenlin and Ermon, Stefano},
  journal={arXiv preprint arXiv:2010.02502},
  year={2020}
}

@inproceedings{dit,
  title={Scalable diffusion models with transformers},
  author={Peebles, William and Xie, Saining},
  booktitle={Proceedings of the IEEE/CVF international conference on computer vision},
  pages={4195--4205},
  year={2023}
}

@inproceedings{diffusionclip,
  title={Diffusionclip: Text-guided diffusion models for robust image manipulation},
  author={Kim, Gwanghyun and Kwon, Taesung and Ye, Jong Chul},
  booktitle={Proceedings of the IEEE/CVF conference on computer vision and pattern recognition},
  pages={2426--2435},
  year={2022}
}

@inproceedings{instructpix2pix,
  title={Instructpix2pix: Learning to follow image editing instructions},
  author={Brooks, Tim and Holynski, Aleksander and Efros, Alexei A},
  booktitle={Proceedings of the IEEE/CVF conference on computer vision and pattern recognition},
  pages={18392--18402},
  year={2023}
}

@inproceedings{null_text,
  title={Null-text inversion for editing real images using guided diffusion models},
  author={Mokady, Ron and Hertz, Amir and Aberman, Kfir and Pritch, Yael and Cohen-Or, Daniel},
  booktitle={Proceedings of the IEEE/CVF conference on computer vision and pattern recognition},
  pages={6038--6047},
  year={2023}
}

@inproceedings{style_align,
  title={Style aligned image generation via shared attention},
  author={Hertz, Amir and Voynov, Andrey and Fruchter, Shlomi and Cohen-Or, Daniel},
  booktitle={Proceedings of the IEEE/CVF Conference on Computer Vision and Pattern Recognition},
  pages={4775--4785},
  year={2024}
}

@inproceedings{deadiff,
  title={Deadiff: An efficient stylization diffusion model with disentangled representations},
  author={Qi, Tianhao and Fang, Shancheng and Wu, Yanze and Xie, Hongtao and Liu, Jiawei and Chen, Lang and He, Qian and Zhang, Yongdong},
  booktitle={Proceedings of the IEEE/CVF conference on computer vision and pattern recognition},
  pages={8693--8702},
  year={2024}
}

@article{instantstyle,
  title={Instantstyle: Free lunch towards style-preserving in text-to-image generation},
  author={Wang, Haofan and Spinelli, Matteo and Wang, Qixun and Bai, Xu and Qin, Zekui and Chen, Anthony},
  journal={arXiv preprint arXiv:2404.02733},
  year={2024}
}

@article{instantstyle_plus,
  title={Instantstyle-plus: Style transfer with content-preserving in text-to-image generation},
  author={Wang, Haofan and Xing, Peng and Huang, Renyuan and Ai, Hao and Wang, Qixun and Bai, Xu},
  journal={arXiv preprint arXiv:2407.00788},
  year={2024}
}

@article{stablenormal,
  title={Stablenormal: Reducing diffusion variance for stable and sharp normal},
  author={Ye, Chongjie and Qiu, Lingteng and Gu, Xiaodong and Zuo, Qi and Wu, Yushuang and Dong, Zilong and Bo, Liefeng and Xiu, Yuliang and Han, Xiaoguang},
  journal={ACM Transactions on Graphics (TOG)},
  volume={43},
  number={6},
  pages={1--18},
  year={2024},
  publisher={ACM New York, NY, USA}
}

@inproceedings{diffusion_depth,
  title={Repurposing diffusion-based image generators for monocular depth estimation},
  author={Ke, Bingxin and Obukhov, Anton and Huang, Shengyu and Metzger, Nando and Daudt, Rodrigo Caye and Schindler, Konrad},
  booktitle={Proceedings of the IEEE/CVF conference on computer vision and pattern recognition},
  pages={9492--9502},
  year={2024}
}

@article{instant3d,
  title={Instant3d: Fast text-to-3d with sparse-view generation and large reconstruction model},
  author={Li, Jiahao and Tan, Hao and Zhang, Kai and Xu, Zexiang and Luan, Fujun and Xu, Yinghao and Hong, Yicong and Sunkavalli, Kalyan and Shakhnarovich, Greg and Bi, Sai},
  journal={arXiv preprint arXiv:2311.06214},
  year={2023}
}

@article{reference_based,
  title={Reference-based Controllable Scene Stylization with Gaussian Splatting},
  author={Mei, Yiqun and Xu, Jiacong and Patel, Vishal M},
  journal={arXiv preprint arXiv:2407.07220},
  year={2024}
}

@inproceedings{arf,
  title={Arf: Artistic radiance fields},
  author={Zhang, Kai and Kolkin, Nick and Bi, Sai and Luan, Fujun and Xu, Zexiang and Shechtman, Eli and Snavely, Noah},
  booktitle={European Conference on Computer Vision},
  pages={717--733},
  year={2022},
  organization={Springer}
}

@inproceedings{style_transfer,
  title={Image style transfer using convolutional neural networks},
  author={Gatys, Leon A and Ecker, Alexander S and Bethge, Matthias},
  booktitle={Proceedings of the IEEE conference on computer vision and pattern recognition},
  pages={2414--2423},
  year={2016}
}

@inproceedings{gstyle,
  title={G-Style: Stylized Gaussian Splatting},
  author={Kov{\'a}cs, {\'A}ron Samuel and Hermosilla, Pedro and Raidou, Renata G},
  booktitle={Computer Graphics Forum},
  volume={43},
  pages={e15259},
  year={2024},
  organization={Wiley Online Library}
}

@incollection{stylegaussian,
  title={Stylegaussian: Instant 3d style transfer with gaussian splatting},
  author={Liu, Kunhao and Zhan, Fangneng and Xu, Muyu and Theobalt, Christian and Shao, Ling and Lu, Shijian},
  booktitle={SIGGRAPH Asia 2024 Technical Communications},
  pages={1--4},
  year={2024},
}

@inproceedings{in2n,
  title={Instruct-nerf2nerf: Editing 3d scenes with instructions},
  author={Haque, Ayaan and Tancik, Matthew and Efros, Alexei A and Holynski, Aleksander and Kanazawa, Angjoo},
  booktitle={Proceedings of the IEEE/CVF international conference on computer vision},
  pages={19740--19750},
  year={2023}
}

@article{vicanerf,
  title={Vica-nerf: View-consistency-aware 3d editing of neural radiance fields},
  author={Dong, Jiahua and Wang, Yu-Xiong},
  journal={Advances in Neural Information Processing Systems},
  volume={36},
  pages={61466--61477},
  year={2023}
}

@inproceedings{consistdreamer,
  title={Consistdreamer: 3d-consistent 2d diffusion for high-fidelity scene editing},
  author={Chen, Jun-Kun and Bulo, Samuel Rota and M{\"u}ller, Norman and Porzi, Lorenzo and Kontschieder, Peter and Wang, Yu-Xiong},
  booktitle={Proceedings of the IEEE/CVF Conference on Computer Vision and Pattern Recognition},
  pages={21071--21080},
  year={2024}
}

@article{instantstylegaussian,
  title={Instantstylegaussian: Efficient art style transfer with 3d gaussian splatting},
  author={Yu, Xin-Yi and Yu, Jun-Xin and Zhou, Li-Bo and Wei, Yan and Ou, Lin-Lin},
  journal={arXiv preprint arXiv:2408.04249},
  year={2024}
}

@article{proedit,
  title={Proedit: Simple progression is all you need for high-quality 3d scene editing},
  author={Chen, Jun-Kun and Wang, Yu-Xiong},
  journal={Advances in Neural Information Processing Systems},
  volume={37},
  pages={4934--4955},
  year={2024}
}

@inproceedings{dge,
  title={Dge: Direct gaussian 3d editing by consistent multi-view editing},
  author={Chen, Minghao and Laina, Iro and Vedaldi, Andrea},
  booktitle={European Conference on Computer Vision},
  pages={74--92},
  year={2024},
  organization={Springer}
}

@inproceedings{editsplat,
  title={Editsplat: Multi-view fusion and attention-guided optimization for view-consistent 3d scene editing with 3d gaussian splatting},
  author={Lee, Dong In and Park, Hyeongcheol and Seo, Jiyoung and Park, Eunbyung and Park, Hyunje and Baek, Ha Dam and Shin, Sangheon and Kim, Sangmin and Kim, Sangpil},
  booktitle={Proceedings of the Computer Vision and Pattern Recognition Conference},
  pages={11135--11145},
  year={2025}
}

@article{gigs,
  title={Gi-gs: Global illumination decomposition on gaussian splatting for inverse rendering},
  author={Chen, Hongze and Lin, Zehong and Zhang, Jun},
  journal={arXiv preprint arXiv:2410.02619},
  year={2024}
}

@article{gusir,
  title={GUS-IR: Gaussian Splatting with Unified Shading for Inverse Rendering},
  author={Liang, Zhihao and Li, Hongdong and Jia, Kui and Guo, Kailing and Zhang, Qi},
  journal={arXiv preprint arXiv:2411.07478},
  year={2024}
}

@inproceedings{rng,
  title={Rng: Relightable neural gaussians},
  author={Fan, Jiahui and Luan, Fujun and Yang, Jian and Hasan, Milos and Wang, Beibei},
  booktitle={Proceedings of the Computer Vision and Pattern Recognition Conference},
  pages={26525--26534},
  year={2025}
}

@inproceedings{gs3,
  title={Gs3: Efficient relighting with triple gaussian splatting},
  author={Bi, Zoubin and Zeng, Yixin and Zeng, Chong and Pei, Fan and Feng, Xiang and Zhou, Kun and Wu, Hongzhi},
  booktitle={SIGGRAPH Asia 2024 Conference Papers},
  pages={1--12},
  year={2024}
}

@article{geosplating,
  title={GeoSplating: Towards Geometry Guided Gaussian Splatling for Physically-based Inverse Rendering},
  author={Ye, Kai and Gao, Chong and Li, Guanbin and Chen, Wenzheng and Chen, Baoquan},
  year={2024}
}

@article{whatsup,
  title={What's" up" with vision-language models? investigating their struggle with spatial reasoning},
  author={Kamath, Amita and Hessel, Jack and Chang, Kai-Wei},
  journal={arXiv preprint arXiv:2310.19785},
  year={2023}
}

@inproceedings{lisa,
  title={Lisa: Reasoning segmentation via large language model},
  author={Lai, Xin and Tian, Zhuotao and Chen, Yukang and Li, Yanwei and Yuan, Yuhui and Liu, Shu and Jia, Jiaya},
  booktitle={Proceedings of the IEEE/CVF Conference on Computer Vision and Pattern Recognition},
  pages={9579--9589},
  year={2024}
}

@inproceedings{glamm,
  title={Glamm: Pixel grounding large multimodal model},
  author={Rasheed, Hanoona and Maaz, Muhammad and Shaji, Sahal and Shaker, Abdelrahman and Khan, Salman and Cholakkal, Hisham and Anwer, Rao M and Xing, Eric and Yang, Ming-Hsuan and Khan, Fahad S},
  booktitle={Proceedings of the IEEE/CVF Conference on Computer Vision and Pattern Recognition},
  pages={13009--13018},
  year={2024}
}

@inproceedings{llava_grounding,
  title={Llava-grounding: Grounded visual chat with large multimodal models},
  author={Zhang, Hao and Li, Hongyang and Li, Feng and Ren, Tianhe and Zou, Xueyan and Liu, Shilong and Huang, Shijia and Gao, Jianfeng and Leizhang and Li, Chunyuan and others},
  booktitle={European Conference on Computer Vision},
  pages={19--35},
  year={2024},
  organization={Springer}
}

@article{vpp_llava,
  title={Visual Position Prompt for MLLM based Visual Grounding},
  author={Tang, Wei and Sun, Yanpeng and Gu, Qinying and Li, Zechao},
  journal={arXiv preprint arXiv:2503.15426},
  year={2025}
}

@inproceedings{vqav2,
  title={Making the v in vqa matter: Elevating the role of image understanding in visual question answering},
  author={Goyal, Yash and Khot, Tejas and Summers-Stay, Douglas and Batra, Dhruv and Parikh, Devi},
  booktitle={Proceedings of the IEEE conference on computer vision and pattern recognition},
  pages={6904--6913},
  year={2017}
}

\end{document}